\documentclass[lettersize,journal]{IEEEtran}
\usepackage[caption=false,font=normalsize,labelfont=sf,textfont=sf]{subfig}
\usepackage{amsmath,amsfonts,amssymb} % AMS 数学公式和符号
\usepackage{url} % 网址排版
\usepackage{textcomp} % 文本符号
\usepackage{verbatim} % 文本原样输出

\usepackage{array} % 数组和表格
\usepackage{booktabs} % 更好的三线表
\usepackage{multirow} % 合并表格单元格
\usepackage{makecell} % 表格单元格内换行

\usepackage{float} % 强制浮动体位置
\usepackage{adjustbox} % 调整图片大小和旋转子图排版
\usepackage{textcomp}
\usepackage{stfloats}
\usepackage{url}
\usepackage{verbatim}
\usepackage{graphicx}
\usepackage{adjustbox}

\usepackage{algorithm} % 伪代码浮动体
\usepackage{algorithmicx} % 先加载 algorithmicx
\usepackage{algpseudocode} % 伪代码语法

\usepackage{threeparttable}
\usepackage{amsmath} % 确保在算法包之前加载
\usepackage{ulem} % 更好的下划线支持
\usepackage{xcolor} % 颜色
\usepackage{setspace} % 设置行距
\usepackage[T1]{fontenc} % 字体编码
\usepackage{etoolbox} % 定义新命令
\usepackage[marginal]{footmisc} % 脚注排版
\usepackage{balance} % 最后一页平衡栏目
\usepackage{footnote}
\usepackage{enumitem} % 需要加载这个宏包

\usepackage[colorlinks,
            linkcolor=blue,
            citecolor=blue]{hyperref}
\makeatletter
\renewcommand{\@cite}[1]{\textcolor{blue}{[#1]}}
\makeatother
\usepackage{tablefootnote}

\usepackage{xcolor}
\algrenewcommand{\algorithmiccomment}[1]{%
  \hfill\textcolor{green!50!black}{$\triangleright$~#1}%
}
\newcommand{\best}[1]{\textcolor{red}{\textbf{#1}}}
\newcommand{\secondbest}[1]{\textcolor{blue}{\underline{#1}}}

\usepackage{booktabs}  % \toprule, \midrule, \bottomrule, \cmidrule
\usepackage{graphicx}  % \resizebox
\usepackage{placeins}  % \FloatBarrier
\usepackage{booktabs}
\usepackage{tabularx}
\usepackage{makecell}
\usepackage{pifont}
\usepackage{array}

\newcommand{\cmark}{\ding{51}}
\newcommand{\xmark}{\ding{55}}

\begin{document}

\title{Revisiting Identity and Spectra Dispersion in Media-Bridged Time Series Forecasting: Linking Multivariate Signals and Narrative Flows}

\author{Jierui Lei, Wenjian Zhang, Qingyi Yang, Yuyang Hong, Fangzheng Chen, Zhengbo Zhang, \\ Haina Tang, Shiming Xiang
\thanks{Jierui Lei and Shiming Xiang are with the School of Advanced Interdisciplinary Sciences, University of Chinese Academy of Sciences, Beijing 100049, China, and also with the State Key Laboratory of Multimodal Artificial Intelligence Systems, Institute of Automation, Chinese Academy of Sciences, Beijing 100190, China (e-mail: leijierui24@mails.ucas.ac.cn, smxiang@nlpr.ia.ac.cn).}
\thanks{Wenjian Zhang is with the School of Intelligent Systems Engineering, Sun Yat-sen University, Shenzhen,  Guangdong  518107, China (e-mail: zhangwj289@mail2.sysu.edu.cn).}
\thanks{Qingyi Yang is with the School of Computer Science and Technology, Huazhong University of Science and Technology, Wuhan, Hubei 430074, China (e-mail: yangqingyi@hust.edu.cn).}
\thanks{Yuyang Hong and Zhengbo Zhang are with the State Key Laboratory of Multimodal Artificial Intelligence Systems, Institue of Automation, Chinese Academy of Sciences, Beijing 100190, China, and also with the School of Artificial Intelligence, University of Chinese Academy of Sciences, Beijing 100049, China (e-mail: hongyuyang2023@ia.ac.cn, zhangzhengbo2025@ia.ac.cn).}
\thanks{Fangzheng Chen is with the College of Computer Science, Chongqing University, Chongqing 401331, China (e-mail: 202414131029T@stu.cqu.edu.cn).}
\thanks{Haina Tang  (\textit{Corresponding author}) is with the School of Artificial Intelligence, University of Chinese Academy of Sciences, Beijing 100049, China (e-mail: hntang@ucas.ac.cn).}}
\markboth{Journal of \LaTeX\ Class Files, Vol. 14, No. 8, August 2021}
{Shell \MakeLowercase{\textit{et al.}}: Bare Demo of IEEEtran.cls for IEEE Journals}
\maketitle

\begin{abstract}
Media-bridged time series forecasting is expanding to encompass traditional "multivariate" and emerging "multimodal" (e.g., through textual assistance). Existing Time Series Forecasting (TSF) models still rely on paradigm-specific relation, fusion,
and temporal modules, hindering a common forecasting backbone across
numerical and pre-aligned narrative-flow settings. To explore this, we propose the Multimedia Identity-Aware Prism Network (MIDAPN), a unified spatiotemporal forecasting backbone based on media-general graph adaptation and automatic temporal learning: (1) Following media pre-alignment, our Multimedia Identity-Aware Graph (MIDAG) revisits identity through  static essence, dynamic behavior, and latent commonality, inducing affinities that extend variable-specific dependencies across media. Contextual Identity Modulation (CIM) further refines discriminative aggregation. (2) We develop Spectral Prism Convolution (SPConv) to automatically perform hierarchical temporal analysis, balancing coarse trends and fine-grained details. Meanwhile, its Adaptive Search Guidance configures a scale-efficient architecture for temporal-dimension reconstruction. These decoupled yet synergistic components jointly address media identity disentanglement and temporal-scale mismatch. Comprehensive evaluations involving 16 SOTA TSF models across 13 "multivariate" and 12 "multimodal" datasets, alongside targeted long-context comparisons against 14 time series foundation models and fused pretrained language models, demonstrate MIDAPN's consistent superiority and broad shared backbone compatibility. The code is available at \href{https://github.com/leijieruilq/MIDAPN/tree/main}{https://github.com/MIDAPN}.
\end{abstract}


\section{Introduction}
\IEEEPARstart{W}{ith} rapid development of sensing  \cite{jin2024survey,tnnls-1,tnnls-3} and multimodal applications \cite{baltruvsaitis2018multimodal,jiang2025multi,tnnls-4} in Web-enabled IoT ecosystems, real-world temporal representations are increasingly enriched by diverse auxiliary media \cite{liu2024time,shao2024exploring}. In essence, this media-bridged view of time series analysis spans two representative paradigms: horizontal "multivariate" association \cite{shao2024exploring,yizhixing} for homogeneous numerical temporal variables \cite{wu2020connecting,jin2024survey} and vertical heterogeneous "multimodal" \cite{liu2024time,zhong2025time,mm25} for semantic enhancement within external modalities. Since multi-domain perceptual datasets (e.g., images, audio) remain scarce \cite{ekambaram2020attention,chen2024terra}, aligning time series with cognitive text currently serves as
a broadly adopted "multimodal" setting \cite{liu2024time,li2025language} (i.e., "narrative-flow"). However, current TSF backbones are typically tailored to either multivariate  or multimodal forecasting, with few architectures jointly addressing spatiotemporal dependencies and modality heterogeneity \cite{shao2024exploring,jiang2025multi}.

To overcome the above question, we seek from a forecasting perspective: multivariate backbone models dependencies among numerical variates, whereas narrative-flow multimodal backbone augments temporal observations with aligned textual semantics. Despite this difference in conditioning information, both aim to predict future numerical states. Existing task-specific architectures nevertheless hard-code source differences into customized relation, fusion, and temporal modules \cite{DUET,msgnet,zhong2025time,calf}, limiting shared post-alignment. Fortunately, current pre-alignment casts numerical channels and PLM-derived text representations as time-aligned variates, making a unified forecasting backbone feasible \cite{li2025language}. However, input-level alignment does not imply modeling-level unification: heterogeneous variates still differ in \emph{what they represent} and \emph{how they evolve}. Therefore, a media-general backbone should move beyond input-level and address two key modeling challenges in spatiotemporal decoupling: \textbf{\emph{how to distinguish \& relate media units}, and \emph{how to organize temporal scales}}.

\begin{figure}[t]
\centering
\includegraphics[width=\linewidth]{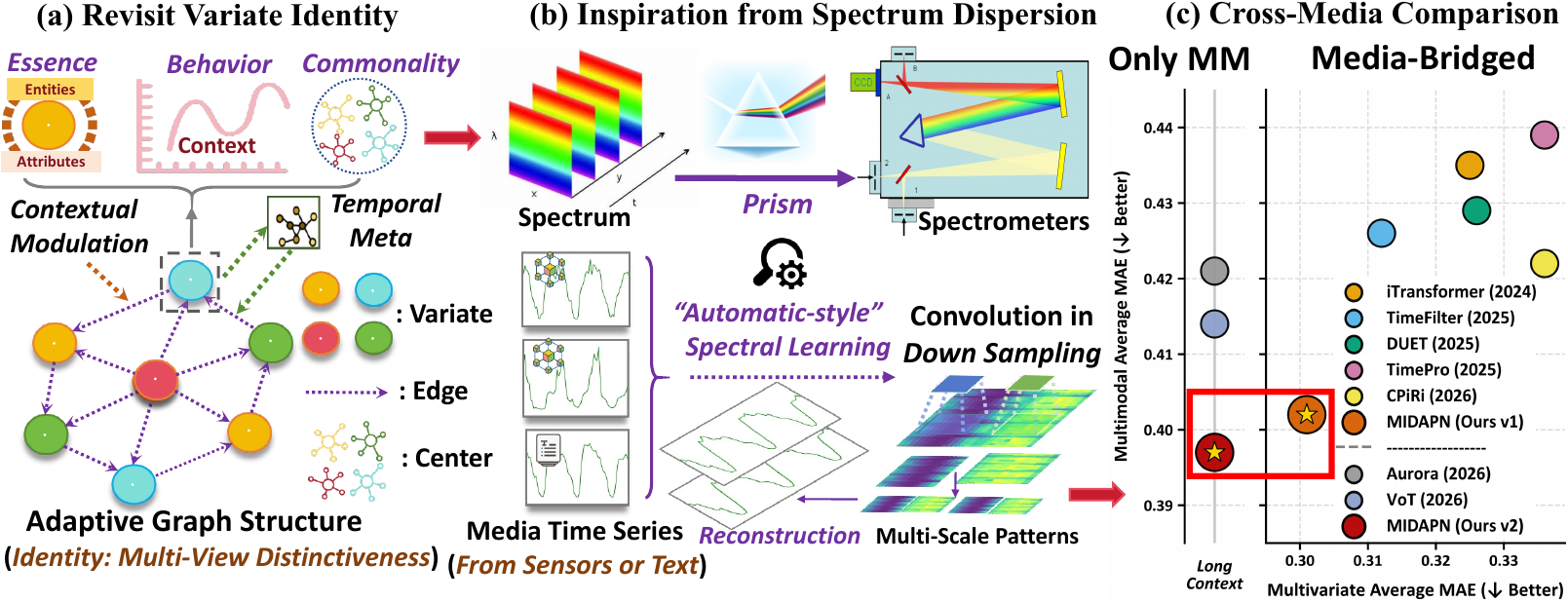}
{\renewcommand{\arraystretch}{0.8}
\resizebox{1.0\linewidth}{!}{
\begin{tabular}{c|cc|cc|c}
\toprule[1.2pt]
\multicolumn{6}{c}{
\textbf{(d) MIDAPN's Controlled Replacement (MIDAG \& SPConv)}
} \\
\midrule
\textbf{Media-Bridged Task}
& \multicolumn{2}{c|}{\textbf{Multivariate}}
& \multicolumn{2}{c|}{\textbf{Multimodal (Time-MMD)}}
& \textbf{Multivariate+Text} \\
\textbf{Replacement ($\uparrow$ Error, \%)}
& Electricity & Weather & Traffic & Energy & LEU \\
\midrule
MIDAG $\rightarrow$ Adp-Graph
& +4.63 & +0.76 & +5.52 & +2.82 & +13.06 \\
SPConv $\rightarrow$ Conv/Freq-Conv
& +14.23/+3.81
& +5.62/+2.79
& +1.27/+4.16
& +4.37/+2.76
& +3.50/+3.74 \\
\bottomrule[1.2pt]
\end{tabular}
}}
\caption{Motivations from identity awareness and temporal capturing
(spatiotemporal learning). MIDAPN achieves cross-media compatibility
across multivariate and multimodal forecasting, while controlled
replacements of MIDAG or SPConv with generic counterparts consistently
increase errors}
\label{intro-midag-spcn}
\end{figure}

(1) \textit{Media-General Identity Entanglement (distinguish \& relate):} From the data-driven perspective, potential yet true spatio-temporal correlations \cite{jin2024survey} should be accomplished during training. Hence the data-driven adaptive graph \cite{gwn,wu2020connecting,tnnls-5} captures multivariate relationships via pairwise affinities among trainable variable embeddings. For instance, causal relationships in changes of air pressure and temperature, and mutual influences of electricity consumption among residents in different regions \cite{tnnls-7}. Despite the absence of explicit topological priors in heterogeneous time series, leveraging adaptive graphs \cite{gwn,megacrn,crossgnn,sageformer,msgnet,agga-mvfln, Lei2026} to learn variable relationships in data-driven form, and pruning redundant correlations within clustering or masked learning strategies \cite{cluster,STD-MAE,DUET,hu2025timefilter} to distinguish spatio-temporal "hallucinations" have become increasingly prevalent approaches. However, these approaches commonly rely on monolithic, entangled variate representations, weakening identity distinctiveness across heterogeneous sources. Conceptually, variate identity is not merely a categorical identifier; it denotes \emph{what it is}, \emph{how it manifests}, and \emph{where it belongs}. In view of this, we rethink Multimedia Identity Graph (MIDAG) adaptation via "essence", "behavior", and "commonality" decomposition. Furthermore, we introduce Contextual Identity Modulation to preserve distinctiveness during graph-conditional aggregation.

(2) \textit{Temporal Scale Mismatch \& Automatic Adaptation (organize):} To efficiently mine temporal patterns, frequency-domain analysis \cite{fremlp,filterts} decomposes complex signals for band-specific learning, while Temporal Convolutional Networks (TCNs) exploit local connectivity and weight sharing for parameter-efficient modeling. Generally, current improvements to TCN primarily focus on expanding receptive fields \cite{micn,SCTNet}, incorporating inter-variable dependencies \cite{moderntcn}, multi-period learning \cite{timesnet}, and stacking convolutions in the frequency domain \cite{DF} to achieve multi-level representation fusion.  However, these incremental modifications have failed to transform TCNs into efficient "automatic-style" architectures, as fixed convolutional hierarchies may not adapt well to varying spectral resolutions. We therefore investigate \textit{whether convolutional hierarchies can be flexibly
tailored based on temporal scales}. Inspired by spectrum dispersion as a conceptual analogy \cite{prism}, we
propose Spectral Prism Convolution (SPConv) with an automatic
``search-by-construction''. Analogous to decomposing composite spectra,
Fourier transforms separate interwoven temporal representations into
frequency components for hierarchical modeling preparation. To achieve scale-efficient  architecture configuration \cite{opt,chain}, an "Adaptive Search Guidance Module" (ASGM) determines the layer depth and channel expansion strategies, enabling targeted temporal reconstruction. Finally, a hierarchical "compression-generation" structure is implemented, functioning as a scale-adaptive "spectrometer" across the spectrum.

In summary, we focus on \textbf{broadly transferable multivariate and text-assisted forecasting} in media-bridged time series, and propose the Multimedia Identity-Aware Prism Network (MIDAPN), a spatiotemporally decoupled single-backbone model trained separately for each task, realizing \textbf{\textit{architecture-level generality without task-specific redesign}}. As shown in Fig. \ref{intro-midag-spcn}(a, b), the core spatiotemporal motivations are driven by identity exploration within adaptive graphs and automated temporal learning based on convolutional composition. Consequently, Fig. \ref{intro-midag-spcn}(c) demonstrates its compatibility across 13 multivariate and 7 multimodal representative benchmarks. Notably, under multimodal-only settings, it consistently outperforms the multimodal time series foundation model Aurora \cite{aurora}, and multi-level fusion alignment VoT \cite{wang2026unlocking}. Moreover, Fig. \ref{intro-midag-spcn}(d) shows that replacing MIDAG or SPConv with generic counterparts consistently increases errors, validating their complementary roles in identity-aware relation modeling and scale-adaptive temporal modeling. Our contributions are:

\noindent $\bullet$ We formulate media-bridged time series forecasting as an underexplored unified task spanning "multivariate" and narrative-flow "multimodal", proposing MIDAPN from media-general variate adaptation and temporal learning.

\noindent $\bullet$ For multimedia correlation learning, we introduce an Identity-Aware Graph to address identity entanglement by revisiting ``variate identity'' through essence, behavior, and commonality, thereby preserving multi-view distinctiveness for spatiotemporal correlations and modality heterogeneity.

\noindent $\bullet$ For automatic temporal learning, we propose Spectral Prism Convolution with hierarchical compression-generation and an Adaptive Search Guidance Module, enabling multi-scale modeling and "search-by-construction" architecture configuration to address scale mismatch across varying forecasting horizons.
    
\noindent $\bullet$ MIDAPN achieves superior results with empirical comparison on 25 real-world "multivariate" and "multimodal" datasets.

\begin{table*}[t]
\centering
\caption{
Comparison with representative multivariate and multimodal models.``+TaTS'' denotes its TaTS-based
multimodal extension.
}
\label{tab:related_comparison}

\setlength{\tabcolsep}{2.6pt}
\renewcommand{\arraystretch}{1.12}

\resizebox{0.9\textwidth}{!}{
\begin{tabular}{lccccccccc}
\toprule

\textbf{Method}
& \textbf{Year}
& \makecell{\textbf{External Paired}\\\textbf{Modality}}
& \makecell{\textbf{Pretrained}\\\textbf{PLM/TSFM}}
& \makecell{\textbf{Explicit Cross-}\\\textbf{Variate Relation}}
& \makecell{\textbf{Intermediate Cross-}\\\textbf{Modal Interaction}}
& \makecell{\textbf{Identity/Group-Aware}\\\textbf{Relation Modeling}}
& \makecell{\textbf{Frequency-Domain}\\\textbf{Modeling}}
& \makecell{\textbf{Explicit Multi-Resolution}\\\textbf{Temporal Modeling}}
& \makecell{\textbf{Search-Guided Hierarchical}\\\textbf{Reconstruction}}
\\

\midrule
\multicolumn{10}{l}{
\textit{Native multivariate models and their TaTS-based extensions}
}
\\

MSGNet~\cite{msgnet}
& 2024
& \xmark
& \xmark
& \cmark
& \xmark
& \xmark
& \cmark
& \cmark
& \xmark
\\

\quad\textit{+ TaTS}~\cite{li2025language}
& 2026
& \cmark
& \cmark
& \cmark
& \cmark
& \xmark
& \cmark
& \cmark
& \xmark
\\

\addlinespace[2pt]

TimeFilter~\cite{hu2025timefilter}
& 2025
& \xmark
& \xmark
& \cmark
& \xmark
& \xmark
& \xmark
& \xmark
& \xmark
\\

\quad\textit{+ TaTS}~\cite{li2025language}
& 2026
& \cmark
& \cmark
& \cmark
& \cmark
& \xmark
& \xmark
& \xmark
& \xmark
\\

\addlinespace[2pt]

DUET~\cite{DUET}
& 2025
& \xmark
& \xmark
& \cmark
& \xmark
& \cmark
& \cmark
& \xmark
& \xmark
\\

\quad\textit{+ TaTS}~\cite{li2025language}
& 2026
& \cmark
& \cmark
& \cmark
& \cmark
& \cmark
& \cmark
& \xmark
& \xmark
\\

\addlinespace[2pt]

VPNet~\cite{vpnet}
& 2026
& \xmark
& \xmark
& \cmark
& \xmark
& \xmark
& \xmark
& \xmark
& \xmark
\\

\quad\textit{+ TaTS}~\cite{li2025language}
& 2026
& \cmark
& \cmark
& \cmark
& \cmark
& \xmark
& \xmark
& \xmark
& \xmark
\\

\addlinespace[2pt]

CrossLinear~\cite{crosslinear}
& 2025
& \xmark
& \xmark
& \cmark
& \xmark
& \xmark
& \xmark
& \xmark
& \xmark
\\

\quad\textit{+ TaTS}~\cite{li2025language}
& 2026
& \cmark
& \cmark
& \cmark
& \cmark
& \xmark
& \xmark
& \xmark
& \xmark
\\

\addlinespace[2pt]

CPiRi~\cite{cpiri}
& 2026
& \xmark
& \cmark
& \cmark
& \xmark
& \xmark
& \xmark
& \xmark
& \xmark
\\

\quad\textit{+ TaTS}~\cite{li2025language}
& 2026
& \cmark
& \cmark
& \cmark
& \cmark
& \xmark
& \xmark
& \xmark
& \xmark
\\

\midrule
\multicolumn{10}{l}{
\textit{Multimodal and PLM/TSFM-based forecasting methods}
}
\\

MM-TSFlib~\cite{liu2024time}
& 2024
& \cmark
& \cmark
& \xmark
& \xmark
& \xmark
& \xmark
& \xmark
& \xmark
\\

GPT4MTS~\cite{jia2024gpt4mts}
& 2024
& \cmark
& \cmark
& \xmark
& \cmark
& \xmark
& \xmark
& \xmark
& \xmark
\\

CALF~\cite{calf}
& 2025
& \xmark
& \cmark
& \xmark
& \cmark
& \xmark
& \xmark
& \xmark
& \xmark
\\

Time-VLM~\cite{zhong2025time}
& 2025
& \xmark
& \cmark
& \xmark
& \cmark
& \xmark
& \xmark
& \xmark
& \xmark
\\

ChatTime~\cite{chattime}
& 2025
& \cmark
& \cmark
& \xmark
& \cmark
& \xmark
& \xmark
& \xmark
& \xmark
\\

TaTS~\cite{li2025language}
& 2026
& \cmark
& \cmark
& \xmark
& \cmark
& \xmark
& \xmark
& \xmark
& \xmark
\\

Aurora~\cite{aurora}
& 2026
& \cmark
& \cmark
& \xmark
& \cmark
& \xmark
& \xmark
& \xmark
& \xmark
\\

SE-LLM~\cite{sellm}
& 2026
& \xmark
& \cmark
& \cmark
& \cmark
& \xmark
& \xmark
& \xmark
& \xmark
\\

VoT~\cite{wang2026unlocking}
& 2026
& \cmark
& \cmark
& \xmark
& \cmark
& \xmark
& \cmark
& \xmark
& \xmark
\\

\midrule

\textbf{MIDAPN}
& \textbf{(Ours)}
& \cmark
& \cmark
& \cmark
& \cmark
& \cmark
& \cmark
& \cmark
& \cmark
\\

\bottomrule
\end{tabular}
}
\end{table*}

\section{Related Work}
\label{rwork}
\subsection{Multivariate Time Series Forecasting}

Multivariate Time Series Forecasting aims to provide comprehensive, multi-channel forecasts across diverse application domains. Within this field, deep learning-based techniques have become the mainstream paradigm as they significantly outperform in terms of predictability and parallelization efficiency. Early deep learning models, typified by RNNs, saw improvements such as SegRNN \cite{segrnn} and P-sLSTM \cite{unlock}, which accelerated RNN iterations through segmental recursion and parallel multi-step prediction; TCN-based architectures \cite{micn,timesnet,DF,moderntcn,SCTNet} have primarily concentrated on expanding the receptive field, incorporating inter-variable dependencies, and leveraging multi-period learning in the frequency domain for more comprehensive multi-scale adaptability.

Concurrently, Transformer- and MLP-based architectures have emerged as popular candidates in the time series community. Initial Transformer-based methods sought to enhance the encoder-decoder's ability to capture diverse temporal dependencies through various strategies, including frequency-domain transforms \cite{freE} and channel-independent patching \cite{patchtst}. Current SOTA methods \cite{itransformer,sageformer,DUET}, however, center more on integrating inter-variable interactions into the Transformer to enhance its ability to learn inter-series correlations. The proliferation of Transformer-based methods has paved the way for fine-tuning Pre-trained Large Language Models \cite{timellm,zhong2025time} or universal Time Series Foundation Models (TSFM) \cite{woo2024unified,liu2025sundial}. At the same time, MLP-based architectures have achieved surprising results by leveraging their simplicity and efficiency, directly integrating various strategies for modeling periodicity (e.g., decomposition \cite{dlinear}, frequency filtering \cite{filterts}, multi-scale mixing \cite{timemixer}) and variable relationships (e.g., channel-independent and mixing \cite{han2024softs}). Furthermore, a few studies are exploring the multi-order Kolmogorov-Arnold Networks \cite{timekan} or the Mamba architecture \cite{timepro} to reconstruct state spaces for variable-specific information and temporal patterns.

\subsection{Multimodal Time Series Forecasting}

Multimodal Time Series Forecasting aims to improve forecasting accuracy
across diverse domains by integrating timestamp-aligned auxiliary information
with time series~\cite{jiang2025multi,zhong2025time}, typically through joint
or coordinated representations~\cite{baltruvsaitis2018multimodal,tnnls-4}.
Given the broad availability and semantic richness of textual narratives,
recent studies increasingly use text as an auxiliary modality, aligning
natural-language semantics with temporal signals to support
``understanding time series with natural language''
\cite{jia2024gpt4mts,zhong2025time}. To facilitate research on multi-domain
multimodal forecasting, several representative benchmarks and frameworks
have emerged. Time-MMD~\cite{liu2024time} spans nine application domains
and introduces MM-TSFlib, which integrates the two modalities through
weighted dual-branch alignment. In parallel, TaTS (Texts as Time Series)
\cite{li2025language} revisits the Platonic representation hypothesis by
casting paired text as auxiliary variates, thereby transforming heterogeneous
multimodal inputs into a structured multivariate form.   

However, recent TSF models still rarely use the same backbone to handle both
``multivariate'' and ``multimodal'' forecasting
\cite{jiang2025multi,mm25}. Even most TSFMs mainly pursue general-purpose forecasting
\cite{blast,gtm}, while multimodal generalization remains less common,
with Aurora representing a recent step in this direction~\cite{aurora}. Table~\ref{tab:related_comparison} illustrates the difference. With
TaTS~\cite{li2025language}, native multivariate backbones can readily
accommodate text-derived variates, but their relational and temporal modeling
largely follow the original design, with limited adaptation to heterogeneous
sources after pre-alignment. Meanwhile, multimodal and PLM-based methods
focus more on semantic alignment and fusion
\cite{liu2024time,calf,zhong2025time,wang2026unlocking}.
MIDAPN targets this post-alignment stage by jointly modeling which
heterogeneous variates interact (MIDAG) and how their relations evolve
across temporal scales (SPConv).

\section{Preliminaries and Media Pre-alignment}
\label{preli}
We formulate the media-bridged time series forecasting by considering its "multivariate" and "multimodal" pre-alignment.

\noindent \textbf{Task 1: Multivariate Time Series Forecasting.} Given a Look-back window of multivariate time series $\textbf{X}_{in}=\{x^{1}_{his}, x^{2}_{his},...,x^{C}_{his}\}$ $\in \mathbb{R}^{C \times T_{h}}$, where $T_{h}$ represents the length of Look-back window, and $C$ denotes the number of variables (Note: the inherent time-related features, e.g. [month, day, week, hour] of timestamp variables can also be flexibly incorporated as covariates $T_{\text{mark}} \in \mathbb{R}^{4}$ \cite{itransformer}). For a time series forecasting model $\mathcal{F}_{\{ \theta_{\text{TSF}} \}}(\cdot)$, the objective is to predict the future $\textbf{X}_{out}$ $\in \mathbb{R}^{C \times T_{f}}$ over a horizon step $T_{f}$ through $\textbf{X}_{in}$.

\noindent \textbf{Task 2: Multimodal Time Series Forecasting (TaTS).} Given paired $\mathcal{D}= \{ \textbf{X},\textbf{S} \}$, where $\textbf{S}_{in}=\{s_{1}, s_{2},..., s_{T_{h}}\}$ represents the input auxiliary text modality and each text input $s_{t}$ is subsequently encoded using a frozen Pre-trained Language Model (PLM). The resulting tokens are then aggregated via average pooling to obtain textual representation $\textbf{V}_{in}=\{v_{1}, v_{2}, ... , v_{T_{h}}\}$ $\in \mathbb{R}^{d_{\text{text}} \times T_{h}}$. Furthermore, $\text{MLP}(\cdot)$ is utilized for dimensionality reduction on $\textbf{V}_{in}$. This projection serves to balance the different modality representations and obtain $\textbf{Z}_{in}=\{z_{1}, z_{2}, ... , z_{T_{h}}\}$ $\in \mathbb{R}^{d_{\text{tproj}} \times T_{h}}$. Here, $\textbf{Z}_{in}$ are treated as auxiliary variables. Concretely, $\textbf{Z}_{in}$ is concatenated with the temporal modality $\textbf{X}_{in}$ to form a unified multimodal sequence $\textbf{U}_{in}=\text{Concat}[\textbf{X}_{in} \mid \mid \textbf{Z}_{in}]$ $\in \mathbb{R}^{(d_{\text{tproj}} + C_{ts}) \times T_{h}}$ in multivariate structure ($C = d_{\text{tproj}} + C_{ts}$). In this way, the "resonance" characteristics are mined between textual periodicity and temporal dynamics. Following TaTS \cite{li2025language}, whose external
text-dropping/-shuffling have established its utility, we \textbf{use the same pre-aligned
$\textbf{U}_{in}$ for all DL TSF backbones
(Table~\ref{tab:comprehensive_merged_categorized}), focusing on
intermediate fusion} to predict
$\textbf{X}_{out}\in\mathbb{R}^{C_{ts}\times T_f}$.

By utilizing a unified forecast backbone $\mathcal{F}_{\{ \theta_{\text{TSF}} \}}(\cdot)$, our goal is to achieve superior media-bridged performance rather than relying on meticulously designed modal-assisted alignments.

\section{Methodology}
\label{method}

\subsection{Overall Architecture and Theoretical Analysis}
\begin{figure*}
\centering
\includegraphics[width=0.9 \linewidth]{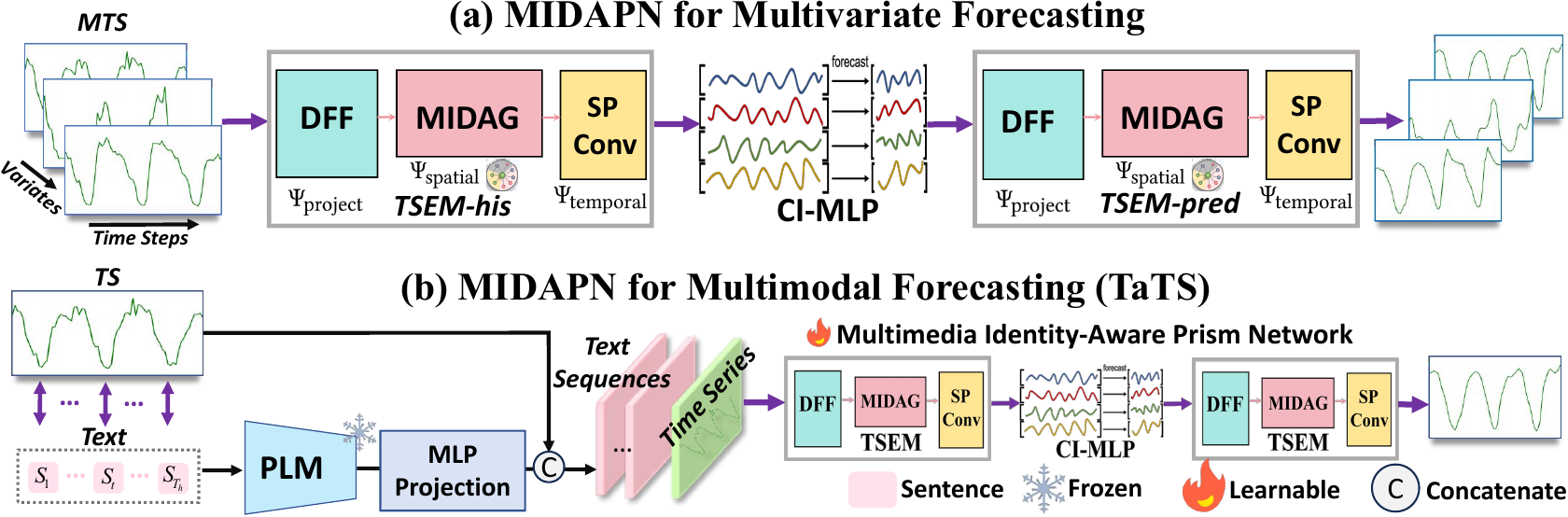}
\caption{MIDAPN architecture. (a) Workflow for "multivariate" scenarios. (b) Workflow for TaTS-style "multimodal" scenarios. In both settings, TSEM-his first extracts
spatiotemporal representations through Decoupled Frequency Fusion,
Multimedia Identity-Aware Graph, and Spectral Prism Convolution from historical windows, then a channel-independent MLP maps the history to the 
future, followed by TSEM-pred for forecast refinement (details in Sec.\ref{method-3})}
\label{VTITG-ASFFN-model}
\end{figure*}

Fig. \ref{VTITG-ASFFN-model} illustrates MIDAPN's "encode–project–refine" architecture for media-bridged forecasting following Sec. \ref{preli}, where PLM supports multimodal pre-alignment. Specifically, the Time Series Embedding Module (TSEM) integrates three components for spatiotemporal decoupling: Decoupled Frequency Fusion (DFF), Multimedia Identity-Aware Graph (MIDAG) and Adaptive Search-Guided Spectral Prism Convolution (SPConv). DFF provides feature projection, while MIDAG and SPConv perform relation learning and temporal scaling, respectively. $\text{TSEM}_{\text{his}}$ first extracts past representations, a Channels-Independent MLP (CI-MLP) projects them to the future horizon, and $\text{TSEM}_{\text{pred}}$ refines the forecast.
To interpret TSEM from an \textbf{\textit{ST-manifold operator perspective}}:
\begin{equation}
\begin{aligned}
\bigl(\Psi_{\mathrm{spatial}}\circ\Psi_{\mathrm{project}}\bigr)(X)
&=
\operatorname{MIDAG}\!\left(\operatorname{DFF}(X)\right),\\
\Psi_{\mathrm{temporal}}\!\left(X;X_{\mathrm{MID}}\right)
&=
\operatorname{SPConv}\!\left(X,X_{\mathrm{MID}}\right).
\end{aligned}
\end{equation}
$\text{DFF}(\cdot)$ achieves flexible projection $\Psi_{\mathrm{project}}$, followed by $\text{MIDAG}(\cdot)$ performing spatial discovery on the variable manifold $\Psi_{\mathrm{spatial}}$, then $\text{SPConv}(\cdot)$ reorganizing topology-conditioned extracting on the temporal manifold $\Psi_{\mathrm{temporal}}$.

This manifold decomposition motivates deeper evaluation of relational and
temporal parts. Let
$\mathcal{R}(f)=\mathbb{E}[\|Y-f(X)\|_F^2]$
and $f^\star(X)=\mathbb{E}[Y\mid X]$
denote the forecasting risk and Bayes predictor. We consider the full MIDAPN
with its variants: $f_{\mathrm{T}}$ removes MIDAG-based relation
modeling, $f_{\mathrm{S}}$ removes explicit SPConv temporal modeling,
and $f_{\mathrm{ST}}$ retains both:
\begin{equation}
\resizebox{0.88\linewidth}{!}{$
\begin{aligned}
X_{\mathrm{MID}}&=\operatorname{MIDAG}(\operatorname{DFF}(X)),\quad
f_{\mathrm{T}}(X)=\operatorname{SPConv}(X,X),\\
f_{\mathrm{S}}(X)&=\text{MLP}(X_{\mathrm{MID}}),\quad
f_{\mathrm{ST}}(X)=\operatorname{SPConv}(X,X_{\mathrm{MID}}).
\end{aligned}
$}
\label{eq:restricted_predictors}
\end{equation}

\noindent\textbf{Proposition 1 (Conditional necessity of MIDAG).}
Define the spatial approximation gap as
\begin{equation}
\Delta_{\mathrm{s}}
=
\inf_{f\in\mathcal{F}_{\mathrm{T}}}\mathcal{R}(f)
-
\mathcal{R}(f^\star).
\end{equation}
\textit{If the target contains \textbf{non-trivial cross-variate or cross-media
dependencies that cannot be approximated arbitrarily well by
$\mathcal{F}_{\mathrm{T}}$}, then $\Delta_{\mathrm{s}}>0$}.
Consequently, any temporal-only predictor incurs
\begin{equation}
\mathcal{R}(f)-\mathcal{R}(f^\star)
\geq
\Delta_{\mathrm{s}},
\qquad
f\in\mathcal{F}_{\mathrm{T}},
\end{equation}
which establishes the necessity of MIDAG.

\noindent\textbf{Proposition 2 (Conditional necessity of SPConv).}
Define the temporal approximation gap as
\begin{equation}
\Delta_{\mathrm{t}}
=
\inf_{f\in\mathcal{F}_{\mathrm{S}}}\mathcal{R}(f)
-
\mathcal{R}(f^\star).
\end{equation}
\textit{If the target contains \textbf{multi-scale temporal dynamics that cannot be
recovered solely from $\mathcal{F}_{\mathrm{S}}$}, then
$\Delta_{\mathrm{t}}>0$}. Consequently,
\begin{equation}
\mathcal{R}(f)-\mathcal{R}(f^\star)
\geq
\Delta_{\mathrm{t}},
\qquad
f\in\mathcal{F}_{\mathrm{S}},
\end{equation}
which establishes the necessity of SPConv after MIDAG.

\noindent\textbf{Corollary 1 (Conditional MIDAG--SPConv complementarity).}
Assume that removing MIDAG or SPConv induces approximation gaps
$\Delta_{\mathrm{s}}>0$ and $\Delta_{\mathrm{t}}>0$, respectively.
Let
$f_{\mathrm{ST}}^\star(X)=\operatorname{SPConv}(X,S^\star(X))$
denote the predictor conditioned on the ideal topology representation
$S^\star(X)$, and define the topology-induced and temporal residuals as
$r_{\mathrm{s}}(X)
=
f_{\mathrm{ST}}(X)-f_{\mathrm{ST}}^\star(X),
r_{\mathrm{t}}(X)
=
f_{\mathrm{ST}}^\star(X)-f^\star(X)$.
Let
$\epsilon_{\mathrm{s}}
=\mathbb{E}\|X_{\mathrm{MID}}-S^\star(X)\|_F^2$
and
$\epsilon_{\mathrm{t}}
=\mathbb{E}\|r_{\mathrm{t}}(X)\|_F^2$.
If SPConv is $L$-Lipschitz in its topology-conditioned input, then
$\mathbb{E}\|r_{\mathrm{s}}(X)\|_F^2
\leq L^2\epsilon_{\mathrm{s}}$.
Using the Bayes-risk decomposition and
$\|a+b\|_F^2\leq2\|a\|_F^2+2\|b\|_F^2$, we obtain
\begin{equation}
\begin{aligned}
\mathcal{R}(f_{\mathrm{ST}})-\mathcal{R}(f^\star)
&=
\mathbb{E}\|r_{\mathrm{s}}(X)+r_{\mathrm{t}}(X)\|_F^2\\
&\leq
2\mathbb{E}\|r_{\mathrm{s}}(X)\|_F^2
+2\mathbb{E}\|r_{\mathrm{t}}(X)\|_F^2\\
&\leq
2L^2\epsilon_{\mathrm{s}}+2\epsilon_{\mathrm{t}}.
\end{aligned}
\label{eq:full_bound}
\end{equation}
Consequently, whenever
$
2L^2\epsilon_{\mathrm{s}}+2\epsilon_{\mathrm{t}}
<
\min\{\Delta_{\mathrm{s}},\Delta_{\mathrm{t}}\},
$
$f_{\mathrm{ST}}$ outperforms either ablated class.
MIDAG identifies \emph{which variables interact}, whereas SPConv
models \emph{how these interactions evolve across temporal scales}.
Thus, they are axis-decoupled but conditionally complementary for
spatiotemporal learning.

\subsection{$\Psi_{\text{project}}$: Decoupled Frequency Fusion}
\label{method-1-1}
\begin{figure*}
\centering
\includegraphics[width=1.0 \linewidth]{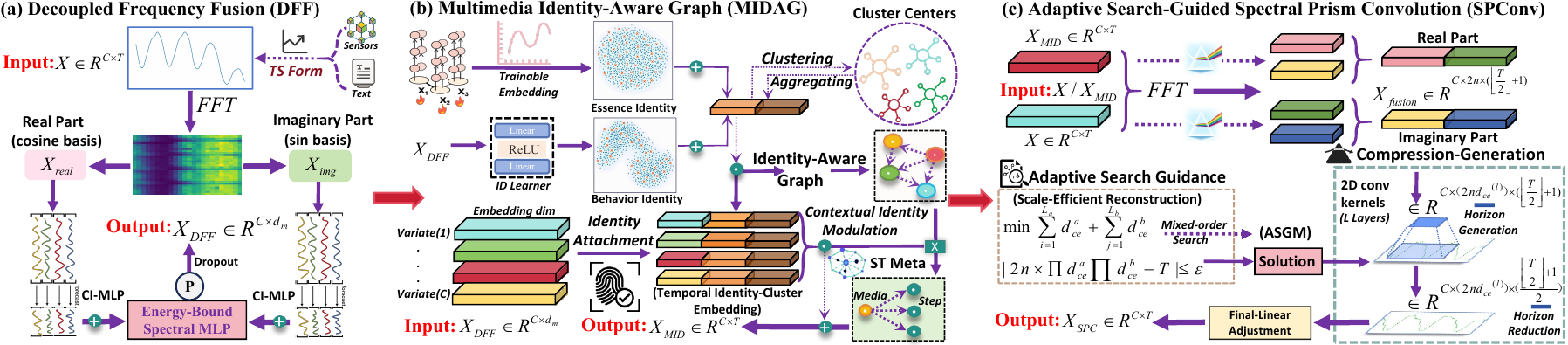}
\caption{Core TSEM modules: (a) Decoupled Frequency Fusion (DFF);
(b) Multimedia Identity-Aware Graph (MIDAG); and
(c) Spectral Prism Convolution (SPConv), whose architecture is configured
by Adaptive Search Guidance (ASGM). Here, $C$ and $T$ denote the numbers
of aligned variates and temporal steps, respectively. MIDAPN learns multivariate dependencies and multimodal heterogeneity within a unified backbone for Spatiotemporal decoupling}
\label{VTITG-ASFFN-model-2}
\end{figure*}

To harness identity-bearing media signals in sliding windows, we leverage the representation power of orthogonal bases \cite{freE,filterts,timebase} in the frequency domain.  As shown in Fig. \ref{VTITG-ASFFN-model-2} (a), we first employ the Fast Fourier Transform (FFT) to map the input signal $X \in \mathbb{R}^{C \times T}$ into the frequency domain, and then we decompose the spectrum into its orthogonal real and imaginary components, denoted as $\mathcal{X}_{\text{real}}(f)$ and $\mathcal{X}_{\text{imag}}(f) \in \mathbb{R}^{C \times (\lfloor \frac{T}{2} \rfloor + 1)}$ \cite{wu2025catch}, which are governed by cosine and sine basis functions respectively. Subsequently, we adopt a Channels-Independent MLP \cite{dlinear,patchtst} to prevent the cross-variable noise interference, ensuring their unique temporal distributions are represented as $\mathcal{Z}_{\text{real}}(f)$, $\mathcal{Z}_{\text{imag}}(f)$ $\in \mathbb{R}^{C \times T}$:
\begin{equation}
\resizebox{0.9\linewidth}{!}{
$\mathcal{Z}[C, T_{\text{out}}] = \sum\nolimits_{T_{\text{in}}} \mathcal{X}[C, T_{\text{in}}] \cdot \mathcal{W}_{\text{CI}}[C, T_{\text{in}}, T_{\text{out}}] + b_{\text{CI}}[C, T_{\text{out}}]$,
}
\end{equation}
\begin{equation}
\resizebox{0.9\linewidth}{!}{
$\mathcal{Z}_{\text{real}}(f), \mathcal{Z}_{\text{imag}}(f) = \text{CI-MLP}_{\text{real}}(\mathcal{X}_{\text{real}}(f)), \text{CI-MLP}_{\text{imag}}(\mathcal{X}_{\text{imag}}(f))$,
}
\end{equation}
where $\text{CI-MLP}(\cdot)$ contains independent weights and biases $\mathcal{W}_{\text{CI}} \in \mathbb{R}^{C \times (\lfloor \frac{T}{2} \rfloor + 1) \times T}$, $ b_{\text{CI}} \in \mathbb{R}^{C \times T}$ for each of the $C$ variates. This sequence-token embedding approach could better address intricate situations in real-world scenarios (e.g., temporal delays and distribution shifts) \cite{itransformer}. Furthermore, we design a straightforward yet effective Energy-Bounded Spectral MLP for additive fusion, which leverages Tanh for saturation damping and dropout for random inhibition: 
\begin{equation}
X_{\text{DFF}}=\text{Drop}(\phi_{\mathrm{Tanh}}(\text{Linear}_{\text{DFF}}(\mathcal{Z}_{\text{real}}(f) + \mathcal{Z}_{\text{imag}}(f)))).
\end{equation}
Note that $\text{Linear}_{\text{DFF}}(\cdot)$ layer is applied along the temporal dimension, mapping the input from $\mathbb{R}^{C \times T}$ to $X_{\text{DFF}} \in \mathbb{R}^{C \times d_{m}}$, here $d_{m}$ denotes model dimension for representation capacity. 

\subsection{$\Psi_{\text{spatial}}$: Multimedia Identity-Aware Graph}
\label{method-1-2}
\subsubsection{Design Rationale} To overcome indistinguishability \cite{shao2024exploring} in media-general graph learning, we revisit "identity" from three core aspects: the intrinsic properties of variates (\textbf{Essence Identity $\mathcal{E}_{\text{id}}$: \emph{what it is}} \cite{gwn,wu2020connecting}), context-dependent dynamics (\textbf{Behavior Identity $\mathcal{B}_{\text{id}}$: \emph{how it manifests}} \cite{megacrn,crossgnn,msgnet}), and shared commonalities (\textbf{Group-Level Identity $\mathcal{V}_{\text{group}}$: \emph{where it belongs}} \cite{DUET,STD-MAE,cluster}). Given the observations $\{\mathcal{O}\}$ in a mini-batch, we adopt a
conceptual probabilistic factorization: \textbf{\textit{how interdependent identities jointly induce}}
varied $\mathcal{G}^{(\mathrm{id})}_{\mathrm{aware}}\in\mathbb{R}^{C\times C}$ in sample-adaptive disentanglement
\begin{equation}
\resizebox{0.88\linewidth}{!}{
$\begin{split}
P\!\left(
\mathcal G_{\mathrm{aware}}^{(\mathrm{id})}\in\mathcal A
\right)
=
\sum\nolimits_{
X,\mathcal E_{\mathrm{id}},
\mathcal B_{\mathrm{id}},
\mathcal V_{\mathrm{group}}
}
P(X)
P(\mathcal E_{\mathrm{id}}\mid X)
P(\mathcal B_{\mathrm{id}}\mid X)
\\
P\!\left(
\mathcal V_{\mathrm{group}}
\mid
\mathcal E_{\mathrm{id}},
\mathcal B_{\mathrm{id}}
\right)
P\!\left(
\mathcal G_{\mathrm{aware}}^{(\mathrm{id})}\in\mathcal A
\mid
\mathcal E_{\mathrm{id}},
\mathcal B_{\mathrm{id}},
\mathcal V_{\mathrm{group}}
\right).
\end{split}$
}
\label{eq-4}
\end{equation}
where $\mathcal{A}$ is optimal set. $\mathcal{E}_{\text{id}}$ $\in \mathbb{R}^{C \times d_{\text{id}}}$, $\mathcal{B}_{\text{id}}$ $\in \mathbb{R}^{C \times d_{\text{id}}}$, and $\mathcal{V}_{\text{group}}$ $\in \mathbb{R}^{C \times d_{\text{cluster}}}$, where $d_{\text{id}}, d_{\text{cluster}}$ signify identity embedding and latent cluster centers dimension respectively. Meanwhile, we design contextual identity modulation to refine graph aggregation without explicitly estimating conditional mutual information (CMI)
\cite{explain}. Next, we will progressively derive \textit{Individual $\&$ Group-Level Identity}, \textit{Graph Construction}, and \textit{Implicit Contextual Identity Modulation} in MIDAG.

\subsubsection{Individual Identity Learning}
To encode the persistent properties of pre-aligned variates, we learn a trainable embedding $\mathcal{E}_{\text{id}}\in\mathbb{R}^{C\times d_{\text{id}}}$ \cite{gwn,wu2020connecting}. During training \cite{meta}, $\mathcal{E}_{\text{id}}$ essentially provides an observation-invariant anchor for distinction (i.e., sensor or TaTS-derived variates). For the behavior identity, we crafted a parameter-friendly "Dynamic ID learner", which is created to  extract instantaneous dynamics:
\begin{equation}
\resizebox{0.8\linewidth}{!}{
    $\mathcal{B}_{\text{id}} = \mathcal{W}_{d_{\text{id}}} \cdot \phi_{\mathrm{ReLU}}(\mathcal{W}_{d_{\text{m}}} X_{\mathrm{DFF}} + b_{d_{\text{m}}}) + b_{d_{\text{id}}}$,
}
\end{equation}
where "Dynamic ID learner" is implemented based on the low-rank hypothesis prevalent in robust learning \cite{yizhixing}. Instead of a direct linear mapping from $\mathbb{R}^{d_{m}} \rightarrow \mathbb{R}^{d_{\text{id}}}$, we factorize this transformation into two consecutive linear layers with weights $\mathcal{W}_{d_{m}}$ $\in \mathbb{R}^{d_{m} \times r}$, $\mathcal{W}_{{d_{\text{id}}}}$ $\in \mathbb{R}^{r \times d_{\text{id}}}$ and bias $b_{d_{\text{m}}}$ $\in \mathbb{R}^{r}$, $b_{d_{\text{id}}}$ $\in \mathbb{R}^{d_{\text{id}}}$. Afterward, we set the rank to a reduced rate $r$ of $\frac{d_{m}}{2}$ to ensure parameter efficiency while preserving adequate representational capacity. Next, we sum the static and dynamic Identities to obtain the individual variate identity $\mathcal{V}_{\text{id}} \in \mathbb{R}^{C \times d_{\text{id}}}$:
\begin{equation}
\mathcal{V}_{\mathrm{id}}(X)
=
\mathcal{E}_{\mathrm{id}}^{\star}(\mathcal{O})
+
\mathcal{B}_{\mathrm{id}}(X;\mathcal{O}),
\label{eq-6}
\end{equation}
where $\mathcal{E}_{\text{id}}$ shows prior expectation, while $\mathcal{B}_{\text{id}}$ refers to posterior update. Hence "residual-like" identity learning improves existing knowledge incrementally through dynamic feedback.

\subsubsection{Group-Level Identity Clustering}
Cluster centers summarize group commonalities
\cite{cluster}. We initialize latent centers
$\vartheta_{\mathrm{cluster}}
\in\mathbb{R}^{M\times d_{\mathrm{cluster}}}$,
where $M$ is the number of groups, and introduce a learnable projection
$\mathcal{W}_{\mathrm{i2c}}
\in\mathbb{R}^{d_{\mathrm{id}}\times d_{\mathrm{cluster}}}$
to align $\mathcal{V}_{\mathrm{id}}$ with
$\vartheta_{\mathrm{cluster}}$ in a shared semantic space.
Under an isotropic-covariance Gaussian-mixture interpretation, the soft-clustered identity is the posterior expectation
$\phi(\mathcal{V}_{\mathrm{id}})
=\sum_{m=1}^{M}
P(\mathrm{label}_{m}\mid\mathcal{V}_{\mathrm{id}})
\vartheta_{\mathrm{cluster}}^{m}$.
Accordingly, we treat projected identities as Queries and cluster
centers as Keys, and approximate posterior assignments by their
dot-product similarities $\mathcal{W}_{\text{cluster}} \in \mathbb{R}^{C \times M}$. This soft assignment allows each variate to retain graded membership across multiple latent groups, rather than being restricted to a single cluster. Moreover, we compute the $\mathcal{V}_{\text{group}} \in \mathbb{R}^{C \times d_{\text{cluster}}}$ for $\mathcal{V}_{\text{id}}$ by performing a superposition of all latent cluster centers:
\begin{equation}
    \mathcal{V}_{\text{group}} = \text{Softmax}(\mathcal{V}_{\text{id}} \mathcal{W}_{\text{i2c}} \vartheta^{T}_{\text{cluster}}) \vartheta_{\text{cluster}}.
\end{equation}
Finally, by concatenating $\mathcal{V}_{\text{id}}$, $\mathcal{V}_{\text{group}}$ at feature axis, we form the Identity-Cluster Embedding $\mathcal{Z}_{\text{id-cluster}} \in \mathbb{R}^{C \times (d_{\text{id}} + d_{\text{cluster}})}$.

\subsubsection{Multimedia Identity-Aware Graph Construction}
Conceptually, $\mathcal{Z}_{\text{id-cluster}}$ can be interpreted as a semantic decomposition of the monolithic node embeddings found in adaptive graph structure, where each segment is endowed with a distinct semantic meaning (i.e., individual \& group identity) \cite{hu2025timefilter}, enabling branch-wise
attribution \cite{freE}. By partitioning
$\mathcal{W}_{\text{gproj}}
=[\mathcal{W}_{\mathrm{I}}^{\top},
\mathcal{W}_{\mathrm{G}}^{\top}]^{\top}$, we have graph base $\mathcal{Z}_{\text{graph}} \in \mathbb{R}^{C \times d_{\text{graph}}}$:
\begin{equation}
\begin{aligned}
\mathcal{Z}_{\mathrm{graph}}
&=
\mathcal{V}_{\mathrm{id}}\mathcal{W}_{\mathrm{I}} + \mathcal{V}_{\mathrm{group}}\mathcal{W}_{\mathrm{G}},\\
\nabla_{\mathcal{W}_{s}}\mathcal{L}
&=
\mathcal{V}_{s}^{\top}
(\nabla_{\mathcal{Z}_{\mathrm{graph}}}\mathcal{L})_{s},
\;s\in\{\mathrm{I},\mathrm{G}\}.
\end{aligned}
\label{eq:branchwise_projection}
\end{equation}
Thus, each branch receives a dedicated gradient,
mitigating cross-semantic interference  (Sec. \ref{6.1.1}).
To retain positive inter-variate affinities and exclude self-loops, we
apply ReLU and a diagonal mask
$\mathcal{D}_{\text{mask}}\in\mathbb{R}^{C\times C}$.
Finally, Softmax yields
$\mathcal{G}^{(\text{id})}_{\text{aware}}\in\mathbb{R}^{C\times C}$ with row-wise probability distributions:
\begin{equation}
\mathcal{G}^{(\text{id})}_{\text{aware}}
=
\operatorname{Softmax}_{\mathrm{row}}\!\left(
\phi_{\mathrm{ReLU}}\!\left(
\mathcal{Z}_{\text{graph}}\mathcal{Z}_{\text{graph}}^{\top}
\right)+\mathcal{D}_{\text{mask}}
\right).
\end{equation}
Here Identity-induced affinity edge aligns individual \& group traits, enabling variate discrimination and cross-media linking: 
\begin{equation}
\resizebox{0.88\linewidth}{!}{$
\begin{aligned}
e^{(\mathrm{id})}_{ij}
&=
\Big(
[\mathcal{V}_{\mathrm{id},i}\Vert
 \mathcal{V}_{\mathrm{group},i}]
W_{\mathrm{gproj}}
\Big)
\Big(
[\mathcal{V}_{\mathrm{id},j}\Vert
 \mathcal{V}_{\mathrm{group},j}]
W_{\mathrm{gproj}}
\Big)^{\top} \\
&=
\mathcal{V}_{\mathrm{id},i}M_{\mathrm{II}}
\mathcal{V}_{\mathrm{id},j}^{\top}
+
\mathcal{V}_{\mathrm{id},i}M_{\mathrm{IG}}
\mathcal{V}_{\mathrm{group},j}^{\top} \\
&\quad+
\mathcal{V}_{\mathrm{group},i}M_{\mathrm{GI}}
\mathcal{V}_{\mathrm{id},j}^{\top}
+
\mathcal{V}_{\mathrm{group},i}M_{\mathrm{GG}}
\mathcal{V}_{\mathrm{group},j}^{\top},
\end{aligned}
$}
\label{eq:affinity_decomposition}
\end{equation}
As \(C\) contains TS variates in multivariate but also projected-text variates under TaTS, the same kernel could capture TS--TS, TS--text, and text--text relations. For affinity decomposition, $M_{\mathrm{II}}$ and $M_{\mathrm{GG}}$ preserve individual specificity and group commonality, while $M_{\mathrm{IG/GI}}$ bridge them, allowing edges to serve variable-specific and cross-media relations (Sec. \ref{AFFD}).

\subsubsection{Contextual Identity Modulation (CIM)} 
\begin{figure}
\centering
\includegraphics[width=1.0 \linewidth]{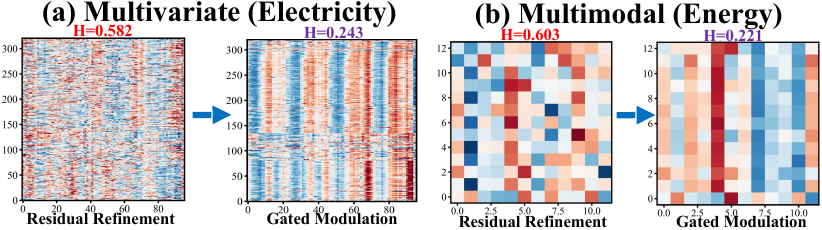}
\caption{Normalized SVD-energy entropy of CIM refinement ($C\times T$); Lower entropy values indicate stronger spectral mode concentration.}
\label{figure-pattern}
\end{figure}

We implicitly realize identity-aware aggregation modulation via \textit{temporal context transferring} and \textit{residual refinement}. For preparation, similar to ST-identity attachment \cite{STID}, we combine the DFF output $\mathbf{X}_{\mathrm{DFF}}$ with individual- and group-level to propagate both identity cues and temporal information. In this way, we get Temporal-Identity-Cluster Embedding $\mathcal{Z}_{\text{t-id-cluster}} \in \mathbb{R}^{C \times (d_{m} + d_{\text{id}} + d_{\text{cluster}})}$ in Fig.  \ref{VTITG-ASFFN-model-2}(b):
\begin{equation}
    \mathcal{Z}_{\text{t-id-cluster}} = [X_{\text{DFF}} \mid \mid \mathcal{V}_{\text{id}} \mid \mid \mathcal{V}_{\text{group}}]. 
\end{equation}
The next stage aims to dynamically infuse multimedia dependencies from the $G^{(\text{id})}_{\text{aware}}$ into the temporal context. For this purpose, we introduce a contextual meta-parameter $\mathcal{W}_{\text{context}} \in \mathbb{R}^{C \times D}$ ($D = d_{m} + d_{\text{id}} + d_{\text{cluster}}$), which acts as a learnable prototype library \cite{megacrn}, storing contextual information to fully leverage the indivisible spatio-temporal heterogeneity \cite{meta}. For infusion, we first perform the discrete Identity-Aware topology transformation of $\mathcal{Z}_{\text{t-id-cluster}}$ over $G^{(\text{id})}_{\text{aware}}$ to approximate probabilistic integration $\int_{\mathcal{G}} \mathcal{Z}  d P(\mathcal{G})$, obtaining an aggregated representation $\mathcal{Z}_{\text{agg}} \in \mathbb{R}^{C \times D}$. Next, a Query-Key computation is utilized for interaction between $\mathcal{Z}_{\text{agg}}$ and $\mathcal{W}_{\text{context}}$:
\begin{equation}
\resizebox{0.88\linewidth}{!}{$
\displaystyle
\mathcal{Z}_{\mathrm{agg}}
=
G_{\mathrm{aware}}^{(\mathrm{id})}
\mathcal{Z}_{\mathrm{t\text{-}id\text{-}cluster}},
\quad
\mathcal{W}_{\mathrm{aware}}^{(\mathrm{id})}
=
\phi_{\mathrm{Tanh}}\!\left(
\mathcal{Z}_{\mathrm{agg}}^{\top}
\mathcal{W}_{\mathrm{context}}
\right).
$}
\label{eq-combined}
\end{equation}
$\mathcal{Z}_{\text{t-id-cluster}}$ is used as a query to perform a self-gated modulation against the $\mathcal{W}^{(\text{id})}_{\text{aware}} \in \mathbb{R}^{D \times D}$ for identity-awareness. Finally, token mixing and residual refinement are integrated using $I_D$:
\begin{equation}
\resizebox{0.88\linewidth}{!}{$
\displaystyle
\mathcal{X}_{\mathrm{MID}}
=
\operatorname{Linear}_{\mathrm{MID}}\!\left(
\mathcal{Z}_{\mathrm{t\text{-}id\text{-}cluster}}
\left[
\operatorname{Dropout}\!\left(
\mathcal{W}_{\mathrm{aware}}^{(\mathrm{id})}
\right)
+
\mathcal{I}_{D}
\right]
\right).
$}
\label{eq-15}
\end{equation}
\textit{Interpretation from CMI:} Rather than using biased auxiliary graph losses \cite{otc}, we interpret the joint modulation under the
aggregated context
$\mathcal{C}\triangleq\mathcal{Z}_{\text{agg}}$. Let
$\mathcal{R}\triangleq\mathcal{Z}I_D$
and
$\mathcal{M}\triangleq
\mathcal{Z}
\mathcal{W}^{(\text{id})}_{\text{aware}}$
denote the ST-residual and gated responses:
\begin{equation}
\resizebox{0.89\linewidth}{!}{$I\!\left(X_{\text{MID}};\mathcal{Y}\mid\mathcal{C}\right)\leq I\!\left([\mathcal{R},\mathcal{M}];\mathcal{Y}\mid\mathcal{C}\right)=I\!\left(\mathcal{R};\mathcal{Y}\mid\mathcal{C}\right)+I\!\left(\mathcal{M};\mathcal{Y}\mid\mathcal{R},\mathcal{C}\right).$}
\label{eq14}
\end{equation}
The CMI chain rule decomposes the information flow into two distinct pathways: (1) ST-identity cues retained via residual refinement, and (2) graph-conditioned responses modulated by identity-aware gating. Although $\mathcal{M}$ varies across samples due to the input-dependent context $\mathcal{C}$, it is deterministically governed by $\mathcal{R}$ and $\mathcal{C}$ under fixed parameters. Consequently, Eq.~(\ref{eq14}) characterizes the architectural processing mechanism rather than the injection of additional conditional information. Empirically, as visualized via SVD-energy entropy in Figure~\ref{figure-pattern}, residual refinement preserves a broader spectrum of ST modes, whereas gating sharply concentrates on dominant modes. Coupled with the ``w/o CIM'' ablation in Sec.~\ref{arr}, these findings firmly substantiate the efficacy of context-guided refinement.

\subsection{$\Psi_{\text{temporal}}$: "Search-Guided" Spectral Prism}
\subsubsection{Spectrum Capturing}
Informed by Parseval's theorem \cite{fremlp} on global time-frequency 
energy equivalence, we adopt the working hypothesis that \textit{potentially 
salient temporal patterns can be revealed through compact, localized 
spectral responses}. The energy within a spectral interval 
$[j+1,j+S]$ can be characterized by summing the squared spectral 
magnitudes $\{|A_k|^2\}$. Rather than explicitly constructing the 
power spectrum, we apply a learnable convolutional kernel to extract local responses from the spectral coefficients:
\begin{equation}
\resizebox{0.85\linewidth}{!}{$
E_j=\sum_{k=j+1}^{j+S}|A_k|^2,\quad
\widehat{R}_j=
\sigma\left(
\sum_{i=1}^{S}W_i^{(\mathrm{conv})}A_{j+i-1}
+\mathrm{bias}
\right).
$}
\end{equation}
Following this spectral analogy \cite{prism}, multi-source signals
$X,X_{\mathrm{MID}}\in\mathbb{R}^{C\times T}$ are mapped into frequency bases,
split into real \& imaginary part and concatenated as
$X_{\mathrm{fusion}}\in\mathbb{R}^{C\times2n\times(\lfloor T/2\rfloor+1)}$
($n=2$). Since spectral resolution varies with $T$, a fixed architecture
may not adapt gracefully across horizons. \textbf{This scale mismatch naturally needs
automatic configuration to balance trends and details}.

\subsubsection{Hierarchical Spectral Scaling}
Instead of separate encoders and decoders for multiscale modeling
\cite{umixer,SCTNet,Jia_Song_Ye_Yuan_2026}, we couple compression-generation hierarchically
via average pooling and channel-expanded convolution: across layers, spectral positions contract as
$T^{(l)}\!\approx\!k_{\mathrm{pool}}^{-l}T^{(0)}$, while channel
width grows as $D^{(l)}\!=\!2n\prod_{r=1}^{l}d_{\mathrm{ce}}^{(r)}$.
Hence,
$D^{(l)}T^{(l)}\!\approx\!\rho^{(l)}
D^{(l-1)}T^{(l-1)}$, $\rho^{(l)}\!=\!d_{\mathrm{ce}}^{(l)}/k_{\mathrm{pool}}$ characterizes the balance between coarse-trend abstraction and fine-detail
preservation:
\begin{equation}
\resizebox{0.89 \linewidth}{!}{
 $X^{(l)}_{\text{fusion}} = \text{AvgPool}(\sum\nolimits_{2n \cdot \prod_{l-1} d^{(l-1)}_{\text{ce}}}\sum\nolimits_{k} W^{(l)}_{\text{conv2d}} (\text{Pad}(X^{(l-1)}_{\text{fusion}})))$.
}
\end{equation}
\noindent where $W^{(l)}_{\text{conv2d}} \in \mathbb{R}^{2n \cdot \prod_{r=1}^{l-1}d^{(r)}_{\text{ce}} \times 2n \cdot \prod_{r=1}^{l}d^{(r)}_{\text{ce}} \times 1 \times k}$ uses $d_{\text{ce}}^{(l)}$ to offset pooling compression. With $k_{\text{pool}}=2, k=3, p_d=2$, convolution preserves length pre-pooling, yielding $T^{(l)}=\left\lfloor \mathcal{G}(T^{(l-1)})/k_{\text{pool}}\right\rfloor$, $\mathcal{G}(T)=T+p_d-k+1$. From hierarchical scaling, the decreasing $T^{(l)}$ ($\approx 2^{-l}$) and expanding $D^{(l)}$ unfold multi-scale frequencies along the new axis for reconstruction. In layer $L$, $\text{Linear}_{\text{SPC}}(\cdot)$ adjusts the $T$ dimension and fuses $X^{(L)}_{\text{fusion}}, X, X_{\text{MID}}$ for $X_{\text{SPC}} \in \mathbb{R}^{C \times T}$. Each layer therefore trades spectral resolution for channel capacity,
turning temporal scale into an explicit architectural progression.

\subsubsection{"Adaptive Search-Guided" Architecture Configuration}
ASGM is not Neural Architecture Search (NAS), it involves no
"search-then-train" candidate evaluation \cite{NAS,Jia_2026}.
Instead, it deterministically instantiates a single architecture at
initialization, allocating layer-wise expansion factors to meet the
reconstruction target, termed "search-by-construction". 

For input horizon $T_{\text{in}}$, ASGM determines core parameters:

\textbf{(1) Network Depth $L$:} The hierarchical convolution-pooling for layer is : $L = \lfloor \frac{\text{log}(\lfloor \frac{T_{\text{in}}}{2} \rfloor + 1)}{\text{log}(k_{\text{pool}})} \rfloor$, which compresses the input temporal resolutions through multiple downsampling. 

\textbf{(2) Expansion Factor $d_{\text{ce}}$:} The unique channel expansion factor $d_{\text{ce}}^{(l)}$ is devised to progressively reconstruct multimedia representations via the newly created multi-sources dimension.

Regarding optimization strategies for expansion factors, fundamental approaches include maximization, minimization, and random selection. However, maximization tends to result in extreme factor distributions (e.g., multiple layers initially set to 1 but eventually converging to $T$), while random selection from the candidate pool $\mathcal{Q}=\{d^{(1)}_{\text{ce}},d^{(2)}_{\text{ce}},...,d^{(L)}_{\text{ce}}\}$ cannot guarantee optimal architecture or robustness. 

For scale-efficient optimization, selecting an expansion strategy is
formulated as a constrained integer program \cite{opt}. We seek an
integer allocation
$\mathcal{Q}=(d_{\mathrm{ce}}^{(1)},\ldots,d_{\mathrm{ce}}^{(L)})$
that minimizes the reconstruction-capacity cost
$\mathcal{C}(\mathcal{Q})$. These factors determine the channel
dimensions of each mapping
$K^{(l)}\in\mathbb{R}^{
2n\prod_{r=1}^{l-1}d_{\mathrm{ce}}^{(r)}
\times
2n\prod_{r=1}^{l}d_{\mathrm{ce}}^{(r)}}$,
and hence bound its attainable rank capacity \cite{timebase}.
Since the learned effective ranks are unavailable before training
\cite{low2}, $\mathcal{C}(\mathcal{Q})$ serves as a structural proxy
that promotes balanced capacity growth. Meanwhile, the scaled cumulative
expansion $2n\prod_{l=1}^{L}d_{\mathrm{ce}}^{(l)}$ should ideally
match the reconstructed temporal dimension $T_{\mathrm{rec}}$,
ensuring sufficient reconstruction capacity without redundant
channel inflation:
\begin{equation}
\resizebox{0.88\linewidth}{!}{
$\textbf{Min}\quad \mathcal{C}(\mathcal{Q})
= \sum_{l=1}^{L} d_{\text{ce}}^{(l)}, \qquad
\textbf{s.t.}\quad
\left\lvert 2n \cdot \prod_{l=1}^{L} d_{\text{ce}}^{(l)}
- T_{\text{rec}} \right\rvert \leq \varepsilon$,
}
\label{eq-23}
\end{equation}
where $\varepsilon$ is an idealized tolerance. Leveraging the Arithmetic-Geometric Mean Inequality $\frac{\sum_{l=1}^{L} d_{\text{ce}}^{(l)}}{L} \geq \left( \prod_{l=1}^{L} d_{\text{ce}}^{(l)} \right)^{\frac{1}{L}}$, a balanced reference is $d_{\text{ce}}^{(l)} = (\frac{T_{\text{rec}}}{2n})^{\frac{1}{L}}$. However, this uniform expansion restricts layer-wise capacity flexibility, while strict capacity matching may also be infeasible under integer constraints. We thus adopt a capacity-guided mixed-order search.

\subsubsection{Mixed-order Allocations} 
Given candidate expansion factors $\mathcal{Q}_{d}\subset\mathbb{Z}^{+}$,
exhaustive $L$-layer search incurs $\mathcal{O}(|\mathcal{Q}_{d}|^{L})$
complexity. We thus restrict allocations to at most two distinct factors.
With $d_{\mathrm{ce}}^{a}=\lfloor(T_{\mathrm{rec}}/(2n))^{1/L}\rfloor$,
each split $m\in\{1,\ldots,L-1\}$ defines
$\mathcal{Q}(m,d_{\mathrm{ce}}^{b}) =
[(d_{\mathrm{ce}}^{a})^{\times m},
(d_{\mathrm{ce}}^{b})^{\times(L-m)}]$:
\begin{equation}
\resizebox{0.89\linewidth}{!}{%
$\begin{aligned}
&\textbf{Min}_{m}\quad
\mathcal{C}(m,d_{\mathrm{ce}}^{b})
=
m d_{\mathrm{ce}}^{a}
+
(L-m)d_{\mathrm{ce}}^{b},\\
&\textbf{s.t.}\quad
2n(d_{\mathrm{ce}}^{a})^{m}
(d_{\mathrm{ce}}^{b})^{L-m}
\leq T_{\mathrm{rec}}
<
2n(d_{\mathrm{ce}}^{a})^{m}
(d_{\mathrm{ce}}^{b})^{L-m+1}.
\end{aligned}$%
}
\label{eq:mixed_order_constraint}
\end{equation}
For each fixed $m$, let $j=L-m$ and
$A_m=T_{\mathrm{rec}}/[2n(d_{\mathrm{ce}}^{a})^m]$.
The capacity-bracketing condition requires
$(d_{\mathrm{ce}}^{b})^j\leq A_m<
(d_{\mathrm{ce}}^{b})^{j+1}$.
Accordingly, we retain
$d_{\mathrm{ce}}^{b}=\lfloor A_m^{1/j}\rfloor$
if feasible. The corresponding cost
$\mathcal{C}_{m}(d)
=m d_{\mathrm{ce}}^{a}+(L-m)d$ satisfies
\begin{equation}
\mathcal{C}_{m}(d+1)-\mathcal{C}_{m}(d)
=
L-m
>
0.
\label{eq:mixed_order_monotonicity}
\end{equation}
Although $\mathcal{C}_{m}(d)$ increases with $d$,
the floor-based construction selects the largest integer
$d_{\mathrm{ce}}^{b}$ that does not overshoot the target for each $m$.
We compare the valid candidates across splits,
selecting the one with the lowest total expansion cost.
This is a greedy choice among constructed
solutions
(Algorithm~\ref{alg:multi_order}).

\begin{algorithm}[t]
\caption{Greedy Search for Mixed-order Strategy}
\label{alg:multi_order}
\begin{algorithmic}[1]
\Require $T_{\mathrm{in}}, T_{\mathrm{rec}}\geq 8$,
$d_{\mathrm{ce}}^{a}$, $L$, $2n$
\Ensure $\mathcal{Q}_{\mathrm{best}}^{*}$
\Comment{Selected expansion strategy}

\State $(\mathcal{Q}_{\mathrm{best}}^{*},c^{*})
\gets([d_{\mathrm{ce}}^{a}]^{\times L},\infty)$
\Comment{Basic initialization}

\For{$i=1,\ldots,L-1$}
\Comment{$L=\left\lfloor
\frac{\log(\lfloor T_{\mathrm{in}}/2\rfloor+1)}
{\log(k_{\mathrm{pool}})}
\right\rfloor$}

\State $j\gets L-i,\quad
A_i\gets \frac{T_{\mathrm{rec}}}
{2n(d_{\mathrm{ce}}^{a})^i}$, \quad $d_{\mathrm{ce}}^{b}\gets
\lfloor A_i^{1/j}\rfloor,\quad
c_i\gets i d_{\mathrm{ce}}^{a}
+j d_{\mathrm{ce}}^{b}$
\Comment{Candidate and expansion cost}

\If{$c_i<c^{*}\land d_{\mathrm{ce}}^{b}>1
\land (d_{\mathrm{ce}}^{b})^j
\leq A_i<(d_{\mathrm{ce}}^{b})^{j+1}$}

\State $(\mathcal{Q}_{\mathrm{best}}^{*},c^{*})
\gets(\mathcal{Q}(i,d_{\mathrm{ce}}^{b}),c_i)$
\Comment{Lowest-cost}
\EndIf
\EndFor

\State $\mathcal{Q}_{\mathrm{best}}^{*}
\gets\operatorname{Interleave}(\mathcal{Q}_{\mathrm{best}}^{*})$
\end{algorithmic}
\end{algorithm}

\begin{table}[htbp]
  \centering
  \caption{Forecasting comparison of different expansion-factor
  strategies. The best results are highlighted in \textbf{bold}.}
  \label{tab:expansion_strategy}
  \resizebox{\columnwidth}{!}{
    \begin{tabular}{lccccccc}
      \toprule
      \multirow{2}{*}{\textbf{Dataset}}
      & \multirow{2}{*}{\textbf{Horizon}}
      & \multicolumn{2}{c}{\textbf{Max-Sum}}
      & \multicolumn{2}{c}{\textbf{Random}}
      & \multicolumn{2}{c}{\textbf{Min-Sum}} \\
      \cmidrule(lr){3-4}
      \cmidrule(lr){5-6}
      \cmidrule(lr){7-8}
      & & MSE & MAE & MSE & MAE & MSE & MAE \\
      \midrule
      Flight
      & 96
      & \textbf{0.144} & \textbf{0.248}
      & 0.145 & 0.249
      & \textbf{0.144} & \textbf{0.248} \\

      Weather
      & 96
      & 0.147 & 0.188
      & \textbf{0.145} & 0.186
      & \textbf{0.145} & \textbf{0.185} \\

      ETTm2
      & 96
      & 0.171 & 0.252
      & 0.172 & 0.253
      & \textbf{0.168} & \textbf{0.249} \\
      \midrule
      Flight
      & 720
      & 0.197 & 0.294
      & 0.196 & 0.292
      & \textbf{0.193} & \textbf{0.290} \\

      ETTm2
      & 720
      & 0.396 & 0.396
      & 0.394 & 0.393
      & \textbf{0.393} & \textbf{0.391} \\

      Electricity
      & 720
      & 0.202 & 0.292
      & 0.201 & 0.293
      & \textbf{0.194} & \textbf{0.287} \\
      \bottomrule
    \end{tabular}
  }
\end{table}

\begin{figure}
\centering
\includegraphics[width=1.0 \linewidth]{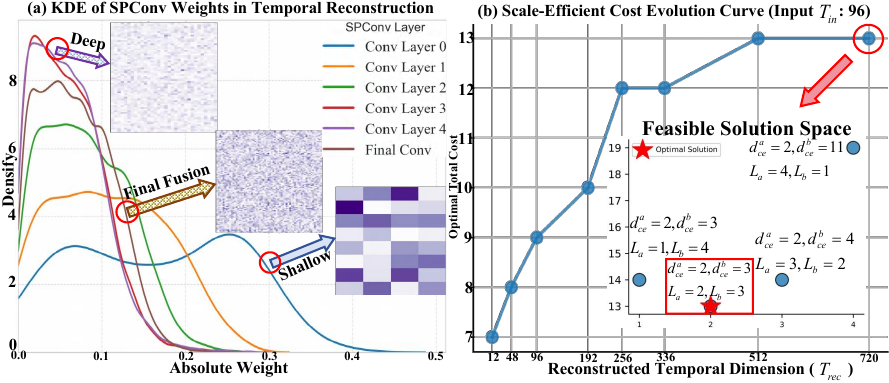}
\caption{"Adaptive Search-Guided" SPConv configuration (Left: multi-scale distributions; Right: $d_{\mathrm{ce}}$ search, cost evolution, and selected low-cost solutions), illustrating scale progression and compact architecture configuration}
\label{figure-pattern-vis}
\end{figure}

For computational analysis, each split $m$ only evaluates
one floor-based candidate $d_{\mathrm{ce}}^{b}$ in $\mathcal{Q}_{d}$. Since there are
$L-1$ possible values of $m$, the search complexity is
$\mathcal{O}(L)=\mathcal{O}(\log T)$ and remains efficient. Finally, the two selected factors are interleaved across SPConv for balanced distribution, avoiding the adjacent layers’ drastic fluctuation.

Overall, in SPConv, $L$ expands abstraction, while $\mathcal{Q}^{*}_{\mathrm{best}}$ allocates capacity efficiently. Table \ref{tab:expansion_strategy} demonstrates that this minimum proxy outperforms alternatives by preventing the representational redundancy of unbalanced allocations in long-horizon forecasting (e.g., $T=720$).

\subsubsection{Patterns Distribution and Scale-Efficient Architecture Searching} From Fig. \ref{figure-pattern-vis} subplots (a), the adoption of mixed-order expansion reveals the internal multi-scales from shallow to deep layers. For temporal reconstruction, with increasing dimensions in Fig. \ref{figure-pattern-vis} subplots (b), ASGM cost rises stepwise while remaining within a controllable horizon range. This indicates ASGM can robustly identify a compact architecture under cost minimization. Within the feasible space, a more balanced and less redundant reconstruction path was selected.

\subsection{MIDAPN Formulation and Forward Propagation}
\label{method-3}
\subsubsection{Instance Normalization and Predict Residuals} To balance the stationarity and smoothness of forecasted multimedia time series, we optionally or jointly employ the Reversible Instance Normalization (RevIN) \cite{revin} and Predict Residuals Regularization (PRReg) \cite{yizhixing} based on domain-specific nature.
\subsubsection{Historical Time Series Embedding} learning historical spatio-temporal correlations and variations for $X_{\text{his}} \in \mathbb{R}^{C \times T_{h}}$:
\begin{equation}
X_{\text{his}} = \text{TSEM}_{\text{his}}\left(\text{RevIN}_{\text{forward}}(X_{1:T_{h}} - X_{T_{h}}\right)),
\end{equation}
\subsubsection{Prediction} The historical embedding $X_{\text{his}}$ is projected to the future horizons through Channels-Independent MLP:
\begin{equation}
Y = \text{CI-MLP}\left(X_{\text{his}}\right), \quad Y \in \mathbb{R}^{C \times T_{f}}.
\end{equation}
\subsubsection{Future Time Series Refinement} The initial prediction $Y \in \mathbb{R}^{C \times T_{f}}$  undergoes the parameter-independent TSEM architecture to produce the final forecasted $Y_{\text{pred}} \in \mathbb{R}^{C \times T_{f}}$:
\begin{equation}
Y_{\text{pred}} = \text{RevIN}_{\text{reverse}}(\text{TSEM}_{\text{pred}}\left(Y\right) + X_{T_{h}}).
\end{equation}
\subsubsection{Operator-Level Interpretation} Essentially, the forward computation process of MIDAPN aligns with the principles of classical embedding theory \cite{embedding}. The $\Psi_{\text{TSEM}_{\text{his}}}(\cdot)$ module first estimates the dynamical state of the system from observation set $\{\mathcal{O}\}$, then the $\mathcal{F}_{\text{CI-MLP}}(\cdot)$ directly evolves it into the predicted state, and the $\Psi_{\text{TSEM}_{\text{pred}}}(\cdot)$ ensures the embedding validity. Thereby, the prediction task can be reformulated as:
\begin{equation}
    Y_{\text{pred}} = \Psi_{\text{TSEM}_{\text{pred}}} \circ \mathcal{F}_{\text{CI-MLP}} \circ \Psi_{\text{TSEM}_{\text{his}}}(X).
\end{equation}
\subsubsection{Tailored Training Loss} We use robust MAE for multivariate forecasting to stabilize fitting across many
variates, whereas MSE is adopted for multimodal forecasting to place greater
emphasis on pointwise deviations and local temporal changes.
Let $z\in\{0,1\}$ denote the fixed task indicator:
\begin{equation}
\mathcal{L}
=
(1-z)\mathcal{L}_{\mathrm{MAE}}
+
z\mathcal{L}_{\mathrm{MSE}}
+
\beta\|\theta\|_2^2,
\end{equation}
where $z$=0 is multivariate, and $z$=1 is multimodal. $\mathcal{L}_{\text{MAE}}$=$\frac{1}{c \times t} \sum\nolimits_{c}\sum\nolimits_{t}|Y_{\text{pred}}-Y_{\text{true}}|$ and $\mathcal{L}_{\text{MSE}}$=$\frac{1}{c \times t}\sum\nolimits_{c}\sum\nolimits_{t}|Y_{\text{pred}}-Y_{\text{true}}|^{2}$, where $Y_{\text{true}}$ is the ground truth of predicted time series. $\beta$ is the optional weight decay coefficient for non-overfitting.

\begin{table}[ht]
\centering
\caption{Statistics of Datasets and the Actual Use of RevIN/PRReg for MIDAPN. All time series are standardized}
\label{tab:dataset_stats_wide}
\resizebox{1.0\linewidth}{!}{
\begin{tabular}{lccccc}
\toprule[1.5pt]
\textbf{Datasets}
& \textbf{Variates}
& \textbf{RevIN}
& \textbf{PRReg}
& \textbf{Timestamps}
& \textbf{Frequency} \\
\midrule

ETTm2
& 7
& $\boldsymbol{\times}$
& $\boldsymbol{\checkmark}$
& 69680
& 15mins \\

ETTh2
& 7
& $\boldsymbol{\times}$
& $\boldsymbol{\checkmark}$
& 17420
& 1h \\

Flight
& 7
& $\boldsymbol{\checkmark}$
& $\boldsymbol{\times}$
& 26304
& 1h \\

Weather
& 21
& $\boldsymbol{\checkmark}$
& $\boldsymbol{\times}$
& 52696
& 10mins \\

Traffic
& 862
& $\boldsymbol{\checkmark}$
& $\boldsymbol{\times}$
& 17544
& 1h \\

Electricity
& 321
& $\boldsymbol{\checkmark}$
& $\boldsymbol{\times}$
& 26304
& 1h \\

Solar
& 137
& $\boldsymbol{\times}$
& $\boldsymbol{\times}$
& 35040
& 15mins \\

PEMS03
& 358
& $\boldsymbol{\times}$
& $\boldsymbol{\times}$
& 26208
& 5mins \\

PEMS04
& 307
& $\boldsymbol{\times}$
& $\boldsymbol{\times}$
& 16992
& 5mins \\

PEMS07
& 883
& $\boldsymbol{\times}$
& $\boldsymbol{\times}$
& 28224
& 5mins \\

PEMS08
& 170
& $\boldsymbol{\times}$
& $\boldsymbol{\times}$
& 17856
& 5mins \\

ILI
& 7
& $\boldsymbol{\checkmark}$
& $\boldsymbol{\checkmark}$
& 966
& 1week \\

NASDAQ
& 12
& $\boldsymbol{\checkmark}$
& $\boldsymbol{\checkmark}$
& 3914
& 1day \\

\midrule

Agriculture
& \multirow{9}{3cm}{\centering 1 (time series) \\ +12 (text dim)}
& $\boldsymbol{\checkmark}$
& $\boldsymbol{\checkmark}$
& 496
& Monthly \\

Climate
&
& $\boldsymbol{\checkmark}$
& $\boldsymbol{\checkmark}$
& 496
& Monthly \\

Economy
&
& $\boldsymbol{\times}$
& $\boldsymbol{\checkmark}$
& 423
& Monthly \\

Energy
&
& $\boldsymbol{\checkmark}$
& $\boldsymbol{\times}$
& 1479
& Weekly \\

Environment
&
& $\boldsymbol{\checkmark}$
& $\boldsymbol{\checkmark}$
& 11102
& Daily \\

Health
&
& $\boldsymbol{\times}$
& $\boldsymbol{\checkmark}$
& 1389
& Weekly \\

Security
&
& $\boldsymbol{\checkmark}$
& $\boldsymbol{\times}$
& 297
& Monthly \\

SocialGood
&
& $\boldsymbol{\times}$
& $\boldsymbol{\checkmark}$
& 900
& Monthly \\

Traffic
&
& $\boldsymbol{\checkmark}$
& $\boldsymbol{\checkmark}$
& 531
& Monthly \\

\midrule

MSPG
& 27 (time series) +12 (text dim)
& $\boldsymbol{\times}$
& $\boldsymbol{\checkmark}$
& 37536
& 15mins \\

LEU
& 16 (time series) +12 (text dim)
& $\boldsymbol{\times}$
& $\boldsymbol{\checkmark}$
& 34848
& 30mins \\

PTF
& 32 (time series) +12 (text dim)
& $\boldsymbol{\times}$
& $\boldsymbol{\checkmark}$
& 8640
& 1h \\

\bottomrule[1.5pt]
\end{tabular}
}
\end{table}

\begin{table*}[!htbp]
\centering
\caption{Averaged Comparison across 25 ``multivariate'' and ``multimodal'' datasets in multiple future horizons. \best{Red} indicates best, while \secondbest{blue} indicates second. Prefixes: [MV] Multivariate, [MM] Multimodal. For fairness: (1) In multivariate, 96 predict \{12, 24, 48, 96\} length for PEMS, 36 predict \{24, 36, 48, 60\} length for ILI and NASDAQ, other datasets use 96 predict \{96, 192, 336, 720\} in long-terms. (2) In multimodal, we use 8 predict \{8, 10, 12\} length for unified-granularity evaluation within TaTS \cite{li2025language}}
\label{tab:comprehensive_merged_categorized}
\resizebox{1.0\linewidth}{!}{%
\begin{tabular}{@{}c *{17}{c c}@{}}
\toprule[1.5pt]

\multicolumn{1}{c}{Models}
& \multicolumn{2}{c}{MIDAPN}
& \multicolumn{2}{c}{TimeFilter}
& \multicolumn{2}{c}{VPNet}
& \multicolumn{2}{c}{CPiRi}
& \multicolumn{2}{c}{SEMixer}
& \multicolumn{2}{c}{DUET}
& \multicolumn{2}{c}{TimeKAN}
& \multicolumn{2}{c}{FilterTS}
& \multicolumn{2}{c}{TimePro}
& \multicolumn{2}{c}{P-sLSTM}
& \multicolumn{2}{c}{ModernTCN}
& \multicolumn{2}{c}{TimeMixer}
& \multicolumn{2}{c}{iTransformer}
& \multicolumn{2}{c}{Time-LLM}
& \multicolumn{2}{c}{MSGNet}
& \multicolumn{2}{c}{PatchTST}
& \multicolumn{2}{c}{TimesNet} \\

\multicolumn{1}{c}{(Years)}
& \multicolumn{2}{c}{(Ours)}
& \multicolumn{2}{c}{(ICML'25)}
& \multicolumn{2}{c}{(ICLR'26)}
& \multicolumn{2}{c}{(ICLR'26)}
& \multicolumn{2}{c}{(WWW'26)}
& \multicolumn{2}{c}{(KDD'25)}
& \multicolumn{2}{c}{(ICLR'25)}
& \multicolumn{2}{c}{(AAAI'25)}
& \multicolumn{2}{c}{(ICML'25)}
& \multicolumn{2}{c}{(AAAI'25)}
& \multicolumn{2}{c}{(ICLR'24)}
& \multicolumn{2}{c}{(ICLR'24)}
& \multicolumn{2}{c}{(ICLR'24)}
& \multicolumn{2}{c}{(ICLR'24)}
& \multicolumn{2}{c}{(AAAI'24)}
& \multicolumn{2}{c}{(ICLR'23)}
& \multicolumn{2}{c}{(ICLR'23)} \\

\cmidrule(lr){2-3}
\cmidrule(lr){4-5}
\cmidrule(lr){6-7}
\cmidrule(lr){8-9}
\cmidrule(lr){10-11}
\cmidrule(lr){12-13}
\cmidrule(lr){14-15}
\cmidrule(lr){16-17}
\cmidrule(lr){18-19}
\cmidrule(lr){20-21}
\cmidrule(lr){22-23}
\cmidrule(lr){24-25}
\cmidrule(lr){26-27}
\cmidrule(lr){28-29}
\cmidrule(lr){30-31}
\cmidrule(lr){32-33}
\cmidrule(lr){34-35}

\multicolumn{1}{c}{Metric}
& MSE & MAE
& MSE & MAE
& MSE & MAE
& MSE & MAE
& MSE & MAE
& MSE & MAE
& MSE & MAE
& MSE & MAE
& MSE & MAE
& MSE & MAE
& MSE & MAE
& MSE & MAE
& MSE & MAE
& MSE & MAE
& MSE & MAE
& MSE & MAE
& MSE & MAE \\
\midrule

% ==================== MV ====================
ETT (Avg) & \best{0.318} & \best{0.353} & \best{0.318} & \secondbest{0.359} & 0.334 & 0.365 & 0.332 & 0.367 & \best{0.318} & 0.363 & 0.326 & 0.361 & 0.330 & 0.364 & \secondbest{0.324} & \secondbest{0.359} & 0.329 & 0.365 & 0.336 & 0.371 & 0.329 & 0.361 & 0.329 & 0.365 & 0.336 & 0.369 & 0.332 & 0.366 & 0.342 & 0.373 & 0.334 & 0.366 & 0.352 & 0.380 \\
Flight & \secondbest{0.161} & \best{0.262} & \secondbest{0.161} & 0.268 & \secondbest{0.161} & 0.269 & \best{0.160} & 0.268 & \best{0.160} & \secondbest{0.267} & 0.163 & 0.271 & 0.179 & 0.289 & 0.165 & 0.269 & 0.164 & 0.272 & 0.171 & 0.281 & 0.172 & 0.282 & 0.172 & 0.274 & 0.180 & 0.292 & 0.179 & 0.291 & 0.208 & 0.321 & 0.163 & 0.275 & 0.191 & 0.304 \\
Weather & \best{0.235} & \best{0.259} & \secondbest{0.239} & \secondbest{0.269} & 0.246 & 0.275 & 0.253 & 0.276 & 0.256 & 0.281 & 0.251 & 0.273 & 0.243 & 0.272 & 0.244 & 0.274 & 0.251 & 0.276 & 0.261 & 0.283 & 0.245 & 0.273 & 0.245 & 0.275 & 0.258 & 0.278 & 0.263 & 0.284 & 0.249 & 0.278 & 0.259 & 0.281 & 0.259 & 0.287 \\
Traffic & 0.461 & \secondbest{0.276} & \best{0.409} & \best{0.269} & 0.448 & 0.291 & 0.450 & 0.306 & 0.505 & 0.324 & 0.451 & \best{0.269} & 0.601 & 0.363 & 0.471 & 0.315 & 0.440 & 0.286 & 0.453 & 0.286 & 0.490 & 0.326 & 0.484 & 0.297 & \secondbest{0.428} & 0.282 & 0.541 & 0.358 & 0.660 & 0.381 & 0.481 & 0.304 & 0.620 & 0.336 \\
Electricity & \best{0.166} & \best{0.258} & \secondbest{0.169} & 0.262 & 0.181 & 0.271 & 0.174 & 0.266 & 0.192 & 0.278 & 0.170 & \secondbest{0.261} & 0.197 & 0.286 & 0.180 & 0.271 & \secondbest{0.169} & 0.262 & 0.187 & 0.274 & 0.182 & 0.280 & 0.189 & 0.278 & 0.179 & 0.269 & 0.209 & 0.289 & 0.194 & 0.300 & 0.187 & 0.290 & 0.193 & 0.295 \\
Solar & \best{0.200} & \best{0.231} & 0.213 & 0.261 & 0.210 & 0.244 & 0.206 & 0.264 & \secondbest{0.204} & 0.260 & 0.237 & \secondbest{0.233} & 0.250 & 0.297 & 0.232 & 0.278 & 0.232 & 0.266 & 0.238 & 0.275 & 0.227 & 0.260 & 0.216 & 0.280 & 0.233 & 0.262 & 0.224 & 0.268 & 0.269 & 0.300 & 0.270 & 0.307 & 0.261 & 0.302 \\
PEMS (Avg) & \best{0.100} & \best{0.190} & \secondbest{0.102} & \secondbest{0.206} & 0.128 & 0.233 & 0.123 & 0.231 & 0.179 & 0.280 & 0.107 & 0.209 & 0.229 & 0.314 & 0.150 & 0.241 & 0.134 & 0.236 & 0.162 & 0.249 & 0.135 & 0.237 & 0.189 & 0.279 & 0.119 & 0.218 & 0.201 & 0.290 & 0.150 & 0.251 & 0.216 & 0.311 & 0.149 & 0.246 \\
ILI & \best{1.934} & \best{0.852} & 2.098 & \secondbest{0.871} & 2.247 & 0.897 & 2.386 & 0.950 & 2.441 & 1.023 & 2.449 & 0.988 & 2.621 & 0.996 & 2.186 & 0.926 & 2.303 & 0.940 & 2.250 & 0.926 & 2.112 & 0.899 & 2.020 & 0.878 & \secondbest{1.993} & 0.888 & 2.147 & 0.891 & 2.492 & 0.914 & 2.208 & 0.910 & 2.403 & 0.982 \\
NASDAQ & \best{0.180} & \best{0.271} & 0.198 & 0.288 & 0.198 & 0.288 & 0.197 & 0.287 & 0.194 & 0.285 & 0.195 & 0.286 & 0.209 & 0.299 & 0.203 & 0.293 & 0.187 & \secondbest{0.277} & 0.197 & 0.287 & 0.199 & 0.289 & \secondbest{0.186} & 0.282 & 0.207 & 0.297 & 0.205 & 0.295 & 0.252 & 0.341 & 0.198 & 0.287 & 0.254 & 0.343 \\

\cmidrule{1-35}
\multicolumn{1}{c}{\textbf{MV-1\textsuperscript{st} Count}}
& \multicolumn{2}{c}{\best{15}}
& \multicolumn{2}{c}{\secondbest{3}}
& \multicolumn{2}{c}{0}
& \multicolumn{2}{c}{1}
& \multicolumn{2}{c}{2}
& \multicolumn{2}{c}{1}
& \multicolumn{2}{c}{0}
& \multicolumn{2}{c}{0}
& \multicolumn{2}{c}{0}
& \multicolumn{2}{c}{0}
& \multicolumn{2}{c}{0}
& \multicolumn{2}{c}{0}
& \multicolumn{2}{c}{0}
& \multicolumn{2}{c}{0}
& \multicolumn{2}{c}{0}
& \multicolumn{2}{c}{0}
& \multicolumn{2}{c}{0} \\

% ==================== MM ====================
\midrule
Agriculture & \best{0.104} & \best{0.207} & 0.107 & 0.214 & \secondbest{0.105} & 0.212 & \secondbest{0.105} & 0.212 & 0.107 & 0.212 & 0.113 & 0.227 & 0.128 & 0.240 & 0.107 & \secondbest{0.211} & 0.107 & 0.217 & 0.110 & 0.213 & 0.106 & \best{0.207} & 0.126 & 0.239 & \best{0.104} & \secondbest{0.211} & 0.108 & 0.214 & 0.112 & 0.224 & 0.106 & 0.214 & 0.109 & 0.222 \\
Climate & \best{0.890} & \best{0.757} & 0.914 & 0.767 & 0.912 & 0.767 & 0.911 & 0.766 & 0.927 & 0.772 & 0.909 & 0.765 & 0.912 & 0.767 & 0.930 & 0.775 & 0.956 & 0.789 & 0.931 & 0.778 & 0.915 & 0.771 & 0.910 & 0.766 & 0.918 & 0.770 & 0.926 & 0.772 & \secondbest{0.906} & 0.764 & 0.912 & 0.766 & 0.913 & \secondbest{0.761} \\
Economy & \best{0.204} & \best{0.357} & 0.215 & 0.370 & \secondbest{0.212} & \secondbest{0.363} & 0.229 & 0.382 & 0.216 & 0.367 & 0.223 & 0.377 & 0.237 & 0.391 & 0.220 & 0.372 & 0.214 & 0.368 & 0.230 & 0.384 & 0.216 & 0.369 & 0.240 & 0.392 & 0.224 & 0.379 & 0.222 & 0.377 & 0.233 & 0.389 & 0.226 & 0.379 & 0.223 & 0.376 \\
Energy & \best{0.172} & \best{0.323} & 0.176 & 0.327 & 0.178 & 0.330 & 0.181 & 0.332 & 0.177 & 0.327 & \secondbest{0.175} & \secondbest{0.324} & 0.184 & 0.330 & 0.177 & \secondbest{0.324} & 0.182 & 0.334 & 0.180 & 0.330 & 0.178 & 0.325 & 0.183 & 0.330 & 0.178 & 0.330 & 0.183 & 0.330 & 0.183 & 0.331 & 0.182 & 0.334 & 0.181 & 0.331 \\
Environment & \best{0.267} & \best{0.366} & \secondbest{0.268} & 0.368 & \secondbest{0.268} & 0.370 & \secondbest{0.268} & 0.368 & \best{0.267} & 0.369 & 0.269 & 0.372 & \secondbest{0.268} & 0.369 & \secondbest{0.268} & 0.371 & 0.269 & \secondbest{0.367} & \secondbest{0.268} & 0.371 & \best{0.267} & 0.371 & \secondbest{0.268} & 0.371 & 0.270 & 0.370 & 0.269 & 0.372 & \secondbest{0.268} & 0.370 & 0.269 & 0.368 & 0.269 & \best{0.366} \\
Health & \best{0.732} & \best{0.572} & 0.767 & 0.583 & \secondbest{0.742} & 0.580 & 0.754 & 0.580 & 0.756 & 0.586 & 0.777 & 0.587 & 0.759 & 0.587 & 0.808 & 0.608 & 0.761 & 0.579 & 0.751 & 0.578 & 0.766 & 0.585 & 0.768 & 0.584 & 0.758 & \secondbest{0.574} & 0.765 & 0.585 & 0.750 & 0.577 & 0.760 & 0.584 & 0.798 & 0.596 \\
Security & \best{74.483} & \best{3.838} & 75.309 & 3.934 & \secondbest{74.669} & \secondbest{3.847} & 75.189 & 3.954 & 76.194 & 3.973 & 75.502 & 3.913 & 75.085 & 3.858 & 76.602 & 4.065 & 77.666 & 4.150 & 76.237 & 3.994 & 75.685 & 3.925 & 75.250 & 3.910 & 76.401 & 4.020 & 76.883 & 4.096 & 75.602 & 3.928 & 75.801 & 3.935 & 75.270 & 3.939 \\
SocialGood & \best{0.889} & \best{0.476} & 0.957 & 0.527 & 0.930 & 0.530 & 0.926 & \secondbest{0.492} & 0.982 & 0.532 & \secondbest{0.922} & 0.500 & 0.946 & 0.518 & 0.925 & 0.511 & 0.929 & 0.516 & 0.935 & 0.522 & 1.029 & 0.557 & 1.009 & 0.561 & 0.933 & 0.507 & 0.940 & 0.510 & 0.947 & 0.508 & 0.956 & 0.563 & 0.957 & 0.529 \\
Traffic & \best{0.163} & 0.229 & 0.181 & 0.228 & 0.181 & 0.233 & 0.177 & 0.227 & 0.184 & 0.236 & \secondbest{0.170} & 0.234 & 0.186 & 0.253 & 0.177 & 0.246 & 0.177 & \best{0.222} & 0.178 & \secondbest{0.224} & 0.177 & 0.228 & 0.183 & 0.245 & 0.178 & 0.225 & 0.180 & 0.230 & 0.185 & 0.242 & 0.174 & \best{0.222} & 0.198 & 0.248 \\
PTF & \best{0.256} & \best{0.340} & 0.361 & 0.424 & 0.387 & 0.435 & \secondbest{0.356} & \secondbest{0.415} & 0.686 & 0.586 & 0.421 & 0.453 & 1.104 & 0.792 & 0.731 & 0.623 & 0.501 & 0.489 & 0.875 & 0.684 & 0.481 & 0.466 & 1.043 & 0.766 & 0.585 & 0.542 & 1.011 & 0.755 & 0.431 & 0.467 & 0.912 & 0.710 & 0.416 & 0.439 \\
LEU & \best{0.663} & \best{0.470} & 0.798 & 0.522 & \secondbest{0.765} & 0.515 & 0.797 & 0.517 & 0.834 & 0.533 & 0.842 & 0.499 & 0.908 & 0.519 & 0.881 & 0.511 & 0.913 & 0.519 & 0.907 & 0.516 & 0.825 & 0.496 & 0.899 & 0.512 & 0.845 & 0.498 & 0.919 & 0.525 & 0.799 & 0.495 & 0.910 & 0.518 & 0.800 & \secondbest{0.494} \\
MSPG & \best{0.361} & \best{0.326} & 0.420 & 0.353 & \secondbest{0.418} & \secondbest{0.348} & 0.422 & 0.352 & 0.449 & 0.365 & 0.501 & 0.379 & 0.617 & 0.428 & 0.561 & 0.400 & 0.623 & 0.426 & 0.538 & 0.392 & 0.499 & 0.371 & 0.579 & 0.411 & 0.527 & 0.380 & 0.602 & 0.423 & 0.534 & 0.390 & 0.569 & 0.405 & 0.558 & 0.401 \\

\cmidrule{1-35}
\multicolumn{1}{c}{\textbf{MM-1\textsuperscript{st} Count}}
& \multicolumn{2}{c}{\best{23}}
& \multicolumn{2}{c}{0}
& \multicolumn{2}{c}{0}
& \multicolumn{2}{c}{0}
& \multicolumn{2}{c}{1}
& \multicolumn{2}{c}{0}
& \multicolumn{2}{c}{0}
& \multicolumn{2}{c}{0}
& \multicolumn{2}{c}{1}
& \multicolumn{2}{c}{0}
& \multicolumn{2}{c}{\secondbest{2}}
& \multicolumn{2}{c}{0}
& \multicolumn{2}{c}{1}
& \multicolumn{2}{c}{0}
& \multicolumn{2}{c}{0}
& \multicolumn{2}{c}{1}
& \multicolumn{2}{c}{1} \\

\bottomrule[1.5pt]
\end{tabular}%
}
\end{table*}

\begin{figure*}
\centering
\includegraphics[width=1.0 \linewidth]{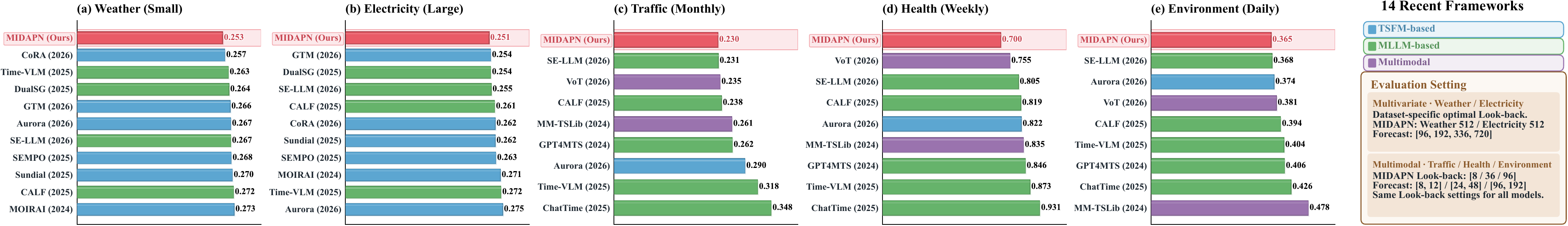}
\caption{TSFMs, fused-PLMs and multimodal frameworks comparison (sorted by averaged MAE from low to high). MIDAPN consistently achieves the lowest MAE across the selected benchmarks, highlighting its accuracy advantage over pretrained forecasting approaches under extended look-back settings}
\label{figure-mtsfm}
\end{figure*}

\section{Experiments}
\label{exp}
\subsection{Experimental Settings}
\label{set}
\subsubsection{Datasets} We have summarized 25 public datasets from both "multivariate" and "multimodal" scenarios in Table \ref{tab:dataset_stats_wide}:

$\bullet$ \textit{Multivariate:}
\textit{ETT\{m2, h2\}}: Records temperature and power load characteristics of electricity transformers at minute-level and hour-level granularity. \textit{Flight}: Flight data variations from seven major European airports during the COVID-19 pandemic. \textit{Weather}: Meteorological indicators from German weather stations. \textit{Traffic}: Road occupancy rates measured on San Francisco Bay Area freeways. \textit{Electricity}: Records of customer electricity kWh consumption. \textit{Solar}: Data from photovoltaic power plants in Alabama State. \textit{PEMS\{03, 04, 07, 08\}}: Highway traffic flow from California. \textit{ILI}: Trends in the number of influenza patients in the United States. \textit{NASDAQ}: Daily Nasdaq index and key economic indicators.

$\bullet$ \textit{Multimodal (Time-MMD):}
\textit{Agriculture}: Composite retail broiler chicken data. \textit{Climate}: Climate trends and impact on environmental policies. \textit{Economy} (corrected order \cite{wang2026unlocking}): United States international trade in goods and services. \textit{Energy}: Gasoline prices variation in the energy market. \textit{Environment}: New York air quality data. \textit{Health}: Influenza-like illness statistics. \textit{Security}: Billion-dollar disasters and emergency declarations. \textit{SocialGood}: Employment trends across different racial groups. \textit{Traffic}: Vehicle miles traveled on highways. 

$\bullet$ \textit{Multivariate + Multimodal (ChatTime):} MSPG records 13 months of solar generation from 27 photovoltaic sites in Melbourne , LEU contains 24 months of electricity usage from 16 London households, and PTF includes 12 months of traffic flow from 32 Paris detectors. Note that LEU, MSPG, and PTF originally consist of dispersed
time-series--text pairs. We reorganize and convert them into the TaTS-compatible format.

\subsubsection{Baselines} We initially selected 16 SOTA native deep TSF models. (1) Transformer-based: PatchTST \cite{patchtst}, iTransformer \cite{itransformer}, DUET \cite{DUET}; (2) MLP-based: SEMixer \cite{semixer}, TimeMixer \cite{timemixer}, FilterTS \cite{filterts}, CrossLinear + CPiRi (plug-in) \cite{cpiri,crosslinear}; (3) TCN-based: TimesNet \cite{timesnet}, ModernTCN \cite{moderntcn}, VPNet \cite{vpnet}; (4) LLM-based: Time-LLM \cite{timellm}; (5) RNN-based: P-sLSTM \cite{unlock}; (6) GCN-based: MSGNet \cite{msgnet}, TimeFilter \cite{hu2025timefilter};  (7) KAN-based: TimeKAN \cite{timekan}; (8) Mamba-based: TimePro \cite{timepro}. Furthermore, 14 latest TSFMs \& fused-PLMs are also included. (1) TSFM: Aurora \cite{aurora}, GTM \cite{gtm}, CoRA (plug-in) \cite{cheng2026cora}, Sundial \cite{liu2025sundial}, SEMPO \cite{he2025sempo}, MOIRAI \cite{woo2024unified}; (2) PLM-fused: SE-LLM \cite{sellm}, Time-VLM \cite{zhong2025time}, CALF \cite{calf}, DualSG \cite{dualsg}, GPT4MTS \cite{jia2024gpt4mts}, ChatTime \cite{chattime}; (3) Modality Fused Strategy: VoT \cite{wang2026unlocking}, MM-TSLib \cite{liu2024time}.

\subsubsection{Implementation Details} Experiments were trained on NVIDIA GeForce RTX 4090 GPUs. For "multivariate", we conformed the Time-series Library \cite{timesnet}; To equip deep TSF models with "multimodal" capabilities, we applied the TaTS framework \cite{li2025language} to provide modal pre-alignment (numerical \& PLM-text projection) for baselines in time series form, making it straightforward to assess the scaling potential. For train/valid/test sets division, the ETT\{m2, h2\} and PEMS\{03, 04, 07, 08\} were 6:2:2 ratio, while the others were 7:1:2 ratio. For metrics, Mean Absolute Error (MAE) and Mean Squared Error (MSE) metrics are used to assess robustness and sensitivity.
Table \ref{tab:comprehensive_merged_categorized} provides comparison under the same Look-back window for prediction limits. As TSFMs and PLM-fused models are inherently designed for longer contexts, Fig. \ref{figure-mtsfm} supports this by reporting optimal long-context results.

\begin{figure}
\centering
\includegraphics[width=1.0 \linewidth]{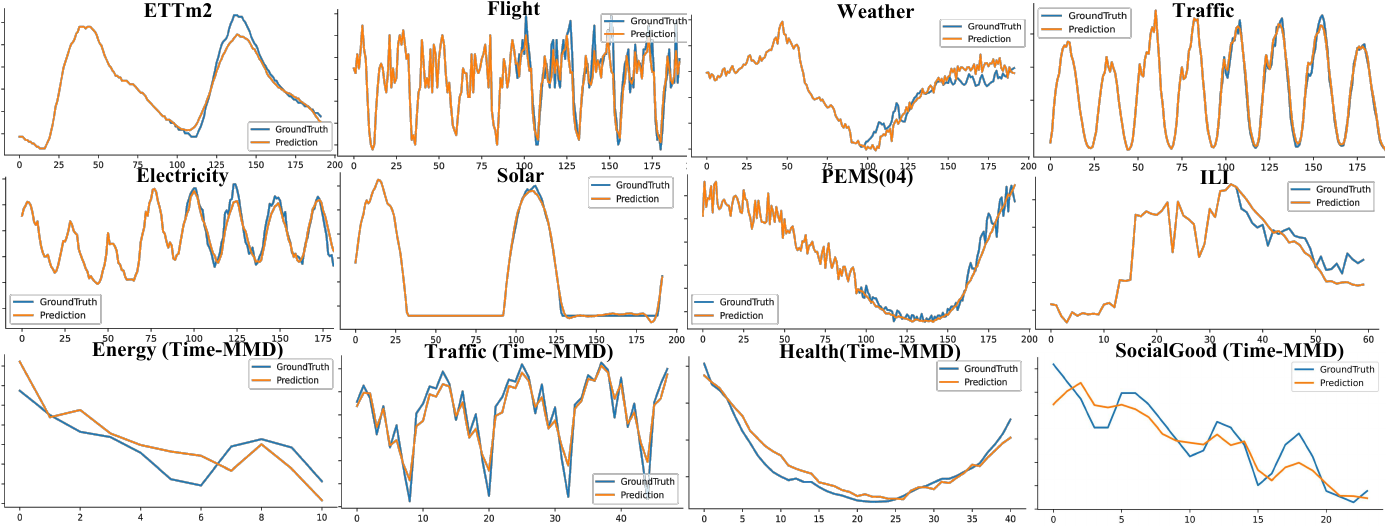}
\caption{Forecasting across media-bridged domains with periodic, volatile, and trend-dominated patterns. MIDAPN generally tracks major trends and recurring oscillations, with localized deviations around abrupt changes}
\label{fig:all-vis}
\end{figure}

\subsection{Experimental Results}
\label{erha}
\textbf{(1) "Multivariate":} From Table \ref{tab:comprehensive_merged_categorized}, MIDAPN achieves consistent and competitive improvements of (11.6\%, 14.2\%) in (MAE, MSE), ranking first in (15 / 18) metrics. In contrast to the redundancy-pruning approach (TimeFilter) of spatio-temporal graph adaptation, MIDAPN enhances structural semantics of variable identity, realizing SOTA improvement of (3.88\%, 3.56\%) in (MAE, MSE). \textbf{(2) "Multimodal":} MIDAPN exhibits remarkable performance, ranking first in (23 / 24) metrics. Although VPNet's local sufficiency hypothesis \cite{vpnet} partially eliminates redundant inter-modal correlations, MIDAPN essentially treats temporal-textual patterns as identity-bearing media signals (Sec. \ref{6.1.1}), indicating gains from genuine post-alignment. Particularly on the PTF, LEU, and MSPG datasets, MIDAPN effectively bridges textual backgrounds with multivariate. \textbf{(3) "Long Context vs TSFM \& PLM":} We conducted a definitive assessment across datasets with diverse multivariate scales and multimodal temporal granularities, extended Look-back window is allowed to leverage the temporal context. Fig. \ref{figure-mtsfm} suggests that MIDAPN remains competitive with recent TSFM- and PLM-based methods,
indicating that its effectiveness extends beyond lightweight forecasters. \textbf{(4) "Overall Performance":} MIDAPN achieves outstanding results in 25 real-world datasets, ranking first in (38 / 42) and top two in (40 / 42) metrics. Fig. \ref{fig:all-vis} visualizes its multi-domain performance. These findings validate our ST-extension—a unified backbone elegantly disentangles identities and weaves temporal scales for media-general modeling.

\subsection{Ablation and Replacement Results}
\label{arr}

We conducted MIDAPN ablation (component mechanisms and temporal/text validation)  in multimedia context. To clarify the notation, \textbf{"w/o" "(DFF/MIDAG/SPConv)"} signify removing its components while \textbf{"w/o-w"} represent replacement.

From Table \ref{table-Ablation-study-multivariate-avg}: (1) For \textbf{overall ablation}, removing any core components of MIDAPN incurs a substantial penalty. Especially in context-guided multivariate analysis, MIDAG demonstrates its indispensable media-bridged role, we then provide its detailed analysis. Moreover, ablating the Decoupled Frequency Fusion module  results in an incomplete and impoverished representation; while removing the Spectral Prism Convolution discards the hierarchical strategy within automatic multi-scale temporal extraction.  (2) For \textbf{architecture ablation}, removing either $\mathrm{TSEM}_{\mathrm{his}}$ or $\mathrm{TSEM}_{\mathrm{pred}}$ compromises embedding validity. Concurrently, eliminating all TSEM modules inevitably undermines the integrity of the hierarchical spatio-temporal manifold. Actually $\text{TSEM}_{\text{his}}$ for historical spatio-temporal manifold tends to be more pronounced. (3) For \textbf{module replace}, substituting proposed modules with basic alternatives also degrades performance  (i.e., MIDAG to Adaptive Graph, SPConv to Conv or Freq-Conv), which demonstrates the advantages of MIDAG and SPConv for media-aware correlations and temporal dependencies. (4) \textbf{MIDAG ablations},  ablating the essence identity eliminates prior knowledge of multivariate relationships, which obscures the conceptual distinction \cite{meta} between variates; Conversely, removing the behavior identity renders it insensitive to dynamic patterns; Meanwhile, the latent cluster centers missing disrupts MIDAPN to assign group commonalities modeling in heterogeneous media characterization. In addition, omitting contextual identity modulation allows noise to propagate and ignores underlying ST-structure patterns. In summary, the absence of any MIDAG modules leads to impoverished Identity-Aware Graph. (5) \textbf{Mechanistic evidence} shows that reversing MIDAPN's $\Psi_{\text{spatial}}$, $\Psi_{\text{temporal}}$ order hurts performance, while its sensitivity to narrative flow shuffling highlights robustness.

\begin{table}[t]
\centering
\caption{Ablation studies (For averaged horizons, multivariate: \{96, 192, 336, 720\}, multimodal: \{8, 10, 12\}). The "Drop" is the error percentage increase in Average of MAE and MSE}
\label{table-Ablation-study-multivariate-avg}
\resizebox{\linewidth}{!}{
\begin{tabular}{c|cccccccccc}
\toprule[1.5pt]
Task Types & \multicolumn{4}{c}{Multivariate} & \multicolumn{4}{c}{Multimodal} & \multicolumn{2}{c}{Multivariate + Text} \\
\cmidrule(lr){2-5} \cmidrule(lr){6-9} \cmidrule(lr){10-11}
Datasets & \multicolumn{2}{c}{Electricity} & \multicolumn{2}{c}{Weather} & \multicolumn{2}{c}{Traffic} & \multicolumn{2}{c}{Energy} & \multicolumn{2}{c}{LEU} \\
\cmidrule(lr){2-3} \cmidrule(lr){4-5} \cmidrule(lr){6-7} \cmidrule(lr){8-9} \cmidrule(lr){10-11}
Strategies & Avg & Drop & Avg & Drop & Avg & Drop & Avg & Drop & Avg & Drop \\
\midrule
\textbf{MIDAPN} & \textbf{0.2135} & --- & \textbf{0.2468} & --- & \textbf{0.1963} & --- &\textbf{0.2478} & --- & \textbf{0.5665} & --- \\
\midrule
w/o DFF & 0.2389 & \textbf{11.89\%} & 0.2541 & \textbf{2.99\%}  &0.2122 &\textbf{8.06\%}  &0.2560  &\textbf{3.30\%} & 0.6233 & \textbf{10.03\%} \\
w/o MIDAG & 0.2384 & \textbf{11.65\%} & 0.2526 & \textbf{2.38\%} &0.2113  &\textbf{7.64\%}  &0.2555 &\textbf{3.09\%}  & 0.7015 & \textbf{23.83\%} \\
w/o SPConv & 0.2428 & \textbf{13.70\%} & 0.2546 & \textbf{3.19\%} &0.2117  &\textbf{7.81\%}  &0.2508 &\textbf{1.21\%}  & 0.5792 & \textbf{2.24\%} \\
\midrule
w/o TSEM-his &0.2188  & \textbf{2.46\%} &0.2486  & \textbf{0.76\%} &0.2100  &\textbf{6.96\%}  &0.2495 &\textbf{0.67\%}  & 0.5983 & \textbf{5.62\%} \\
w/o TSEM-pred &0.2170  & \textbf{1.64\%} &0.2486  & \textbf{0.76\%} &0.2130  &\textbf{8.49\%}  &0.2487 &\textbf{0.34\%}  & 0.5855 & \textbf{3.35\%} \\
w/o all & 0.2455 & \textbf{14.99\%} & 0.2551 & \textbf{3.39\%} &0.2250  &\textbf{14.60\%}  &0.2795 &\textbf{12.77\%}  & 0.7162 & \textbf{26.42\%} \\
\midrule
w/o-w ADP &0.2234  & \textbf{4.63\%} &0.2486 & \textbf{0.76\%} &0.2072  &\textbf{5.52\%}  &0.2548 &\textbf{2.82\%} & 0.6405 & \textbf{13.06\%} \\
w/o-w Conv & 0.2439 & \textbf{14.23\%} & 0.2606 & \textbf{5.62\%} &0.1988  &\textbf{1.27\%}  &0.2587 &\textbf{4.37\%}  & 0.5863 & \textbf{3.50\%} \\
w/o-w Freq-Conv & 0.2216 & \textbf{3.81\%} & 0.2536 & \textbf{2.79\%} &0.2045  &\textbf{4.16\%}  &0.2547 &\textbf{2.76\%}  & 0.5877 & \textbf{3.74\%} \\
\midrule
w/o MIDAG-EID & 0.2159 & \textbf{1.11\%} & 0.2539 & \textbf{2.89\%} &0.2077  &\textbf{5.77\%}  &0.2497 &\textbf{0.74\%}  & 0.5735 & \textbf{1.24\%} \\
w/o MIDAG-BID & 0.2180 & \textbf{2.11\%} & 0.2580 & \textbf{4.56\%} & 0.2120 &\textbf{7.98\%}  &0.2545 &\textbf{2.69\%}  & 0.6217 & \textbf{9.74\%} \\
w/o MIDAG-Cluster & 0.2163 & \textbf{1.29\%} & 0.2508 & \textbf{1.62\%} &0.2117  &\textbf{7.81\%}  &0.2523 &\textbf{1.82\%}  & 0.5747 & \textbf{1.44\%} \\
w/o MIDAG-CIM &0.2168  & \textbf{1.52\%} &0.2478 & \textbf{0.41\%} &0.2067  &\textbf{5.26\%}  &0.2557 &\textbf{3.16\%}  & 0.5802 & \textbf{2.41\%} \\
\midrule 
SPConv-MIDAG &0.2400 & \textbf{12.41\%} &0.2558 & \textbf{3.65\%} &0.2087  &\textbf{6.28\%}  &0.2505 &\textbf{1.08\%}  &0.5767 & \textbf{1.79\%} \\
Temporal/Text Shuffling &0.8353  & \textbf{291.22\%} &0.2951 & \textbf{16.13\%} &0.2222  &\textbf{13.16\%}  &0.3007 &\textbf{21.32\%}  &0.6623 & \textbf{16.92\%} \\
\bottomrule[1.5pt]
\end{tabular}
}
\end{table}

\section{Discussion}
\label{discuss}
\subsection{Interpretability and Generality Study}
\label{Interpretability}
\subsubsection{Identity-Aware Graph Adaptation} \label{6.1.1} 
\begin{figure}
\centering
\includegraphics[width=1.0 \linewidth]{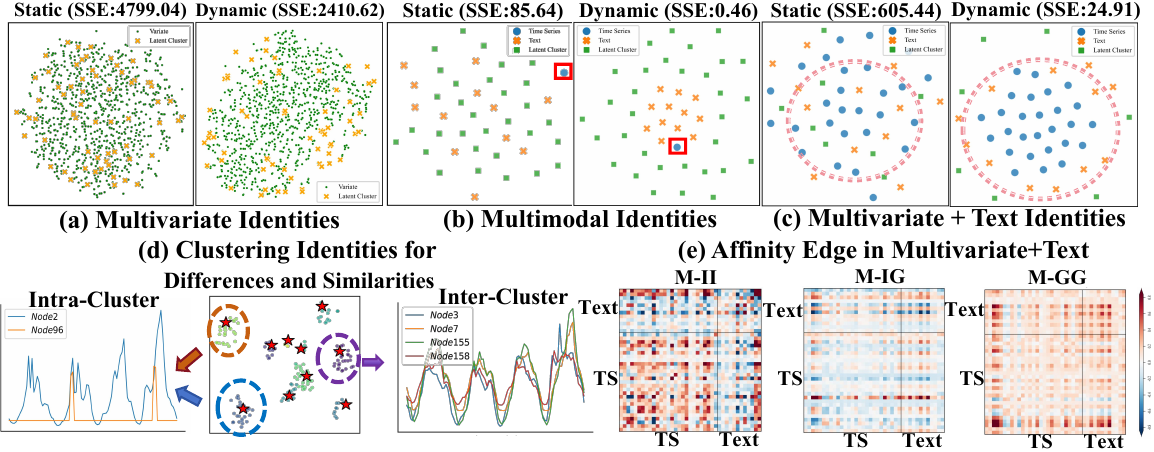}
\caption{Identity-Aware Graph Adaptation (multivariate: Traffic, multimodal: Climate, multivariate + text: MSPG). (a, b, c): Variate identity visualizations; (d): Clustering visualization; (e): Cross-level Affinity Decomposition}
\label{fig:graph-vis}
\end{figure}

Rather
than adaptive-graph variants only fit unimodal TSF, MIDAG is
generated in media-bridged space. From the 2D t-SNE visualization: (1) For "multivariate" in Fig. \ref{fig:graph-vis} (a): The essence identities are distributed in a relatively loose and uniform manner, reflecting the intrinsic variate properties. In contrast, the behavior identities are conditioned upon the immediate Look-back window, hence they exhibit clear convergence. In addition, overlapping cluster centers typically indicate a similarity structure. (2) For "multimodal" in Fig. \ref{fig:graph-vis} (b,c): The essence identities reveal the natural separation between time-series and textual auxiliary variates, which stems from modality heterogeneity, while the behavior identities narrow the inter-modal distance. Overall, this joint adaptation covers property \& heterogeneity ($\mathcal{E}_{\text{id}}$), and dynamism \& aggregation ($\mathcal{B}_{\text{id}}$) knowledge.

\subsubsection{Clustering Analysis} In the Fig. \ref{fig:graph-vis} (d), we select 10 latent cluster centers from the sensors subset in a local region of the multivariate Traffic dataset. As depicted, time-series variates within the same cluster (e.g., variate 3, 7, 155, and 158 in the purple circle) exhibit highly similar evolutionary patterns within their Look-back windows. In contrast, distant clusters such as the brown and blue circles correspond to variate (2 and 96) with distinctly different temporal characteristics, respectively showcasing periodic cycles versus abrupt patterns. This phenomenon reflects that the latent cluster centers can perform implicit group clustering \cite{cluster}.

\subsubsection{Affinity Decomposition} \label{AFFD} Based on MSPG, we decompose affinity into individual ($M_{\mathrm{II}}$), cross-level ( $M_{\mathrm{IG/GI}}$), and group ($M_{\mathrm{GG}}$) interactions. Fig. ~\ref{fig:graph-vis}(e) reveals: $M_{\mathrm{II}}$ reinforces TS–TS dependencies, prioritizes forecast-relevant factors within compressed sentence embeddings. $M_{\mathrm{GG}}$ establishes shared intra-media structure, $M_{\mathrm{IG/GI}}$ suppresses redundant affinities for selective TS–TS/Text alignment.

\subsubsection{"Narrative-Flow" Case Study} 
\begin{figure}
\centering
\includegraphics[width=1.0 \linewidth]{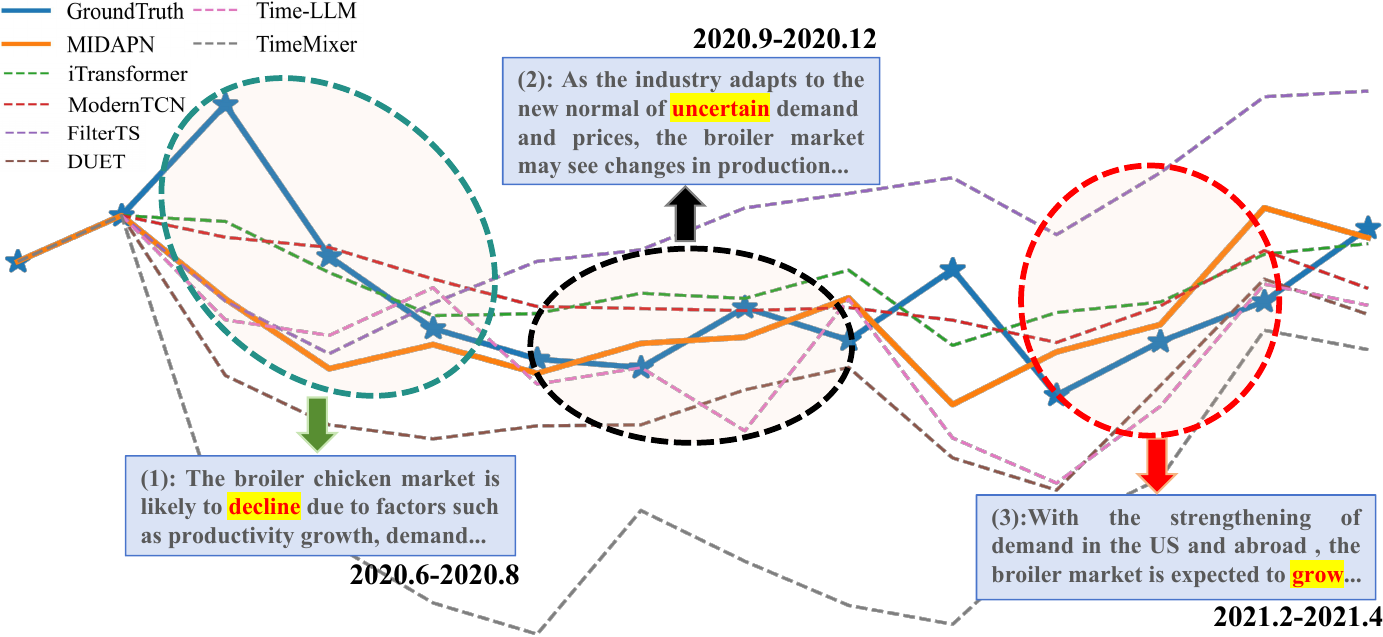}
\caption{``Narrative-Flow'' forecasting on the Agriculture dataset across
declining, uncertain, and rising narrative-guided market periods}
\label{fig:modal-vis}
\end{figure}

We focus time series-text pairs (April 2020 to May 2021) on: (1) Declining: Jun–Aug 2020; (2) Uncertain: Sep–Dec 2020; (3) Rising: Feb–Apr 2021. Fig. \ref{fig:modal-vis} shows MIDAPN's best forecast approximation.

\subsubsection{"Graph-Informed" Generality}
Table \ref{tab:generality} demonstrates MIDAG's consistent plug-and-play capacity across mainstream TSF architectures. Crucially, PTF inherently suffers from redundancy due to identical "clone nodes". MIDAG's explicit identity awareness effectively improves this anomaly.

\begin{table}[htbp]
\centering
\caption{MIDAG Generalization in "multivariate" (input-96-predict-96) and "multimodal" (input-8-predict-12) scenarios}
\label{tab:generality}
\resizebox{\linewidth}{!}{%
\begin{tabular}{l|cc|cc|cc|cc|cc}
\toprule[1.5pt]
% --- 一级表头 ---
Task Types & \multicolumn{4}{c|}{Multivariate} & \multicolumn{4}{c|}{Multimodal} & \multicolumn{2}{c}{\begin{tabular}[c]{@{}c@{}}Multivariate \\ + Text\end{tabular}} \\
\cmidrule(lr){2-5} \cmidrule(lr){6-9} \cmidrule(lr){10-11}
% --- 二级表头 ---
Datasets & \multicolumn{2}{c|}{Solar} & \multicolumn{2}{c|}{Weather} & \multicolumn{2}{c|}{Climate} & \multicolumn{2}{c|}{Agriculture} & \multicolumn{2}{c}{PTF} \\
\cmidrule(lr){2-3} \cmidrule(lr){4-5} \cmidrule(lr){6-7} \cmidrule(lr){8-9} \cmidrule(lr){10-11}
% --- 三级表头 (指标) ---
Backbone \& Models & MAE & MSE & MAE & MSE & MAE & MSE & MAE & MSE & MAE & MSE \\
\midrule
% --- Transformer 数据行 ---
SageFormer \cite{sageformer} & 0.259 & 0.197 & 0.206 & 0.162 & 0.766 & 0.912 & 0.242 & 0.136 & 0.595 & 0.674 \\
(Transformer)+MIDAG & \textbf{0.235} & \textbf{0.189} & \textbf{0.198} & \textbf{0.161} & \textbf{0.762} & \textbf{0.909} & \textbf{0.240} & \textbf{0.131} & \textbf{0.548} & \textbf{0.635} \\
\midrule
% --- Linear 数据行 ---
DLinear \cite{dlinear} & 0.236 & 0.312 & 0.255 & 0.196 & 0.765 & 0.920 & 0.304 & 0.194 & 0.953 & 1.454 \\
(Linear)+MIDAG & \textbf{0.229} & \textbf{0.190} & \textbf{0.218} & \textbf{0.164} & \textbf{0.748} & \textbf{0.902} & \textbf{0.286} & \textbf{0.152} & \textbf{0.568} & \textbf{0.628} \\
\midrule
% --- LLM 数据行 ---
Time-LLM \cite{timellm} & 0.249 & 0.196 & 0.219 & 0.178 & 0.774 & 0.933 & 0.240 & 0.137 & 0.780 & 1.068 \\
(LLM)+MIDAG & \textbf{0.224} & \textbf{0.191} & \textbf{0.204} & \textbf{0.164} & \textbf{0.764} & \textbf{0.915} & \textbf{0.238} & \textbf{0.135} & \textbf{0.757} & \textbf{1.038} \\
\midrule
% --- TSFM 数据行 ---
UniCA (Chronos-Bolt) \cite{han2026unica} &0.280 &0.241 & 0.299 & 0.239 & 0.763 & 0.912 & 0.260 & 0.141 & 0.512 & 0.523 \\
(TSFM)+MIDAG  &\textbf{0.275} &\textbf{0.237} &\textbf{0.243} &\textbf{0.191} & \textbf{0.753} & \textbf{0.912} & \textbf{0.259} & \textbf{0.124} & \textbf{0.258} & \textbf{0.340} \\
\midrule
% --- KAN 数据行 ---
TimeKAN \cite{timekan} & 0.275 & 0.215 & 0.208 & 0.162 & 0.765 & 0.916 & 0.261 & 0.154 & 0.825 & 1.193 \\
(KAN)+MIDAG & \textbf{0.220} & \textbf{0.194} & \textbf{0.198} & \textbf{0.162} & \textbf{0.764} & \textbf{0.908} & \textbf{0.251} & \textbf{0.145} & \textbf{0.759} & \textbf{0.957} \\
\midrule
% --- Mamba 数据行 ---
TimePro \cite{timepro} & 0.237 & 0.196 & 0.207 & 0.166 & 0.799 & 0.975 & 0.241 & 0.133 & 0.521 & 0.544 \\
(Mamba)+MIDAG & \textbf{0.204} & \textbf{0.178} & \textbf{0.195} & \textbf{0.159} & \textbf{0.788} & \textbf{0.958} & \textbf{0.238} & \textbf{0.133} & \textbf{0.498} & \textbf{0.519} \\
\bottomrule[1.5pt]
\end{tabular}
}
\end{table}

\subsection{Hyperparameter Analysis}
We analyze MIDAPN hyperparameters using two settings: "multivariate" PEMS04 (Input-96-predict-12) for stable sensitivity evaluation under rich temporal context, and "multimodal" datasets for grid-search optimality analysis (Input-8-predict-12) due to coarse timestamps and sample sparsity.

\subsubsection{Sensitivity Analysis} Based on Fig. \ref{figure-hyper-vis}(a), the optimal hyperparameter configurations (red boxes) are distinctly different. Furthermore, continuously increasing a hyperparameter's value (e.g., Model Dimension or Graph Dimension) does not always lead to performance gains, suggesting that an optimal balance point must be found experimentally. Meanwhile, the performance (blue curve) does not fluctuate drastically.

\subsubsection{Optimal Selection in Multi-Domain}
\label{OS}
\begin{figure}
\centering
\includegraphics[width=1.0 \linewidth]{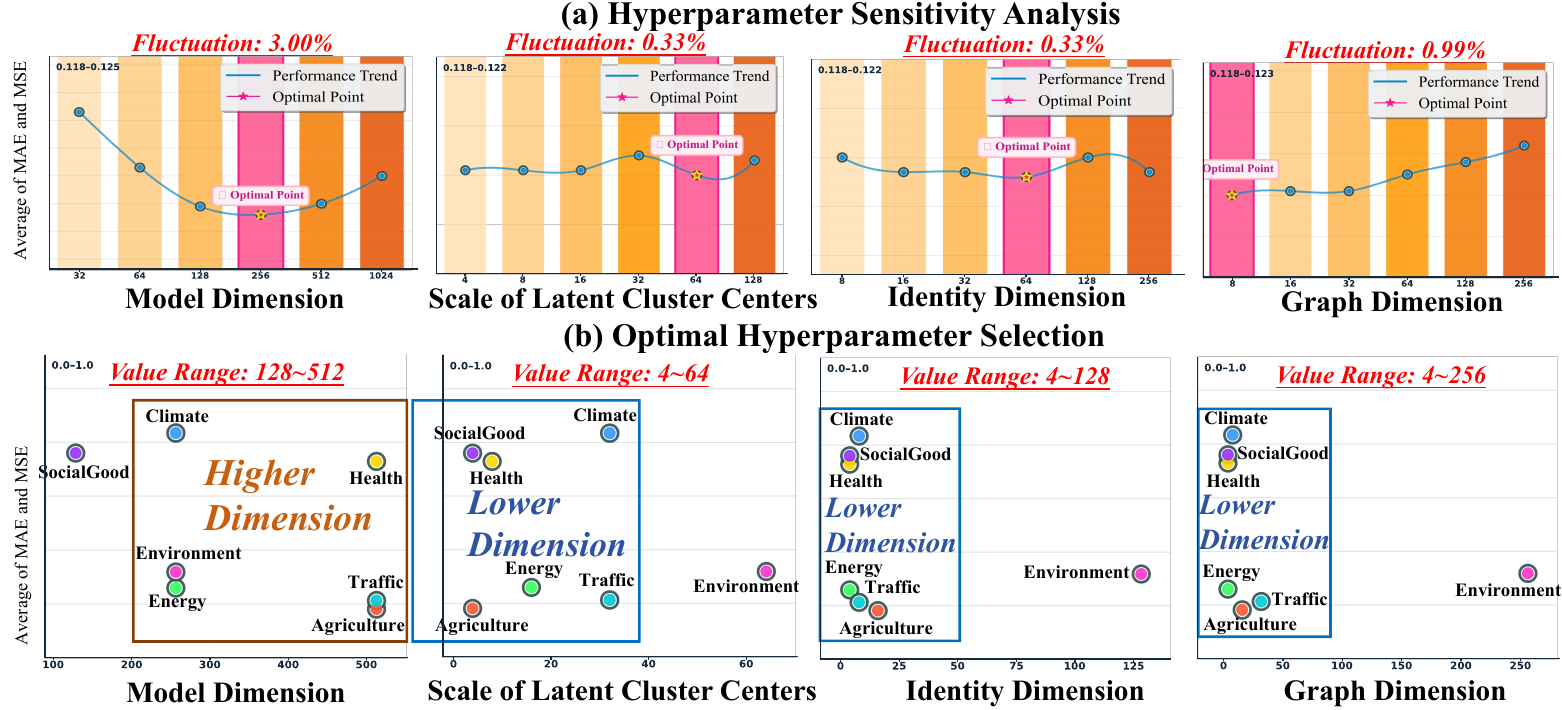}
\caption{Hyperparameter analysis. (a): Sensitivity analysis (PEMS04, Input-96-predict-12); (b): Optimal Selection (Multimodal, Input-8-predict-12)}
\label{figure-hyper-vis}
\end{figure}

From observations in Fig. \ref{figure-hyper-vis}(b), Model Dimension exhibits larger value generally, which may require greater model capacity to mine effective representations. In contrast, adopting more lightweight core dimensions for Identity and Graph helps suppress cross-modal noise. However, scaling the dimensionality can amplify capacity during extensive time steps conditions (e.g., Environment).

\subsubsection{Text Encoder Analysis} 
\begin{figure}
\centering
\includegraphics[width=1.0 \linewidth]{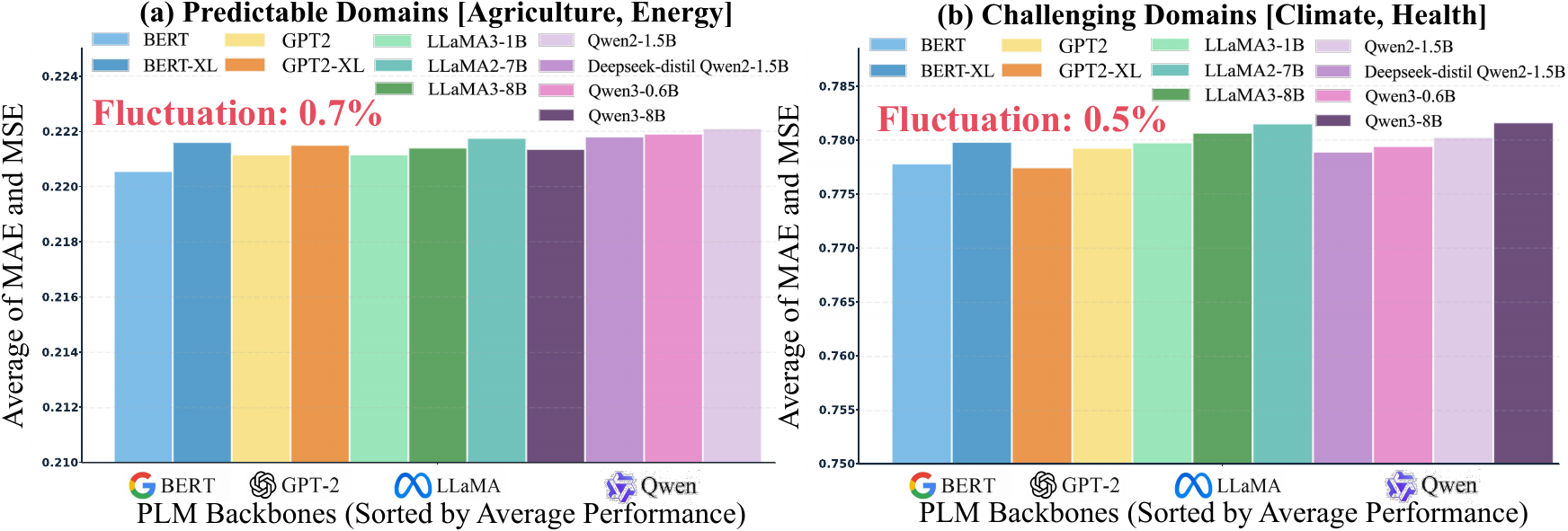}
\caption{Text encoder evaluation.  (a): Easily predictable domains; (b): Challenging domains. (Embedding dimension: [BERT:768, BERT-XL:1024, GPT2:768, GPT2-XL:1600, LLaMA3.2-1B:2048, LLaMA2-7B/LLaMA3-8B:4096, Qwen3-0.6B:1024, Qwen2/Deepseek-distil-1.5B:1536, Qwen3-8B:4096], Input-8-predict-12). The contribution of PLMs remains inconclusive}
\label{figure-plm}
\end{figure}

Despite our experimental setup employing efficient BERT-style PLMs \cite{devlin2019bert}, a comprehensive comparison of text encoders remains necessary. Therefore, we switch to typical LLM backbones (i.e., GPT2 \cite{radford2019language}, LLaMA \cite{touvron2023llama,grattafiori2024llama}, Qwen \cite{team2024qwen2,yang2025qwen3} and Deepseek-distil-Qwen2 \cite{Guo2025}). Observations from the Fig. \ref{figure-plm} in "multimodal" scenario indicate that smaller BERT and GPT2 maintain the robustest performance in both predictable and challenging domains. This finding can be attributed to lower embedding dimension for fine-tuning effective projection within limited data \cite{liu2024time}, while the prompt capabilities of Qwen and distilled reasoning of DeepSeek-R1 only translate to alignment paradigm partially. Moreover, the minimal variation \cite{li2025language} (0.5\% $\sim$ 0.7\%) among different PLMs for textual encoding highlights MIDAPN's robustness in media-aware pre-alignment setting.   

\subsection{Efficiency Analysis}
\begin{figure}
\centering
\includegraphics[width=1.0 \linewidth]{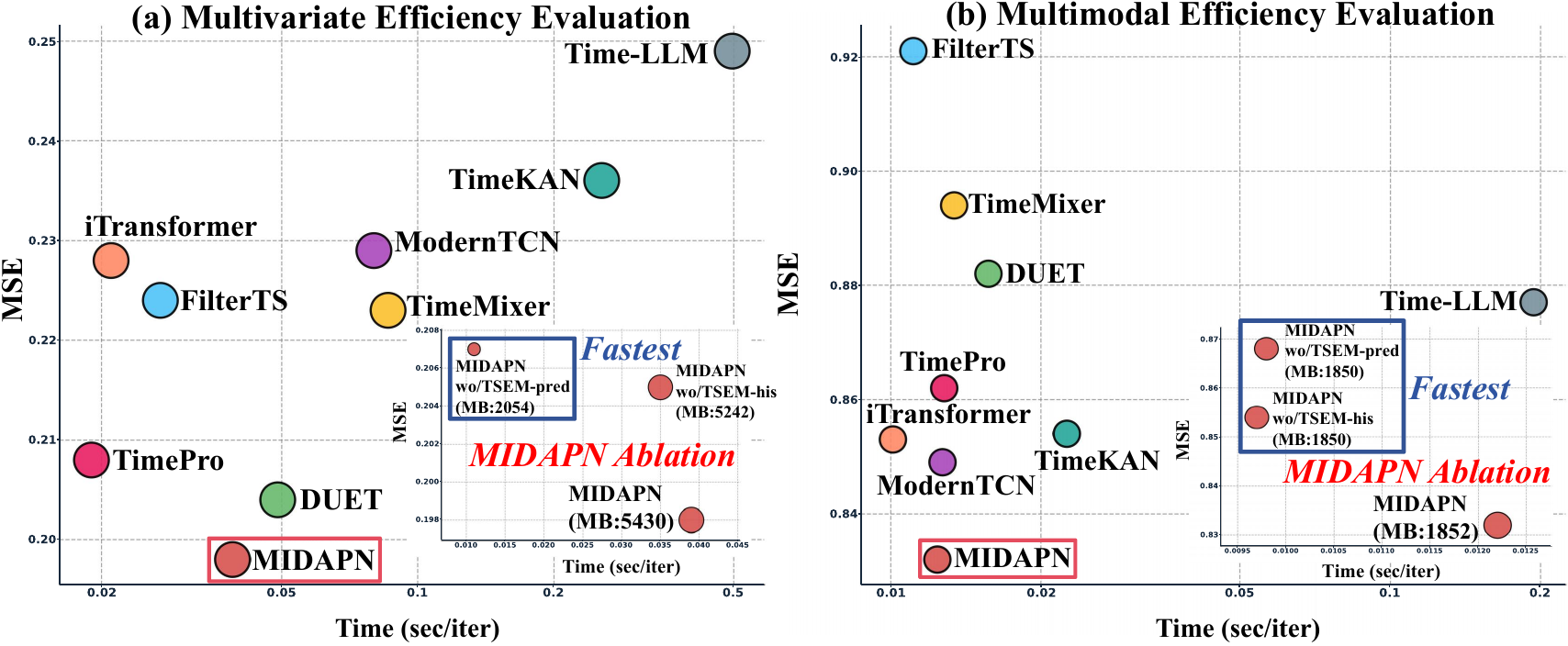}
\caption{Performance, efficiency and memory footprint comparison. (a): Input-96-predict-720 for Electricity; (b): Input-8-predict-12 for Health. By selectively ablating $\text{TSEM}_{\text{his}}$ and $\text{TSEM}_{\text{pred}}$, MIDAPN can achieve the most expeditious training speed compared to other SOTA TSF models}
\label{figure-eff-vis}
\end{figure}

We report the MSE and per-iteration training time of each TSF model. As shown in Fig. \ref{figure-eff-vis}, (1) MIDAPN achieves the lowest error while remaining the third- or fourth-fastest in both "multivariate" and "multimodal" settings. Although iTransformer and FilterTS are faster, MIDAPN offers better accuracy. In fact, small identity and graph dimensions keep $D$ manageable in MIDAG's $O(C^2D)$ complexity, while SPConv uses lightweight TCN-style operations (i.e., localized connections \& parameter sharing) with spectral transforms complexity in $O(CT \log T)$. Notably, the ASGM search is completed at initialization with negligible $O(L)$ overhead, adding no training burden. Thus, MIDAPN balances accuracy and efficiency well. (2) Moreover, removing $\text{TSEM}_{\text{his}}$ or $\text{TSEM}_{\text{pred}}$ gives MIDAPN a clear speed advantage, enabling faster training than SOTA TSF models with only marginal performance loss. (3) For a TSEM block of length $T$, DFF
$\Psi_{\text{project}}$, MIDAG $\Psi_{\text{spatial}}$, and SPConv
$\Psi_{\text{temporal}}$ contribute $O(CT^2)$, $O(C^2D)$, and
$O(Ck\sum_{l=1}^{L}F_{l-1}D^{(l-1)}D^{(l)}
+CF_LD^{(L)}T)$, where $F_l$ and $D^{(l)}$
denote spectral length and expanded channel width.
Thus, MIDAPN costs
$\mathcal{C}_{\rm TSEM}(T_{\rm in})
+O(CT_{\rm in}T_{\rm rec})
+\mathcal{C}_{\rm TSEM}(T_{\rm rec})$,
while ASGM adds only $O(L)$ initialization overhead.

\section{Conclusion}
\label{conclu}
In this paper, we propose MIDAPN to realize a unified spatiotemporal backbone for media-bridged time series forecasting, which collaboratively comprises
two innovative components based on media-general graph structure and automatic
temporal learning: (1) Guided by contextual modulation, the Multimedia Identity-Aware Graph captures intrinsic spatio-temporal state evolution, while bridging the heterogeneity gap across multimedia through a dual mechanism of individual identity learning and group-level identity clustering; (2) “Adaptive Search-Guided” Spectral Prism Convolution automatically constructs a hierarchical compression-generation architecture via scale-efficient optimization, effectively capturing multi-scale dynamism in the frequency domain. Experiments in 25 real-world datasets demonstrate MIDAPN outperforms SOTA deep TSF, TSFM, and PLM-fused methods, providing a new perspective linking "multivariate" and narrative-flow "multimodal" for spatio-temporal dependencies and modality heterogeneity solving. For future research, we may extend to integrate perceptual modalities (e.g., images, audio and video).

\bibliographystyle{IEEEtran}
\bibliography{mybib.bib}

\section*{Appendix}

\section{Joint Conditional-Information View of CIM}
\label{app:cmi_cim}

Let $\mathcal{R}$ and $\mathcal{M}$ denote the residual and gated
responses in CIM, respectively, and let
$\mathcal{C}\triangleq\mathcal{Z}_{\mathrm{agg}}$ denote the
graph-aggregated context. Here, $[\mathcal{R},\mathcal{M}]$ denotes
their joint observation rather than additive fusion. By the
definition of conditional mutual information,
\begin{equation}
\begin{aligned}
&I([\mathcal{R},\mathcal{M}];\mathcal{Y}\mid\mathcal{C}) \\
={}&
H(\mathcal{Y}\mid\mathcal{C})
-
H(\mathcal{Y}\mid\mathcal{R},\mathcal{M},\mathcal{C}).
\end{aligned}
\label{eq:cmi_joint}
\end{equation}
Here, $H(\cdot\mid\cdot)$ denotes conditional entropy.
Adding and subtracting
$H(\mathcal{Y}\mid\mathcal{R},\mathcal{C})$ gives
\begin{equation}
\begin{aligned}
&I([\mathcal{R},\mathcal{M}];\mathcal{Y}\mid\mathcal{C}) \\
={}&
\underbrace{
H(\mathcal{Y}\mid\mathcal{C})
-
H(\mathcal{Y}\mid\mathcal{R},\mathcal{C})
}_{I(\mathcal{R};\mathcal{Y}\mid\mathcal{C})}
\\
&+
\underbrace{
H(\mathcal{Y}\mid\mathcal{R},\mathcal{C})
-
H(\mathcal{Y}\mid\mathcal{R},\mathcal{M},\mathcal{C})
}_{I(\mathcal{M};\mathcal{Y}\mid\mathcal{R},\mathcal{C})}.
\end{aligned}
\label{eq:cmi_chain_derivation}
\end{equation}
Therefore,
\begin{equation}
\begin{aligned}
I([\mathcal{R},\mathcal{M}];\mathcal{Y}\mid\mathcal{C})
={}&
I(\mathcal{R};\mathcal{Y}\mid\mathcal{C}) \\
&+
I(\mathcal{M};\mathcal{Y}\mid\mathcal{R},\mathcal{C}).
\end{aligned}
\label{eq:cmi_chain}
\end{equation}
Conditioned on $\mathcal{C}$, this induces the Markov relation
\begin{equation}
\mathcal{Y}
\longrightarrow
[\mathcal{R},\mathcal{M}]
\longrightarrow
X_{\mathrm{MID}}.
\end{equation}
The conditional data-processing inequality consequently yields
\begin{equation}
I(X_{\mathrm{MID}};\mathcal{Y}\mid\mathcal{C})
\leq
I([\mathcal{R},\mathcal{M}];\mathcal{Y}\mid\mathcal{C}).
\label{eq:cmi_dpi}
\end{equation}
Thus, the decomposition in Eq.~(\ref{eq:cmi_chain}) provides an
interpretation of the two CIM paths before fusion.

\section{Datasets Description}
\label{dd}

 We have summarized 25 public datasets from both "multivariate" and "multimodal" scenarios in Table \ref{tab:dataset_stats_wide}.

Note that LEU, MSPG, and PTF originally consist of dispersed
time-series--text pairs. We use Gemini 3.1 Pro to organize and
convert them into the TaTS-compatible format.

\begin{table*}[ht]
\centering
\caption{Statistics of Datasets and the Actual Use of RevIN/PRReg for MIDAPN}
\label{tab:dataset_stats_wide}
\resizebox{0.9\linewidth}{!}{
\begin{tabular}{lcccccll} 
\toprule[1.5pt]
\textbf{Datasets} & \textbf{Variates} &\textbf{RevIN} &\textbf{PRReg} & \textbf{Timestamps} & \textbf{Frequency} & \textbf{Properties} &\textbf{Date}\\
\midrule
 ETTm2  & 7 &$\boldsymbol{\times}$ &$\boldsymbol{\checkmark}$ & 69680 & 15mins  & Volatility \& Periodicity &2016.7.1-2018.6.26 \\
 ETTh2  & 7 &$\boldsymbol{\times}$ &$\boldsymbol{\checkmark}$ & 17420  & 1h  & Volatility \& Periodicity &2016.7.1-2018.6.26 \\
 Flight  & 7 &$\boldsymbol{\checkmark}$ &$\boldsymbol{\times}$ & 26304  & 1h  & Volatility & 2019.1.1-2021.12.31\\
 Weather  & 21 &$\boldsymbol{\checkmark}$ &$\boldsymbol{\times}$ & 52696  & 10mins & Fluctuation & 2020.1.1-2021.1.1\\
 Traffic  & 862 &$\boldsymbol{\checkmark}$ &$\boldsymbol{\times}$ & 17544  & 1h  & Stable Periodicity &2016.7.1-2018.7.2 \\
 Electricity  & 321 &$\boldsymbol{\checkmark}$ &$\boldsymbol{\times}$ & 26304  & 1h & Stable Periodicity &2016.7.1-2019.7.2 \\
 Solar  & 137 &$\boldsymbol{\times}$ &$\boldsymbol{\times}$ & 35040  & 15mins  & Stable Periodicity &2006.1.1-2006.12.31 \\
 PEMS03  & 358 &$\boldsymbol{\times}$ &$\boldsymbol{\times}$ & 26208  & 5mins & Unstable Periodicity &2018.9.1 - 2018.11.30 \\
 PEMS04  & 307 &$\boldsymbol{\times}$ &$\boldsymbol{\times}$ & 16992  & 5mins & Unstable Periodicity &2018.1.1-2018.2.28 \\
 PEMS07  & 883 &$\boldsymbol{\times}$ &$\boldsymbol{\times}$ & 28224 & 5mins & Unstable Periodicity & 2017.5.1-2017.8.31\\
 PEMS08  & 170 &$\boldsymbol{\times}$ &$\boldsymbol{\times}$ & 17856 & 5mins & Unstable Periodicity &2016.7.1-2016.8.31 \\
 ILI  & 7 &$\boldsymbol{\checkmark}$ &$\boldsymbol{\checkmark}$  & 966 & 1week  & Transition \& Periodicity &2002.1.1-2020.6.30 \\
 NASDAQ & 12 &$\boldsymbol{\checkmark}$ &$\boldsymbol{\checkmark}$ & 3914 & 1day  & Volatility & 2010.1.4-2024.10.25 \\
\midrule
 Agriculture  & \multirow{8}{3cm}{\centering 1 (time series) \\ +12 (text)} &$\boldsymbol{\checkmark}$ &$\boldsymbol{\checkmark}$ & 496 & Monthly  & Trend \& Transition &1983.1.1-2024.4.30 \\
 Climate  & &$\boldsymbol{\checkmark}$ &$\boldsymbol{\checkmark}$ & 496  & Monthly & Volatility \& Periodicity &1983.1.1-2024.4.30 \\
   Economy  & &$\boldsymbol{\times}$ &$\boldsymbol{\checkmark}$ & 423  & Monthly & Volatility \& Transition &1989.1.1-2024.4.30 \\
 Energy  & &$\boldsymbol{\checkmark}$ &$\boldsymbol{\times}$ & 1479  & Weekly  & Trend \& Transition & 1996.1.1-2024.5.5\\
 Environment  & &$\boldsymbol{\checkmark}$ &$\boldsymbol{\checkmark}$ & 11102  & Daily  & Volatility \& Periodicity &1982.1.1-2023.9.30 \\
 Health  & &$\boldsymbol{\times}$ &$\boldsymbol{\checkmark}$ & 1389  & Weekly & Volatility \& Periodicity &1997.9.29-2024.5.12 \\
 Security  & &$\boldsymbol{\checkmark}$ &$\boldsymbol{\times}$ & 297  & Monthly & Mutant \& Fluctuation &1999.9.1-2024.5.31 \\
 SocialGood  & &$\boldsymbol{\times}$ &$\boldsymbol{\checkmark}$ & 900  & Monthly & Transition \& Periodicity &1950.1.1-2024.12.31 \\
 Traffic  & &$\boldsymbol{\checkmark}$ &$\boldsymbol{\checkmark}$ & 531  & Monthly & Trend \& Periodicity & 1980.1.1-2024.3.31\\
 \midrule
  MSPG  &  27 (time series) +12 (text) &$\boldsymbol{\times}$ &$\boldsymbol{\checkmark}$ & 37536 & 15mins  & Stable Periodicity &2021.3.6-2022.3.31 \\
  LEU  &  16 (time series) +12 (text) &$\boldsymbol{\times}$ &$\boldsymbol{\checkmark}$ & 34848 & 30mins  & Trend \& Transition &2012.1.6-2013.12.31 \\
  PTF  &  32 (time series) +12 (text) &$\boldsymbol{\times}$ &$\boldsymbol{\checkmark}$ & 8640 & 1h  & Trend \& Transition &2022.1.6-2022.12.31 \\
\bottomrule[1.5pt]
\end{tabular}
}
\end{table*}

\section{Baselines} 
\label{baseline}
\subsection{Basic TSF Models}

\noindent \textit{(1) Transformer-based:} PatchTST (2023) learns patterns from patched subsequences within a channel-independent strategy; iTransformer (2024) enhances variable interaction by inverting the self-attention dimension; DUET (2025) boosts forecasting via masked spatio-temporal clustering in frequency domain.

\noindent \textit{(2) MLP-based:} TimeMixer (2024) fuses multi-scale patterns from decomposed and down-sampled time series; FilterTS (2025) synergistically learns stationary and dynamic frequency components; CrossLinear (2025) captures time-invariant and direct variable dependencies through lightweight cross-correlation embedding; SEMixer (2026) fuses multi-scale patterns from Random Attention Mechanism and Progressive Mixing Chain, while CPiRi (2026) serves as a plug-in for channel permutation invariant regularization.

\noindent \textit{(3) TCN-based:} TimesNet (2023) reshapes 1D series into a 2D representation for intra- and inter-period dynamics learning; ModernTCN (2024) uses depthwise and grouped pointwise convolutions to learn cross-variable and temporal dependencies. VPNet (2026) builds on the local sufficiency hypothesis and introduces variate–patch field and TCNBlock to balance accuracy and efficiency in high-dimensional settings.

\noindent \textit{(4) RNN-based:} P-sLSTM (2025) integrates exponential gating, memory mixing, patching-channel independence.

\noindent \textit{(5) GCN-based:} MSGNet (2024) constructs multiple graph structures within and between key frequency-domain scales to self-learn inter-series correlations. TimeFilter (2025) employs graph neural network and dynamic routing mechanism based on the Mixture of Experts to preserve the most critical spatiotemporal correlations.

\noindent \textit{(6) LLM-based:} Time-LLM (2024) fine-tunes the PLM to align the natural language semantics with temporal patterns.

\noindent \textit{(7) KAN-based:} TimeKAN (2025) represents different frequencies by multi-order Kolmogorov-Arnold Networks.

\noindent \textit{(8) Mamba-based:} TimePro (2025) reconstructs variate and temporal hyper-states through the Mamba architecture.

\subsection{TSFM \& Multimodal PLM Frameworks}

14 latest TSFM and Multimodal PLM Frameworks for driving the general-purpose multimedia temporal modeling:

\noindent \textit{(1) TSFM-based.} Aurora (2026): The first generative multimedia TSFM on both multivariate and multimodal scenarios. GTM (2026): GTM integrates frequency-domain attention and hybrid masking into a task-agnostic generative framework for universal time series analysis. CoRA (2026): A correlation-aware adapter that requires only fine-tuning of the TSFM to capture relationships across different types. Sundial (2025): Introducing flow-matching idea into auto-regressive time series generation, enabling the base model to learn any complex distribution on continuous value domain. SEMPO (2025): A novel lightweight foundational model that integrates an energy-aware spectral decomposition module and a prompt hybridization Transformer to learn heterogeneous temporal patterns. MOIRAI (2024): A universal time series forecasting Transformer based on mask encoder.

\noindent \textit{(2) MLLM-based.} SE-LLM (2026): SE-LLM aligns temporal embeddings with LLM semantic spaces to separate anomalous and denormalized semantics, thereby enhancing the freezing effect of large language models through cross-modal temporal semantic alignment. Time-VLM  (2025): A multimodal framework that enhances prediction capabilities by integrating visual and textual representations of time series data with pre-trained visual-linguistic models. CALF (2025): By employing dual-branch cross-modal contrastive learning with temporal and text sources, it reduces the distribution differences between text and temporal data. DualSG (2025): The dual-stream architecture employs LLMs as semantic guides to refine traditional predictions, thereby avoiding the complex implicit alignment of latent spaces. ChatTime (2025): An out-of-the-box multimodal time series foundation model, which innovatively characterizes time series as a foreign language for LLM adaptation. GPT4MTS (2024): This model is a prompt-based LLM framework that enables multimodal time series prediction by simultaneously utilizing numerical data and textual information.

\noindent \textit{(3) Multimodal Framework.} VoT (2026) employs event-driven reasoning and multi-level alignment to leverage rich information from exogenous texts. MM-TSLib (2024): A framework based on weighted alignment of dual-modality branches.

\subsection{Implementation Details} 

\subsubsection{Settings} All experiments were executed on NVIDIA GeForce RTX 4090 and A100-SXM4 GPUs. For the multivariate scenario, we conformed the Time-series Library . Within the multimodal scenario that employs the TaTS framework , we used classic bert-base-uncased  as text encoder. This choice reflects the trade-off between performance gain and computational cost  (i.e., Occam's Razor principle for simplicity and effectiveness), making it straightforward to assess the true capability of TSF models. Meanwhile the textual dimension $d_{\text{tproj}}$ was empirically set to 12 to maintain consistency . The training, validation, and test sets were split according to standards in previous research. Specifically, the ETT\{m2, h2\} and PEMS\{03, 04, 07, 08\} datasets were 6:2:2 ratio, while the others were 7:1:2 ratio. To ensure a uniform comparison, we cited results from the reports of prior work . For any missing results or inconsistent settings , we reran them within moderately optimized configuration, while the resilient L1 loss serves as an optional guide for training under highly volatile conditions (i.e. QPS/ILI/PEMS). To mitigate sample indistinguishability , we optionally use the normalization or predict residuals method for baselines based on performance. For example, in some scenarios with frequent transitions (i.e., Health and Social), adopting PRReg instead of RevIN enables backtracking to the nearest known history for improvements.

\subsubsection{Model Configuration} The default hyperparameters for MIDAPN were set as: 2 Multimedia Identity-Aware Graph layers, 2 Spectral Prism Convolution layers, 512 model dimension $d_{m}$, 32 latent cluster centers $M$, 32 graph dimension $d_{\text{graph}}$, and 8 embedding dimension for both identity and latent cluster centers $d_{\text{id}}$, $d_{\text{cluster}}$. A limited search was also conducted for key training parameters in "multivariate", including a learning rate in $1 e^{- 4}$ $\sim$ $1 e^{- 2}$ with adaptive schedules \{exponential decay, cosine annealing\} for Adam optimizer, a batch size in \{4, 8, 16, 32\}, and training epoch in 10 $\sim$ 100. To ensure equitable exploration of "multimodal", we unify the initial learning rate in $3 e^{- 4}$ within exponential decay, training epochs in 10 for all models. In addition, the historical timestamps $T_{\text{mark}}$ were added as covariates for MIDAPN in QPS (multivariate) and Agriculture/Traffic/SocialGood (multimodal) scenes.

\subsubsection{Forecasting Configuration} For "multivariate" sub-task, the long-term prediction horizons were \{96, 192, 336, 720\}, while the short-term horizons were \{12, 24, 48, 96\} for PEMS\{03, 04, 07, 08\} and \{24, 36, 48, 60\} for ILI and NASDQ; For "multimodal" sub-task, the future horizons were \{8, 10, 12\}. This setting is only in accordance with relative time steps and real-world timestamp granularity is disregarded.

\section{Discussion of Auxiliary Training Objectives for Multimedia Scenarios}
\label{noloss}

To explain why we do not introduce an auxiliary training objective for the
multimedia identity-aware graph in the main model, we examine an
assignment-sharpening objective commonly considered in graph optimization:
$
\mathcal{L}_{\mathrm{assign}}
=
-\frac{1}{C}
\sum_{i=1}^{C}
\sum_{j=1}^{M}
\vartheta_{i,j}^{\mathrm{cluster}}
\log\left(
\vartheta_{i,j}^{\mathrm{cluster}}+\epsilon
\right)$,
where $\vartheta_{i,j}^{\mathrm{cluster}}$ denotes the probability of assigning
the $i$-th variable to the $j$-th representative cluster. When this auxiliary
objective is included, the total training objective becomes
$
\mathcal{L}_{\mathrm{total}}
=
\mathcal{L}_{\mathrm{task}}
+
\lambda\mathcal{L}_{\mathrm{assign}}$,
where $\lambda$ controls the auxiliary objective.

Minimizing $\mathcal{L}_{\mathrm{assign}}$ reduces the entropy of the
assignment distribution and encourages a sharp, nearly one-hot assignment.
However, such an explicit assignment constraint may weaken the overlapping
effects of different representative clusters. For example, in multimodal
scenarios, temporal sequences and their textual descriptions may provide
complementary information. Similarly, in multi-domain scenarios, an anomalous
sensor may simultaneously affect multiple sensor regions.  We consequently omit this auxiliary constraint from the main model and learn the multimedia identity-aware graph directly through task supervision. Importantly, $G^{\mathrm{aware}}$ remains sample-dependent and is dynamically updated end-to-end through $\mathcal{L}_{\mathrm{task}}$ throughout training. Table~\ref{tab:auxiliary_loss_comparison} compares the model without the
auxiliary objective against its variants using $\lambda=0.001$ and
$\lambda=0.01$. The assignment-sharpening objective provides no consistent
improvement, supporting our choice to learn the
identity-aware graph using task supervision.

\begin{table}[!t]
\centering
\caption{
Performance comparison with the auxiliary assignment.
\textcolor{red}{\textbf{Red}}/\textcolor{blue}{\textbf{Blue}}
indicate
\textcolor{red}{\textbf{1st}}/\textcolor{blue}{\textbf{2nd}}
}
\label{tab:auxiliary_loss_comparison}

\setlength{\tabcolsep}{9pt}
\renewcommand{\arraystretch}{1.08}

\resizebox{0.88\columnwidth}{!}{%
\begin{tabular}{lcccccc}
\toprule
& \multicolumn{2}{c}{\textbf{MIDAPN}-nonloss}
& \multicolumn{2}{c}{\textbf{MIDAPN}-loss-0.001}
& \multicolumn{2}{c}{\textbf{MIDAPN}-loss-0.01} \\
\cmidrule(lr){2-3}
\cmidrule(lr){4-5}
\cmidrule(lr){6-7}

\textbf{Dataset}
& \textbf{MSE} & \textbf{MAE}
& \textbf{MSE} & \textbf{MAE}
& \textbf{MSE} & \textbf{MAE} \\
\midrule

Agriculture
& \best{0.1314}
& \best{0.2317}
& \secondbest{0.1315}
& \secondbest{0.2320}
& 0.1323
& 0.2334 \\

Climate
& \secondbest{0.9061}
& \secondbest{0.7618}
& \best{0.9059}
& \best{0.7617}
& 0.9116
& 0.7638 \\

Energy
& \best{0.1834}
& \best{0.3359}
& 0.1968
& 0.3436
& \secondbest{0.1877}
& \secondbest{0.3407} \\

Environment
& \secondbest{0.2691}
& \best{0.3677}
& 0.2694
& 0.3719
& \best{0.2691}
& \secondbest{0.3682} \\

Health
& \best{0.8322}
& \best{0.6260}
& \secondbest{0.8365}
& \secondbest{0.6287}
& 0.8716
& 0.6423 \\

\midrule

ETTm2
& \secondbest{0.2733}
& \best{0.3160}
& 0.2763
& 0.3180
& \best{0.2731}
& \secondbest{0.3172} \\

Flight
& \best{0.1620}
& \best{0.2630}
& \secondbest{0.1625}
& 0.2648
& 0.1630
& \secondbest{0.2645} \\

Illness
& \best{1.9335}
& \best{0.8515}
& 2.0906
& 0.8787
& \secondbest{2.0394}
& \secondbest{0.8660} \\

Electricity
& \best{0.1673}
& \best{0.2598}
& \secondbest{0.1685}
& \secondbest{0.2604}
& 0.1704
& 0.2628 \\

\bottomrule
\end{tabular}%
}
\end{table}

\begin{table}[!t]
  \centering
  \caption{Forecasting results over five random seeds}
  \label{tab:five_seeds}
  \resizebox{0.8\columnwidth}{!}{%
    \begin{tabular}{llcc}
      \toprule
      \textbf{Dataset}
      & \textbf{Model}
      & \textbf{MSE}
      & \textbf{MAE} \\
      \midrule

      \multicolumn{4}{l}{\textit{Multivariate datasets}} \\
      \addlinespace[2pt]

      ETTh2
      & MIDAPN
      & $0.368 \pm 0.003$
      & $0.392 \pm 0.002$ \\

      ETTm2
      & MIDAPN
      & $0.271 \pm 0.001$
      & $0.316 \pm 0.001$ \\

      Flight
      & MIDAPN
      & $0.161 \pm 0.000$
      & $0.262 \pm 0.000$ \\

      Weather
      & MIDAPN
      & $0.236 \pm 0.001$
      & $0.260 \pm 0.001$ \\

      PEMS03
      & MIDAPN
      & $0.096 \pm 0.001$
      & $0.201 \pm 0.001$ \\

      PEMS04
      & MIDAPN
      & $0.086 \pm 0.001$
      & $0.190 \pm 0.002$ \\

      \midrule
      \multicolumn{4}{l}{\textit{Multimodal datasets}} \\
      \addlinespace[2pt]

      \multirow{2}{*}{Agriculture}
      & MIDAPN
      & \best{$0.106 \pm 0.001$}
      & \best{$0.212 \pm 0.003$} \\
      & VPNet
      & \secondbest{$0.107 \pm 0.001$}
      &  \secondbest{$0.216 \pm 0.003$} \\

      \addlinespace[2pt]
      \multirow{2}{*}{Climate}
      & MIDAPN
      & \best{$0.905 \pm 0.009$}
      & \best{$0.764 \pm 0.005$} \\
      & VPNet
      & \secondbest{$0.917 \pm 0.004$}
      & \secondbest{$0.769 \pm 0.002$} \\

      \addlinespace[2pt]
      \multirow{2}{*}{Energy}
      & MIDAPN
      & \best{$0.177 \pm 0.003$}
      & \best{$0.327 \pm 0.002$} \\
      & VPNet
      & \secondbest{$0.179 \pm 0.002$}
      & \secondbest{$0.330 \pm 0.001$} \\

      \bottomrule
    \end{tabular}%
  }
\end{table}

\begin{table}[t]
  \centering
  \caption{Performance under different textual dimensions}
  \label{tab:text_emb_average}
  \resizebox{0.8\columnwidth}{!}{%
    \begin{tabular}{llcccccc}
      \toprule
      \multirow{2}{*}{\textbf{Dataset}}
      & \multirow{2}{*}{\textbf{Metric}}
      & \multicolumn{6}{c}{\textbf{Text-Embedding Dimension}} \\
      \cmidrule(lr){3-8}
      & & \textbf{4} & \textbf{8} & \textbf{12}
        & \textbf{24} & \textbf{48} & \textbf{96} \\
      \midrule

      \multirow{2}{*}{Climate}
      & MSE & 0.910 & 0.910 & \textbf{0.890}
            & 0.905 & 0.906 & 0.907 \\
      & MAE & 0.766 & 0.765 & \textbf{0.757}
            & 0.763 & 0.763 & 0.764 \\
      \addlinespace

      \multirow{2}{*}{Energy}
      & MSE & 0.182 & 0.176 & \textbf{0.172}
            & 0.178 & 0.178 & 0.177 \\
      & MAE & 0.330 & 0.327 & \textbf{0.324}
            & 0.327 & 0.328 & 0.327 \\
      \addlinespace

      \multirow{2}{*}{Environment}
      & MSE & 0.269 & 0.268 & \textbf{0.267}
            & 0.268 & 0.270 & 0.270 \\
      & MAE & 0.369 & 0.369 & \textbf{0.366}
            & 0.368 & 0.368 & 0.370 \\
      \addlinespace

      \multirow{2}{*}{Traffic}
      & MSE & 0.167 & 0.178 & \textbf{0.163}
            & 0.176 & 0.177 & 0.176 \\
      & MAE & 0.237 & 0.244 & \textbf{0.229}
            & 0.250 & 0.252 & 0.256 \\
      \addlinespace

      \multirow{2}{*}{Health}
      & MSE & 0.788 & 0.774 & \textbf{0.732}
            & 0.789 & 0.791 & 0.846 \\
      & MAE & 0.593 & 0.588 & \textbf{0.573}
            & 0.590 & 0.595 & 0.614 \\

      \bottomrule
    \end{tabular}%
  }
\end{table}

\section{Result Robustness}
\label{rr}

To assess robustness across multimodal and multivariate settings, we report
the mean and standard deviation over five random seeds in
Table~\ref{tab:five_seeds}. For the multimodal datasets, we additionally
compare MIDAPN with VPNet, and each seed's results are averaged over
prediction horizons $\{8,10,12\}$ (without searching the best hyperparameter setting for each seed, unlike comparison). These results improve transparency and
quantify performance variability. Meanwhile, we also search the TaTS's optimal textual embedding in Table~\ref{tab:text_emb_average}, which suggests that a moderate embedding dimension provides a favorable representation capacity for modal alignment.

\section{Full Results}
\label{full}

please see Table \ref{tab:comprehensive_long_term_updated} (long-term, multivariate), Table \ref{tab:comprehensive_short_term_updated} (short-term, multivariate), Table \ref{tab:comprehensive_horizons_all} (multimodal), Table \ref{tab:weather_electricity_comparison} (TSFM \& PLM: multivariate), Table \ref{tab:multi_domain_comparison} (TSFM \& PLM: multimodal) for details.

 download this file. Check that its computer is connected and try again
\begin{table*}[!htbp]
\centering
\caption{Comprehensive performance comparison in long-term forecasting. Results are highlighted strictly following the Dense Ranking principle. \textcolor{red}{\textbf{Red}}/\textcolor{blue}{\textbf{Blue}} indicate \textcolor{red}{\textbf{1st}}/\textcolor{blue}{\textbf{2nd}}}
\label{tab:comprehensive_long_term_updated}
\resizebox{\linewidth}{!}{%
\begin{tabular}{@{}cc *{34}{c}@{}}
\toprule

\multicolumn{2}{c}{Models}
& \multicolumn{2}{c}{MIDAPN}
& \multicolumn{2}{c}{TimeFilter}
& \multicolumn{2}{c}{VPNet}
& \multicolumn{2}{c}{CPiRi}
& \multicolumn{2}{c}{SEMixer}
& \multicolumn{2}{c}{DUET}
& \multicolumn{2}{c}{TimeKAN}
& \multicolumn{2}{c}{FilterTS}
& \multicolumn{2}{c}{TimePro}
& \multicolumn{2}{c}{P\_sLSTM}
& \multicolumn{2}{c}{ModernTCN}
& \multicolumn{2}{c}{TimeMixer}
& \multicolumn{2}{c}{iTransformer}
& \multicolumn{2}{c}{TimeLLM}
& \multicolumn{2}{c}{MGSNET}
& \multicolumn{2}{c}{PatchTST}
& \multicolumn{2}{c}{TimesNet} \\

\multicolumn{2}{c}{(Years)}
& \multicolumn{2}{c}{(Ours)}
& \multicolumn{2}{c}{(2025)}
& \multicolumn{2}{c}{(2026)}
& \multicolumn{2}{c}{(2026)}
& \multicolumn{2}{c}{(2026)}
& \multicolumn{2}{c}{(2025)}
& \multicolumn{2}{c}{(2025)}
& \multicolumn{2}{c}{(2025)}
& \multicolumn{2}{c}{(2025)}
& \multicolumn{2}{c}{(2025)}
& \multicolumn{2}{c}{(2024)}
& \multicolumn{2}{c}{(2024)}
& \multicolumn{2}{c}{(2024)}
& \multicolumn{2}{c}{(2024)}
& \multicolumn{2}{c}{(2024)}
& \multicolumn{2}{c}{(2023)}
& \multicolumn{2}{c}{(2023)} \\

\cmidrule(lr){3-4}
\cmidrule(lr){5-6}
\cmidrule(lr){7-8}
\cmidrule(lr){9-10}
\cmidrule(lr){11-12}
\cmidrule(lr){13-14}
\cmidrule(lr){15-16}
\cmidrule(lr){17-18}
\cmidrule(lr){19-20}
\cmidrule(lr){21-22}
\cmidrule(lr){23-24}
\cmidrule(lr){25-26}
\cmidrule(lr){27-28}
\cmidrule(lr){29-30}
\cmidrule(lr){31-32}
\cmidrule(lr){33-34}
\cmidrule(lr){35-36}

\multicolumn{2}{c}{Metric}
& MSE & MAE & MSE & MAE & MSE & MAE & MSE & MAE
& MSE & MAE & MSE & MAE & MSE & MAE & MSE & MAE
& MSE & MAE & MSE & MAE & MSE & MAE & MSE & MAE
& MSE & MAE & MSE & MAE & MSE & MAE & MSE & MAE
& MSE & MAE \\
\midrule

% ETTm2
\multirow{5}{*}{ETTm2}
& 96 & \best{0.168} & \best{0.250} & \secondbest{0.169} & 0.255 & 0.173 & \secondbest{0.254} & 0.178 & 0.259 & 0.176 & 0.262 & 0.174 & 0.255 & 0.174 & 0.255 & 0.172 & 0.255 & 0.178 & 0.260 & 0.181 & 0.267 & 0.175 & 0.256 & 0.174 & 0.258 & 0.180 & 0.264 & 0.178 & 0.262 & 0.177 & 0.262 & 0.175 & 0.259 & 0.187 & 0.267 \\
& 192 & \best{0.230} & \best{0.294} & \secondbest{0.235} & 0.299 & 0.247 & 0.305 & 0.245 & 0.305 & 0.243 & 0.310 & 0.243 & 0.302 & 0.239 & 0.299 & 0.237 & 0.299 & 0.242 & 0.303 & 0.250 & 0.312 & 0.239 & \secondbest{0.298} & 0.239 & 0.302 & 0.250 & 0.309 & 0.243 & 0.304 & 0.247 & 0.307 & 0.241 & 0.302 & 0.249 & 0.309 \\
& 336 & \secondbest{0.295} & \best{0.330} & \best{0.293} & \secondbest{0.336} & 0.322 & 0.349 & 0.306 & 0.344 & 0.306 & 0.351 & 0.304 & 0.341 & 0.301 & 0.340 & 0.299 & 0.398 & 0.303 & 0.342 & 0.315 & 0.353 & 0.298 & \secondbest{0.336} & 0.296 & 0.340 & 0.311 & 0.348 & 0.308 & 0.345 & 0.312 & 0.346 & 0.305 & 0.343 & 0.321 & 0.351 \\
& 720 & \secondbest{0.391} & \best{0.389} & \best{0.390} & \secondbest{0.393} & 0.430 & 0.412 & 0.403 & 0.401 & 0.407 & 0.416 & 0.399 & 0.397 & 0.395 & 0.396 & 0.397 & 0.394 & 0.400 & 0.399 & 0.418 & 0.413 & 0.401 & 0.395 & 0.393 & 0.397 & 0.412 & 0.407 & 0.408 & 0.407 & 0.414 & 0.403 & 0.402 & 0.400 & 0.408 & 0.403 \\
\cmidrule(lr){2-36}
& Avg & \best{0.271} & \best{0.316} & \secondbest{0.272} & \secondbest{0.321} & 0.293 & 0.330 & 0.283 & 0.327 & 0.283 & 0.335 & 0.280 & 0.324 & 0.277 & 0.323 & 0.276 & \secondbest{0.321} & 0.281 & 0.326 & 0.291 & 0.336 & 0.278 & \secondbest{0.321} & 0.276 & 0.324 & 0.288 & 0.332 & 0.284 & 0.330 & 0.288 & 0.330 & 0.281 & 0.326 & 0.291 & 0.333 \\
\midrule

% ETTh2
\multirow{5}{*}{ETTh2}
& 96 & \best{0.280} & \best{0.329} & \secondbest{0.283} & \secondbest{0.337} & 0.304 & 0.350 & 0.294 & 0.346 & 0.286 & 0.343 & 0.296 & 0.345 & 0.290 & 0.340 & 0.290 & 0.338 & 0.293 & 0.345 & 0.298 & 0.348 & 0.305 & 0.349 & 0.293 & 0.344 & 0.297 & 0.349 & 0.294 & 0.343 & 0.328 & 0.371 & 0.302 & 0.348 & 0.340 & 0.374 \\
& 192 & \secondbest{0.363} & \best{0.382} & \best{0.362} & 0.392 & 0.379 & 0.397 & 0.373 & 0.396 & 0.369 & 0.401 & 0.368 & \secondbest{0.389} & 0.375 & 0.392 & 0.374 & 0.390 & 0.367 & 0.394 & 0.376 & 0.396 & 0.389 & 0.401 & 0.374 & 0.396 & 0.380 & 0.400 & 0.378 & 0.393 & 0.402 & 0.414 & 0.388 & 0.400 & 0.402 & 0.414 \\
& 336 & \secondbest{0.403} & \best{0.415} & 0.404 & 0.424 & \best{0.402} & \secondbest{0.420} & 0.425 & 0.436 & 0.404 & 0.427 & 0.411 & 0.422 & 0.423 & 0.435 & 0.406 & \secondbest{0.420} & 0.419 & 0.431 & 0.419 & 0.430 & 0.407 & \secondbest{0.420} & 0.416 & 0.430 & 0.428 & 0.432 & 0.418 & 0.429 & 0.435 & 0.443 & 0.426 & 0.433 & 0.452 & 0.452 \\
& 720 & 0.410 & \best{0.431} & \best{0.407} & \secondbest{0.433} & \secondbest{0.409} & \secondbest{0.433} & 0.431 & 0.448 & 0.425 & 0.449 & 0.412 & 0.434 & 0.443 & 0.449 & 0.418 & 0.437 & 0.427 & 0.445 & 0.426 & 0.444 & 0.414 & 0.434 & 0.440 & 0.452 & 0.427 & 0.445 & 0.426 & 0.444 & 0.417 & 0.441 & 0.431 & 0.446 & 0.462 & 0.468 \\
\cmidrule(lr){2-36}
& Avg & \secondbest{0.364} & \best{0.389} & \secondbest{0.364} & 0.397 & 0.374 & 0.400 & 0.381 & 0.407 & \best{0.353} & \secondbest{0.390} & 0.372 & 0.398 & 0.383 & 0.404 & 0.372 & 0.396 & 0.377 & 0.403 & 0.380 & 0.405 & 0.379 & 0.401 & 0.381 & 0.406 & 0.383 & 0.407 & 0.379 & 0.402 & 0.396 & 0.417 & 0.387 & 0.407 & 0.414 & 0.427 \\
\midrule

% Flight
\multirow{5}{*}{Flight}
& 96 & \secondbest{0.145} & \best{0.249} & \secondbest{0.145} & \secondbest{0.254} & 0.148 & 0.259 & \best{0.144} & \secondbest{0.254} & 0.147 & 0.256 & 0.146 & 0.256 & 0.165 & 0.278 & 0.151 & 0.258 & \secondbest{0.145} & 0.255 & 0.158 & 0.270 & 0.153 & 0.259 & 0.156 & 0.259 & 0.159 & 0.271 & 0.164 & 0.277 & 0.183 & 0.301 & 0.148 & 0.263 & 0.237 & 0.350 \\
& 192 & \secondbest{0.147} & \best{0.250} & \secondbest{0.147} & \secondbest{0.253} & 0.150 & 0.259 & \best{0.146} & 0.255 & \best{0.146} & 0.254 & \best{0.146} & 0.254 & 0.165 & 0.277 & 0.151 & 0.256 & \secondbest{0.147} & 0.255 & 0.158 & 0.269 & 0.160 & 0.274 & 0.156 & 0.257 & 0.163 & 0.276 & 0.165 & 0.278 & 0.189 & 0.306 & \secondbest{0.147} & 0.260 & 0.224 & 0.337 \\
& 336 & 0.157 & \best{0.260} & 0.160 & 0.267 & \secondbest{0.156} & 0.264 & \best{0.155} & \secondbest{0.263} & \secondbest{0.156} & \secondbest{0.263} & 0.159 & 0.266 & 0.172 & 0.282 & 0.160 & 0.264 & 0.164 & 0.273 & 0.167 & 0.277 & 0.169 & 0.282 & 0.167 & 0.269 & 0.177 & 0.291 & 0.170 & 0.283 & 0.206 & 0.320 & 0.158 & 0.271 & 0.289 & 0.394 \\
& 720 & 0.194 & \best{0.290} & 0.192 & 0.297 & \best{0.189} & \secondbest{0.293} & 0.194 & 0.298 & \secondbest{0.191} & 0.294 & 0.202 & 0.307 & 0.212 & 0.317 & 0.199 & 0.298 & 0.198 & 0.303 & 0.202 & 0.308 & 0.204 & 0.311 & 0.210 & 0.311 & 0.222 & 0.331 & 0.218 & 0.324 & 0.253 & 0.358 & 0.197 & 0.305 & 0.310 & 0.408 \\
\cmidrule(lr){2-36}
& Avg & \secondbest{0.161} & \best{0.262} & \secondbest{0.161} & 0.268 & \secondbest{0.161} & 0.269 & \best{0.160} & 0.268 & \best{0.160} & \secondbest{0.267} & 0.163 & 0.271 & 0.179 & 0.289 & 0.165 & 0.269 & 0.164 & 0.272 & 0.171 & 0.281 & 0.172 & 0.282 & 0.172 & 0.274 & 0.180 & 0.292 & 0.179 & 0.291 & 0.208 & 0.321 & 0.281 & 0.326 & 0.265 & 0.372 \\
\midrule

% Weather
\multirow{5}{*}{Weather}
& 96 & \best{0.146} & \best{0.185} & \secondbest{0.153} & \secondbest{0.199} & 0.162 & 0.206 & 0.172 & 0.211 & 0.175 & 0.219 & 0.163 & 0.202 & 0.162 & 0.208 & 0.162 & 0.207 & 0.166 & 0.207 & 0.180 & 0.222 & 0.158 & 0.203 & 0.166 & 0.213 & 0.174 & 0.214 & 0.178 & 0.219 & 0.163 & 0.212 & 0.177 & 0.218 & 0.172 & 0.220 \\
& 192 & \best{0.196} & \best{0.235} & \secondbest{0.202} & \secondbest{0.246} & 0.209 & 0.251 & 0.216 & 0.252 & 0.222 & 0.260 & 0.218 & 0.252 & 0.207 & 0.249 & 0.209 & 0.252 & 0.216 & 0.254 & 0.226 & 0.261 & 0.208 & 0.250 & 0.209 & 0.251 & 0.221 & 0.254 & 0.229 & 0.262 & 0.212 & 0.254 & 0.225 & 0.259 & 0.219 & 0.261 \\
& 336 & \best{0.255} & \best{0.279} & \secondbest{0.260} & \secondbest{0.289} & 0.266 & 0.294 & 0.272 & 0.295 & 0.276 & 0.299 & 0.274 & 0.294 & 0.263 & 0.290 & 0.263 & 0.292 & 0.273 & 0.296 & 0.280 & 0.300 & 0.266 & 0.294 & 0.264 & 0.293 & 0.278 & 0.296 & 0.286 & 0.303 & 0.272 & 0.299 & 0.278 & 0.297 & 0.280 & 0.306 \\
& 720 & \secondbest{0.342} & \best{0.336} & \secondbest{0.342} & 0.341 & 0.346 & 0.347 & 0.350 & 0.345 & 0.351 & 0.347 & 0.349 & 0.343 & \best{0.338} & \secondbest{0.340} & 0.344 & 0.344 & 0.351 & 0.346 & 0.356 & 0.349 & 0.346 & 0.346 & \secondbest{0.342} & 0.343 & 0.358 & 0.349 & 0.358 & 0.350 & 0.350 & 0.348 & 0.354 & 0.348 & 0.365 & 0.359 \\
\cmidrule(lr){2-36}
& Avg & \best{0.235} & \best{0.259} & \secondbest{0.239} & \secondbest{0.269} & 0.246 & 0.275 & 0.253 & 0.276 & 0.256 & 0.281 & 0.251 & 0.273 & 0.243 & 0.272 & 0.244 & 0.274 & 0.251 & 0.276 & 0.261 & 0.283 & 0.245 & 0.273 & 0.245 & 0.275 & 0.258 & 0.278 & 0.263 & 0.284 & 0.249 & 0.278 & 0.259 & 0.281 & 0.259 & 0.287 \\
\midrule

% Traffic
\multirow{5}{*}{Traffic}
& 96 & 0.431 & 0.257 & \best{0.375} & \best{0.251} & 0.420 & 0.279 & 0.415 & 0.288 & 0.489 & 0.321 & 0.407 & \secondbest{0.252} & 0.629 & 0.354 & 0.448 & 0.309 & 0.408 & 0.269 & 0.428 & 0.275 & 0.474 & 0.318 & 0.462 & 0.285 & \secondbest{0.395} & 0.268 & 0.536 & 0.359 & 0.621 & 0.339 & 0.462 & 0.295 & 0.593 & 0.321 \\
& 192 & 0.444 & \secondbest{0.268} & \best{0.396} & \best{0.262} & 0.436 & 0.284 & 0.438 & 0.298 & 0.490 & 0.316 & 0.431 & \best{0.262} & 0.566 & 0.364 & 0.455 & 0.307 & 0.424 & 0.276 & 0.441 & 0.279 & 0.478 & 0.321 & 0.473 & 0.296 & \secondbest{0.417} & 0.276 & 0.530 & 0.354 & 0.636 & 0.374 & 0.466 & 0.296 & 0.617 & 0.336 \\
& 336 & 0.463 & 0.278 & \best{0.415} & \secondbest{0.271} & 0.452 & 0.292 & 0.456 & 0.307 & 0.503 & 0.320 & 0.456 & \best{0.269} & 0.587 & 0.358 & 0.472 & 0.313 & 0.444 & 0.287 & 0.455 & 0.286 & 0.489 & 0.325 & 0.498 & 0.296 & \secondbest{0.433} & 0.283 & 0.530 & 0.349 & 0.663 & 0.390 & 0.482 & 0.304 & 0.629 & 0.336 \\
& 720 & 0.504 & 0.302 & \best{0.449} & \best{0.291} & 0.484 & 0.310 & 0.491 & 0.329 & 0.536 & 0.338 & 0.509 & \secondbest{0.292} & 0.620 & 0.375 & 0.508 & 0.332 & 0.483 & 0.311 & 0.489 & 0.305 & 0.520 & 0.340 & 0.506 & 0.313 & \secondbest{0.467} & 0.302 & 0.569 & 0.371 & 0.721 & 0.420 & 0.514 & 0.322 & 0.640 & 0.350 \\
\cmidrule(lr){2-36}
& Avg & \best{0.163} & 0.229 & 0.181 & 0.228 & 0.181 & 0.233 & 0.177 & 0.227 & 0.184 & 0.236 & \secondbest{0.170} & 0.234 & 0.186 & 0.253 & 0.177 & 0.246 & 0.177 & \best{0.222} & 0.178 & \secondbest{0.224} & 0.177 & 0.228 & 0.484 & 0.297 & 0.178 & 0.225 & 0.180 & 0.230 & 0.185 & 0.242 & 0.174 & \best{0.222} & 0.198 & 0.248 \\
\midrule

% Electricity
\multirow{5}{*}{Electricity}
& 96 & \best{0.138} & \best{0.231} & 0.143 & 0.237 & 0.152 & 0.245 & 0.147 & 0.239 & 0.167 & 0.256 & 0.144 & \secondbest{0.233} & 0.174 & 0.266 & 0.151 & 0.245 & \secondbest{0.139} & 0.234 & 0.161 & 0.249 & 0.152 & 0.254 & 0.157 & 0.250 & 0.148 & 0.240 & 0.182 & 0.262 & 0.165 & 0.274 & 0.159 & 0.268 & 0.168 & 0.272 \\
& 192 & 0.159 & 0.250 & \secondbest{0.158} & \secondbest{0.249} & 0.166 & 0.257 & 0.162 & 0.254 & 0.176 & 0.264 & 0.160 & \best{0.248} & 0.182 & 0.273 & 0.163 & 0.256 & \best{0.156} & \secondbest{0.249} & 0.171 & 0.258 & 0.171 & 0.269 & 0.177 & 0.263 & 0.162 & 0.253 & 0.194 & 0.275 & 0.184 & 0.292 & 0.177 & 0.278 & 0.184 & 0.289 \\
& 336 & \secondbest{0.172} & \best{0.265} & \best{0.171} & \best{0.265} & 0.183 & 0.274 & 0.176 & 0.269 & 0.193 & 0.280 & 0.173 & 0.270 & 0.197 & 0.286 & 0.180 & 0.274 & \secondbest{0.172} & \secondbest{0.267} & 0.188 & 0.277 & 0.189 & 0.288 & 0.189 & 0.282 & 0.178 & 0.269 & 0.211 & 0.294 & 0.195 & 0.302 & 0.195 & 0.296 & 0.198 & 0.300 \\
& 720 & \best{0.194} & \best{0.287} & \secondbest{0.203} & 0.296 & 0.221 & 0.306 & 0.212 & 0.302 & 0.233 & 0.313 & 0.204 & \secondbest{0.291} & 0.236 & 0.320 & 0.224 & 0.311 & 0.209 & 0.299 & 0.228 & 0.310 & 0.214 & 0.310 & 0.233 & 0.316 & 0.228 & 0.312 & 0.249 & 0.323 & 0.231 & 0.332 & 0.215 & 0.317 & 0.220 & 0.320 \\
\cmidrule(lr){2-36}
& Avg & \best{0.166} & \best{0.258} & \secondbest{0.169} & 0.262 & 0.181 & 0.271 & 0.174 & 0.266 & 0.192 & 0.278 & 0.170 & \secondbest{0.261} & 0.197 & 0.286 & 0.180 & 0.271 & \secondbest{0.169} & 0.262 & 0.187 & 0.274 & 0.182 & 0.280 & 0.189 & 0.278 & 0.179 & 0.269 & 0.209 & 0.289 & 0.194 & 0.300 & 0.187 & 0.290 & 0.193 & 0.295 \\
\midrule

% Solar
\multirow{5}{*}{Solar}
& 96 & \best{0.173} & \secondbest{0.221} & 0.198 & 0.244 & 0.193 & 0.229 & 0.190 & 0.250 & \secondbest{0.188} & 0.254 & 0.200 & \best{0.207} & 0.215 & 0.275 & 0.201 & 0.251 & 0.196 & 0.237 & 0.200 & 0.250 & 0.198 & 0.244 & 0.189 & 0.259 & 0.203 & 0.237 & 0.196 & 0.249 & 0.224 & 0.262 & 0.234 & 0.286 & 0.224 & 0.264 \\
& 192 & \best{0.198} & \best{0.233} & 0.209 & 0.258 & 0.209 & \secondbest{0.244} & 0.205 & 0.263 & \secondbest{0.204} & 0.261 & 0.228 & \best{0.233} & 0.252 & 0.297 & 0.237 & 0.282 & 0.231 & 0.263 & 0.240 & 0.277 & 0.230 & 0.258 & 0.222 & 0.283 & 0.233 & 0.261 & 0.225 & 0.266 & 0.279 & 0.301 & 0.267 & 0.310 & 0.264 & 0.310 \\
& 336 & \best{0.208} & \best{0.233} & 0.221 & 0.269 & 0.214 & 0.249 & 0.216 & 0.274 & \secondbest{0.210} & 0.264 & 0.262 & \secondbest{0.244} & 0.267 & 0.308 & 0.241 & 0.287 & 0.250 & 0.281 & 0.253 & 0.285 & 0.239 & 0.268 & 0.231 & 0.292 & 0.248 & 0.273 & 0.238 & 0.278 & 0.294 & 0.311 & 0.290 & 0.315 & 0.287 & 0.327 \\
& 720 & 0.221 & \best{0.236} & 0.224 & 0.272 & 0.222 & 0.252 & \best{0.212} & 0.267 & \secondbest{0.215} & 0.262 & 0.258 & \secondbest{0.249} & 0.265 & 0.307 & 0.250 & 0.292 & 0.253 & 0.285 & 0.259 & 0.288 & 0.242 & 0.270 & 0.223 & 0.285 & 0.249 & 0.275 & 0.238 & 0.280 & 0.279 & 0.325 & 0.289 & 0.317 & 0.267 & 0.305 \\
\cmidrule(lr){2-36}
& Avg & \best{0.200} & \best{0.231} & 0.213 & 0.261 & 0.210 & 0.244 & 0.206 & 0.264 & \secondbest{0.204} & 0.260 & 0.237 & \secondbest{0.233} & 0.250 & 0.297 & 0.232 & 0.278 & 0.232 & 0.266 & 0.238 & 0.275 & 0.227 & 0.260 & 0.216 & 0.280 & 0.233 & 0.262 & 0.224 & 0.268 & 0.269 & 0.300 & 0.270 & 0.307 & 0.261 & 0.302 \\
\midrule
\multicolumn{2}{l}{1\textsuperscript{st} Count}
& \multicolumn{2}{c}{\textbf{44}}
& \multicolumn{2}{c}{13}
& \multicolumn{2}{c}{2}
& \multicolumn{2}{c}{5}
& \multicolumn{2}{c}{3}
& \multicolumn{2}{c}{6}
& \multicolumn{2}{c}{1}
& \multicolumn{2}{c}{0}
& \multicolumn{2}{c}{2}
& \multicolumn{2}{c}{0}
& \multicolumn{2}{c}{0}
& \multicolumn{2}{c}{0}
& \multicolumn{2}{c}{0}
& \multicolumn{2}{c}{0}
& \multicolumn{2}{c}{0}
& \multicolumn{2}{c}{1}
& \multicolumn{2}{c}{0} \\

\bottomrule
\end{tabular}%
}
\end{table*}

\begin{table*}[!htbp]
\centering
\caption{Comprehensive performance comparison in short-term forecasting. Results are highlighted strictly following the Dense Ranking principle. \textcolor{red}{\textbf{Red}}/\textcolor{blue}{\textbf{Blue}} indicate \textcolor{red}{\textbf{1st}}/\textcolor{blue}{\textbf{2nd}}}
\label{tab:comprehensive_short_term_updated}
\resizebox{\linewidth}{!}{%
\begin{tabular}{@{}cc *{34}{c}@{}}
\toprule

\multicolumn{2}{c}{Models}
& \multicolumn{2}{c}{MIDAPN}
& \multicolumn{2}{c}{TimeFilter}
& \multicolumn{2}{c}{VPNet}
& \multicolumn{2}{c}{CPiRi}
& \multicolumn{2}{c}{SEMixer}
& \multicolumn{2}{c}{DUET}
& \multicolumn{2}{c}{TimeKAN}
& \multicolumn{2}{c}{FilterTS}
& \multicolumn{2}{c}{TimePro}
& \multicolumn{2}{c}{P\_sLSTM}
& \multicolumn{2}{c}{ModernTCN}
& \multicolumn{2}{c}{TimeMixer}
& \multicolumn{2}{c}{iTransformer}
& \multicolumn{2}{c}{TimeLLM}
& \multicolumn{2}{c}{MGSNET}
& \multicolumn{2}{c}{PatchTST}
& \multicolumn{2}{c}{TimesNet} \\

\multicolumn{2}{c}{(Years)}
& \multicolumn{2}{c}{(Ours)}
& \multicolumn{2}{c}{(2025)}
& \multicolumn{2}{c}{(2026)}
& \multicolumn{2}{c}{(2026)}
& \multicolumn{2}{c}{(2026)}
& \multicolumn{2}{c}{(2025)}
& \multicolumn{2}{c}{(2025)}
& \multicolumn{2}{c}{(2025)}
& \multicolumn{2}{c}{(2025)}
& \multicolumn{2}{c}{(2025)}
& \multicolumn{2}{c}{(2024)}
& \multicolumn{2}{c}{(2024)}
& \multicolumn{2}{c}{(2024)}
& \multicolumn{2}{c}{(2024)}
& \multicolumn{2}{c}{(2024)}
& \multicolumn{2}{c}{(2023)}
& \multicolumn{2}{c}{(2023)} \\

\cmidrule(lr){3-4}
\cmidrule(lr){5-6}
\cmidrule(lr){7-8}
\cmidrule(lr){9-10}
\cmidrule(lr){11-12}
\cmidrule(lr){13-14}
\cmidrule(lr){15-16}
\cmidrule(lr){17-18}
\cmidrule(lr){19-20}
\cmidrule(lr){21-22}
\cmidrule(lr){23-24}
\cmidrule(lr){25-26}
\cmidrule(lr){27-28}
\cmidrule(lr){29-30}
\cmidrule(lr){31-32}
\cmidrule(lr){33-34}
\cmidrule(lr){35-36}

\multicolumn{2}{c}{Metric}
& MSE & MAE & MSE & MAE & MSE & MAE & MSE & MAE
& MSE & MAE & MSE & MAE & MSE & MAE & MSE & MAE
& MSE & MAE & MSE & MAE & MSE & MAE & MSE & MAE
& MSE & MAE & MSE & MAE & MSE & MAE & MSE & MAE
& MSE & MAE \\
\midrule

% Pems03
\multirow{5}{*}{Pems03}
& 12 & \best{0.060} & \best{0.161} & 0.067 & 0.169 & 0.066 & 0.170 & 0.077 & 0.190 & 0.081 & 0.191 & \secondbest{0.064} & \secondbest{0.166} & 0.093 & 0.206 & 0.094 & 0.202 & 0.076 & 0.180 & 0.077 & 0.183 & 0.069 & 0.174 & 0.076 & 0.188 & 0.071 & 0.174 & 0.095 & 0.208 & 0.078 & 0.187 & 0.099 & 0.216 & 0.085 & 0.192 \\
& 24 & \best{0.076} & \best{0.181} & 0.085 & 0.192 & 0.090 & 0.201 & 0.097 & 0.213 & 0.122 & 0.236 & \secondbest{0.081} & \secondbest{0.186} & 0.155 & 0.269 & 0.112 & 0.222 & 0.101 & 0.210 & 0.109 & 0.219 & 0.093 & 0.203 & 0.113 & 0.226 & 0.093 & 0.201 & 0.135 & 0.248 & 0.108 & 0.218 & 0.142 & 0.259 & 0.118 & 0.223 \\
& 48 & \best{0.103} & \best{0.211} & 0.126 & 0.236 & 0.138 & 0.252 & 0.140 & 0.253 & 0.195 & 0.302 & \secondbest{0.114} & \secondbest{0.222} & 0.236 & 0.337 & 0.160 & 0.265 & 0.137 & 0.248 & 0.163 & 0.270 & 0.141 & 0.250 & 0.191 & 0.292 & 0.125 & 0.236 & 0.198 & 0.302 & 0.178 & 0.272 & 0.211 & 0.319 & 0.155 & 0.260 \\
& 96 & \best{0.143} & \best{0.248} & 0.173 & 0.283 & 0.201 & 0.311 & 0.174 & 0.289 & 0.255 & 0.354 & 0.175 & 0.283 & 0.306 & 0.399 & 0.208 & 0.307 & 0.184 & 0.292 & 0.209 & 0.309 & 0.208 & 0.306 & 0.288 & 0.363 & \secondbest{0.164} & \secondbest{0.275} & 0.243 & 0.342 & 0.238 & 0.328 & 0.269 & 0.370 & 0.228 & 0.317 \\
\cmidrule(lr){2-36}
& Avg & \best{0.096} & \best{0.200} & 0.113 & 0.220 & 0.124 & 0.234 & 0.122 & 0.236 & 0.163 & 0.271 & \secondbest{0.109} & \secondbest{0.214} & 0.198 & 0.303 & 0.144 & 0.249 & 0.125 & 0.233 & 0.140 & 0.245 & 0.128 & 0.233 & 0.167 & 0.267 & 0.113 & 0.222 & 0.168 & 0.275 & 0.151 & 0.251 & 0.180 & 0.291 & 0.147 & 0.248 \\
\midrule

% Pems04
\multirow{5}{*}{Pems04}
& 12 & \best{0.068} & \best{0.167} & \secondbest{0.069} & \secondbest{0.169} & 0.077 & 0.182 & 0.087 & 0.200 & 0.094 & 0.206 & 0.079 & 0.181 & 0.108 & 0.223 & 0.091 & 0.202 & 0.080 & 0.188 & 0.090 & 0.195 & 0.076 & 0.187 & 0.092 & 0.204 & 0.078 & 0.183 & 0.101 & 0.214 & 0.086 & 0.199 & 0.105 & 0.224 & 0.087 & 0.195 \\
& 24 & \best{0.077} & \best{0.179} & \secondbest{0.083} & \secondbest{0.188} & 0.098 & 0.211 & 0.105 & 0.221 & 0.136 & 0.251 & 0.096 & 0.203 & 0.159 & 0.271 & 0.108 & 0.220 & 0.104 & 0.218 & 0.126 & 0.233 & 0.093 & 0.206 & 0.128 & 0.243 & 0.095 & 0.205 & 0.146 & 0.259 & 0.101 & 0.218 & 0.153 & 0.275 & 0.103 & 0.215 \\
& 48 & \best{0.089} & \best{0.193} & \secondbest{0.109} & \secondbest{0.218} & 0.142 & 0.260 & 0.131 & 0.249 & 0.202 & 0.311 & 0.114 & 0.226 & 0.258 & 0.349 & 0.140 & 0.250 & 0.132 & 0.248 & 0.181 & 0.281 & 0.122 & 0.238 & 0.213 & 0.315 & 0.120 & 0.233 & 0.226 & 0.323 & 0.127 & 0.247 & 0.229 & 0.339 & 0.136 & 0.250 \\
& 96 & \best{0.107} & \best{0.216} & 0.164 & 0.274 & 0.208 & 0.321 & 0.169 & 0.283 & 0.266 & 0.368 & 0.151 & \secondbest{0.262} & 0.342 & 0.412 & 0.175 & 0.285 & 0.165 & 0.278 & 0.233 & 0.321 & 0.171 & 0.284 & 0.307 & 0.384 & \secondbest{0.150} & \secondbest{0.262} & 0.277 & 0.364 & 0.174 & 0.292 & 0.291 & 0.389 & 0.190 & 0.303 \\
\cmidrule(lr){2-36}
& Avg & \best{0.085} & \best{0.189} & \secondbest{0.106} & \secondbest{0.212} & 0.131 & 0.244 & 0.123 & 0.238 & 0.175 & 0.284 & 0.110 & 0.218 & 0.217 & 0.314 & 0.129 & 0.239 & 0.120 & 0.233 & 0.158 & 0.258 & 0.116 & 0.229 & 0.185 & 0.287 & 0.111 & 0.221 & 0.188 & 0.290 & 0.122 & 0.239 & 0.195 & 0.307 & 0.129 & 0.241 \\
\midrule

% Pems07
\multirow{5}{*}{Pems07}
& 12 & \best{0.053} & \best{0.144} & \secondbest{0.055} & \secondbest{0.150} & 0.060 & 0.158 & 0.062 & 0.165 & 0.075 & 0.182 & 0.060 & 0.156 & 0.087 & 0.199 & 0.104 & 0.187 & 0.066 & 0.165 & 0.076 & 0.169 & 0.064 & 0.165 & 0.071 & 0.170 & 0.067 & 0.165 & 0.089 & 0.202 & 0.079 & 0.182 & 0.095 & 0.297 & 0.082 & 0.181 \\
& 24 & \best{0.066} & \best{0.159} & \secondbest{0.069} & \secondbest{0.167} & 0.084 & 0.189 & 0.080 & 0.189 & 0.119 & 0.232 & 0.073 & 0.172 & 0.149 & 0.260 & 0.119 & 0.203 & 0.089 & 0.192 & 0.114 & 0.209 & 0.089 & 0.192 & 0.117 & 0.225 & 0.088 & 0.190 & 0.158 & 0.270 & 0.099 & 0.206 & 0.150 & 0.262 & 0.101 & 0.204 \\
& 48 & \best{0.092} & \best{0.181} & 0.097 & \secondbest{0.200} & 0.126 & 0.238 & 0.110 & 0.223 & 0.210 & 0.314 & \secondbest{0.096} & 0.201 & 0.282 & 0.361 & 0.153 & 0.235 & 0.122 & 0.228 & 0.186 & 0.266 & 0.130 & 0.236 & 0.201 & 0.288 & 0.110 & 0.215 & 0.232 & 0.310 & 0.133 & 0.239 & 0.253 & 0.340 & 0.134 & 0.238 \\
& 96 & \best{0.132} & \best{0.204} & \secondbest{0.133} & 0.240 & 0.183 & 0.293 & 0.141 & 0.251 & 0.305 & 0.381 & \best{0.132} & \secondbest{0.238} & 0.437 & 0.447 & 0.190 & 0.268 & 0.212 & 0.306 & 0.257 & 0.313 & 0.187 & 0.280 & 0.320 & 0.360 & 0.139 & 0.245 & 0.328 & 0.372 & 0.179 & 0.279 & 0.346 & 0.404 & 0.181 & 0.279 \\
\cmidrule(lr){2-36}
& Avg & \best{0.086} & \best{0.172} & \secondbest{0.089} & \secondbest{0.189} & 0.113 & 0.220 & 0.098 & 0.207 & 0.177 & 0.277 & 0.090 & 0.192 & 0.239 & 0.317 & 0.142 & 0.223 & 0.122 & 0.223 & 0.158 & 0.239 & 0.118 & 0.218 & 0.177 & 0.261 & 0.101 & 0.204 & 0.202 & 0.289 & 0.123 & 0.227 & 0.211 & 0.326 & 0.125 & 0.226 \\
\midrule

% Pems08
\multirow{5}{*}{Pems08}
& 12 & 0.077 & \secondbest{0.165} & \best{0.063} & \best{0.162} & 0.073 & 0.173 & 0.077 & 0.180 & 0.094 & 0.200 & \secondbest{0.072} & 0.168 & 0.106 & 0.216 & 0.086 & 0.194 & 0.075 & 0.177 & 0.088 & 0.191 & 0.082 & 0.187 & 0.091 & 0.201 & 0.079 & 0.182 & 0.107 & 0.218 & 0.105 & 0.211 & 0.168 & 0.232 & 0.112 & 0.212 \\
& 24 & 0.096 & \secondbest{0.183} & \best{0.078} & \best{0.181} & 0.100 & 0.203 & 0.113 & 0.216 & 0.143 & 0.247 & \secondbest{0.093} & 0.191 & 0.182 & 0.290 & 0.133 & 0.244 & 0.109 & 0.211 & 0.132 & 0.232 & 0.116 & 0.220 & 0.137 & 0.246 & 0.115 & 0.219 & 0.182 & 0.288 & 0.141 & 0.243 & 0.224 & 0.281 & 0.141 & 0.238 \\
& 48 & 0.128 & \best{0.211} & \best{0.105} & \secondbest{0.212} & 0.155 & 0.253 & 0.177 & 0.271 & 0.233 & 0.317 & \secondbest{0.123} & 0.217 & 0.305 & 0.354 & 0.236 & 0.265 & 0.191 & 0.278 & 0.244 & 0.274 & 0.189 & 0.287 & 0.265 & 0.343 & 0.186 & 0.235 & 0.306 & 0.330 & 0.211 & 0.300 & 0.321 & 0.354 & 0.198 & 0.283 \\
& 96 & 0.221 & \best{0.236} & \best{0.147} & \secondbest{0.259} & 0.239 & 0.313 & 0.231 & 0.310 & 0.335 & 0.390 & \secondbest{0.193} & 0.281 & 0.452 & 0.431 & 0.286 & 0.306 & 0.306 & 0.350 & 0.307 & 0.314 & 0.328 & 0.373 & 0.410 & 0.407 & 0.221 & 0.267 & 0.392 & 0.389 & 0.364 & 0.387 & 0.408 & 0.417 & 0.320 & 0.351 \\
\cmidrule(lr){2-36}
& Avg & 0.131 & \best{0.199} & \best{0.098} & \secondbest{0.204} & 0.142 & 0.236 & 0.150 & 0.244 & 0.201 & 0.289 & \secondbest{0.120} & 0.214 & 0.261 & 0.323 & 0.185 & 0.252 & 0.170 & 0.254 & 0.193 & 0.253 & 0.179 & 0.267 & 0.226 & 0.299 & 0.150 & 0.226 & 0.247 & 0.306 & 0.205 & 0.285 & 0.280 & 0.321 & 0.193 & 0.271 \\
\midrule

% ILI
\multirow{5}{*}{ILI}
& 24 & \best{1.833} & \best{0.827} & 2.231 & 0.902 & 2.539 & 0.919 & 2.763 & 0.992 & 2.412 & 1.002 & 2.180 & 0.922 & 2.728 & 1.024 & 2.231 & 0.881 & 2.369 & 0.941 & 2.548 & 0.979 & 2.299 & 0.924 & 2.110 & 0.879 & 2.004 & 0.860 & \secondbest{1.945} & \secondbest{0.837} & 3.590 & 1.026 & 2.253 & 0.896 & 2.116 & 0.952 \\
& 36 & \best{1.884} & \best{0.853} & 2.235 & 0.897 & 2.346 & 0.906 & 2.391 & 0.940 & 2.447 & 1.021 & 2.666 & 1.041 & 2.753 & 1.013 & 2.209 & 0.954 & 2.463 & 0.949 & 2.224 & 0.924 & 2.313 & 0.931 & 2.084 & 0.890 & \secondbest{1.910} & \secondbest{0.880} & 2.337 & 0.932 & 2.246 & 0.885 & 2.402 & 0.937 & 2.810 & 1.041 \\
& 48 & 2.055 & 0.865 & 2.053 & \best{0.846} & 2.095 & 0.871 & 2.204 & 0.916 & 2.432 & 1.025 & 2.535 & 1.003 & 2.553 & 0.982 & 2.227 & 0.962 & 2.234 & 0.924 & 2.111 & \secondbest{0.862} & \secondbest{2.014} & 0.875 & \best{1.961} & 0.866 & 2.036 & 0.891 & 2.203 & 0.896 & 2.092 & 0.868 & 2.138 & 0.895 & 2.465 & 0.964 \\
& 60 & 1.962 & \secondbest{0.861} & \secondbest{1.873} & \best{0.840} & 2.008 & 0.892 & 2.185 & 0.951 & 2.472 & 1.043 & 2.413 & 0.984 & 2.451 & 0.964 & 2.077 & 0.907 & 2.147 & 0.946 & 2.118 & 0.938 & \best{1.822} & 0.866 & 1.926 & 0.878 & 2.022 & 0.919 & 2.104 & 0.899 & 2.040 & 0.876 & 2.039 & 0.913 & 2.221 & 0.971 \\
\cmidrule(lr){2-36}
& Avg & \best{1.934} & \best{0.852} & 2.098 & \secondbest{0.871} & 2.247 & 0.897 & 2.386 & 0.950 & 2.441 & 1.023 & 2.449 & 0.988 & 2.621 & 0.996 & 2.186 & 0.926 & 2.303 & 0.940 & 2.250 & 0.926 & 2.112 & 0.899 & 2.020 & 0.878 & \secondbest{1.993} & 0.888 & 2.147 & 0.891 & 2.492 & 0.914 & 2.208 & 0.910 & 2.403 & 0.982 \\
\midrule

% NASDAQ
\multirow{5}{*}{NASDAQ}
& 24 & \best{0.119} & \best{0.211} & 0.129 & 0.227 & 0.128 & 0.227 & 0.131 & 0.229 & 0.127 & 0.225 & 0.127 & 0.224 & 0.141 & 0.244 & 0.129 & 0.227 & \best{0.119} & \secondbest{0.216} & 0.130 & 0.229 & 0.129 & 0.227 & \secondbest{0.122} & 0.221 & 0.137 & 0.237 & 0.133 & 0.233 & 0.186 & 0.294 & 0.127 & 0.224 & 0.198 & 0.299 \\
& 36 & \best{0.159} & \best{0.255} & 0.176 & 0.273 & 0.177 & 0.273 & 0.176 & 0.271 & 0.168 & 0.266 & 0.167 & 0.266 & 0.187 & 0.283 & 0.186 & 0.284 & \secondbest{0.165} & \secondbest{0.262} & 0.172 & 0.270 & 0.179 & 0.275 & 0.183 & 0.279 & 0.184 & 0.280 & 0.176 & 0.272 & 0.236 & 0.330 & 0.174 & 0.269 & 0.223 & 0.326 \\
& 48 & \best{0.200} & \best{0.292} & 0.220 & 0.310 & 0.215 & 0.307 & 0.218 & 0.307 & 0.215 & 0.305 & 0.227 & 0.314 & 0.229 & 0.317 & 0.223 & 0.311 & \secondbest{0.208} & \secondbest{0.298} & 0.218 & 0.306 & 0.222 & 0.312 & \best{0.200} & \secondbest{0.298} & 0.229 & 0.318 & 0.234 & 0.323 & 0.271 & 0.355 & 0.225 & 0.314 & 0.267 & 0.352 \\
& 60 & \secondbest{0.241} & \best{0.324} & 0.266 & 0.343 & 0.271 & 0.346 & 0.264 & 0.342 & 0.267 & 0.343 & 0.259 & 0.338 & 0.278 & 0.351 & 0.275 & 0.348 & 0.255 & 0.333 & 0.266 & 0.342 & 0.266 & 0.343 & \best{0.238} & \secondbest{0.328} & 0.279 & 0.352 & 0.277 & 0.350 & 0.315 & 0.385 & 0.265 & 0.339 & 0.327 & 0.394 \\
\cmidrule(lr){2-36}
& Avg & \best{0.180} & \best{0.271} & 0.198 & 0.288 & 0.198 & 0.288 & 0.197 & 0.287 & 0.194 & 0.285 & 0.195 & 0.286 & 0.209 & 0.299 & 0.203 & 0.293 & 0.187 & \secondbest{0.277} & 0.197 & 0.287 & 0.199 & 0.289 & \secondbest{0.186} & 0.282 & 0.207 & 0.297 & 0.205 & 0.295 & 0.252 & 0.341 & 0.198 & 0.287 & 0.254 & 0.343 \\
\midrule
\multicolumn{2}{l}{1\textsuperscript{st} Count}
& \multicolumn{2}{c}{\textbf{48}}
& \multicolumn{2}{c}{9}
& \multicolumn{2}{c}{0}
& \multicolumn{2}{c}{0}
& \multicolumn{2}{c}{0}
& \multicolumn{2}{c}{1}
& \multicolumn{2}{c}{0}
& \multicolumn{2}{c}{0}
& \multicolumn{2}{c}{1}
& \multicolumn{2}{c}{0}
& \multicolumn{2}{c}{1}
& \multicolumn{2}{c}{3}
& \multicolumn{2}{c}{0}
& \multicolumn{2}{c}{0}
& \multicolumn{2}{c}{0}
& \multicolumn{2}{c}{0}
& \multicolumn{2}{c}{0} \\

\bottomrule
\end{tabular}%
}
\end{table*}

\begin{table*}[!htbp]
\centering
\caption{Comprehensive performance comparison across ``multimodal scenarios for different prediction horizons (8, 10, 12). Results are highlighted strictly following the Dense Ranking principle. \textcolor{red}{\textbf{Red}}/\textcolor{blue}{\textbf{Blue}} indicate \textcolor{red}{\textbf{1st}}/\textcolor{blue}{\textbf{2nd}}}
\label{tab:comprehensive_horizons_all}
\resizebox{1.0\linewidth}{!}{%
\begin{tabular}{@{}l l *{34}{c}@{}}
\toprule

\multicolumn{2}{c}{Models}
& \multicolumn{2}{c}{MIDAPN}
& \multicolumn{2}{c}{TimeFilter}
& \multicolumn{2}{c}{VPNet}
& \multicolumn{2}{c}{CPiRi}
& \multicolumn{2}{c}{SEMixer}
& \multicolumn{2}{c}{DUET}
& \multicolumn{2}{c}{TimeKAN}
& \multicolumn{2}{c}{FilterTS}
& \multicolumn{2}{c}{TimePro}
& \multicolumn{2}{c}{P\_sLSTM}
& \multicolumn{2}{c}{ModernTCN}
& \multicolumn{2}{c}{TimeMixer}
& \multicolumn{2}{c}{iTransformer}
& \multicolumn{2}{c}{TimeLLM}
& \multicolumn{2}{c}{MGSNET}
& \multicolumn{2}{c}{PatchTST}
& \multicolumn{2}{c}{TimesNet} \\

\multicolumn{2}{c}{(Years)}
& \multicolumn{2}{c}{(Ours)}
& \multicolumn{2}{c}{(2025)}
& \multicolumn{2}{c}{(2026)}
& \multicolumn{2}{c}{(2026)}
& \multicolumn{2}{c}{(2026)}
& \multicolumn{2}{c}{(2025)}
& \multicolumn{2}{c}{(2025)}
& \multicolumn{2}{c}{(2025)}
& \multicolumn{2}{c}{(2025)}
& \multicolumn{2}{c}{(2025)}
& \multicolumn{2}{c}{(2024)}
& \multicolumn{2}{c}{(2024)}
& \multicolumn{2}{c}{(2024)}
& \multicolumn{2}{c}{(2024)}
& \multicolumn{2}{c}{(2024)}
& \multicolumn{2}{c}{(2023)}
& \multicolumn{2}{c}{(2023)} \\

\cmidrule(lr){3-4}
\cmidrule(lr){5-6}
\cmidrule(lr){7-8}
\cmidrule(lr){9-10}
\cmidrule(lr){11-12}
\cmidrule(lr){13-14}
\cmidrule(lr){15-16}
\cmidrule(lr){17-18}
\cmidrule(lr){19-20}
\cmidrule(lr){21-22}
\cmidrule(lr){23-24}
\cmidrule(lr){25-26}
\cmidrule(lr){27-28}
\cmidrule(lr){29-30}
\cmidrule(lr){31-32}
\cmidrule(lr){33-34}
\cmidrule(lr){35-36}

\multicolumn{2}{c}{Metric}
& MSE & MAE & MSE & MAE & MSE & MAE & MSE & MAE
& MSE & MAE & MSE & MAE & MSE & MAE & MSE & MAE
& MSE & MAE & MSE & MAE & MSE & MAE & MSE & MAE
& MSE & MAE & MSE & MAE & MSE & MAE & MSE & MAE
& MSE & MAE \\
\midrule

% --- Agriculture ---
\multirow{4}{*}{Agriculture}
& 8 & 0.080 & \secondbest{0.182} & 0.083 & 0.191 & 0.082 & 0.188 & \secondbest{0.079} & \secondbest{0.182} & 0.083 & 0.189 & 0.099 & 0.220 & 0.107 & 0.220 & 0.082 & 0.185 & 0.082 & 0.195 & 0.084 & 0.187 & 0.080 & \best{0.180} & 0.107 & 0.221 & \best{0.078} & 0.187 & 0.081 & 0.187 & 0.084 & 0.194 & 0.080 & 0.188 & 0.086 & 0.202 \\
& 10 & \secondbest{0.102} & \best{0.207} & 0.105 & 0.213 & 0.103 & 0.211 & \secondbest{0.102} & 0.210 & 0.103 & 0.211 & 0.105 & 0.221 & 0.124 & 0.238 & 0.105 & 0.210 & 0.105 & 0.216 & 0.109 & 0.214 & 0.103 & \secondbest{0.208} & 0.121 & 0.234 & \best{0.101} & 0.210 & 0.106 & 0.214 & 0.112 & 0.229 & 0.104 & 0.214 & 0.104 & 0.216 \\
& 12 & \best{0.131} & \best{0.232} & 0.134 & 0.237 & \best{0.131} & 0.236 & 0.135 & 0.244 & 0.134 & 0.236 & 0.134 & 0.239 & 0.154 & 0.261 & 0.133 & 0.238 & 0.133 & 0.241 & 0.137 & 0.238 & 0.136 & \secondbest{0.234} & 0.149 & 0.262 & \secondbest{0.132} & 0.235 & 0.137 & 0.240 & 0.141 & 0.248 & 0.133 & 0.240 & 0.136 & 0.248 \\
\cmidrule(lr){2-36}
& Average & \best{0.104} & \best{0.207} & 0.107 & 0.214 & \secondbest{0.105} & 0.212 & \secondbest{0.105} & 0.212 & 0.107 & 0.212 & 0.113 & 0.227 & 0.128 & 0.240 & 0.107 & \secondbest{0.211} & 0.107 & 0.217 & 0.110 & 0.213 & 0.106 & \best{0.207} & 0.126 & 0.239 & \best{0.104} & \secondbest{0.211} & 0.108 & 0.214 & 0.112 & 0.224 & 0.106 & 0.214 & 0.109 & 0.222 \\
\midrule

% --- Climate ---
\multirow{4}{*}{Climate}
& 8 & \best{0.874} & \best{0.751} & 0.902 & 0.764 & 0.903 & 0.763 & 0.904 & 0.765 & 0.905 & 0.763 & 0.897 & 0.763 & 0.906 & 0.768 & 0.916 & 0.769 & 0.941 & 0.780 & 0.920 & 0.774 & 0.904 & 0.769 & 0.906 & 0.768 & 0.908 & 0.767 & 0.925 & 0.771 & \secondbest{0.886} & 0.757 & 0.903 & 0.764 & 0.892 & \secondbest{0.756} \\
& 10 & \best{0.894} & \best{0.759} & 0.914 & 0.766 & 0.924 & 0.773 & \secondbest{0.913} & 0.767 & 0.933 & 0.775 & 0.919 & 0.766 & \secondbest{0.913} & 0.768 & 0.940 & 0.779 & 0.951 & 0.787 & 0.935 & 0.780 & 0.918 & 0.772 & \secondbest{0.913} & 0.766 & 0.922 & 0.772 & 0.919 & 0.770 & 0.919 & 0.768 & 0.916 & 0.769 & 0.920 & \secondbest{0.763} \\
& 12 & \best{0.901} & \best{0.760} & 0.927 & 0.770 & \secondbest{0.908} & 0.766 & 0.917 & 0.766 & 0.942 & 0.779 & 0.912 & 0.766 & 0.916 & 0.765 & 0.933 & 0.777 & 0.975 & 0.799 & 0.939 & 0.780 & 0.924 & 0.771 & 0.912 & \secondbest{0.763} & 0.924 & 0.772 & 0.933 & 0.774 & 0.913 & 0.766 & 0.918 & 0.766 & 0.926 & 0.764 \\
\cmidrule(lr){2-36}
& Average & \best{0.890} & \best{0.757} & 0.914 & 0.767 & 0.912 & 0.767 & 0.911 & 0.766 & 0.927 & 0.772 & 0.909 & 0.765 & 0.912 & 0.767 & 0.930 & 0.775 & 0.956 & 0.789 & 0.931 & 0.778 & 0.915 & 0.771 & 0.910 & 0.766 & 0.918 & 0.770 & 0.926 & 0.772 & \secondbest{0.906} & 0.764 & 0.912 & 0.766 & 0.913 & \secondbest{0.761} \\
\midrule

% --- Economy ---
\multirow{4}{*}{Economy}
& 8 & \best{0.195} & \best{0.349} & 0.207 & 0.363 & \secondbest{0.198} & \secondbest{0.352} & 0.211 & 0.365 & 0.206 & 0.357 & 0.211 & 0.366 & 0.228 & 0.383 & 0.211 & 0.364 & 0.204 & 0.359 & 0.223 & 0.378 & 0.206 & 0.361 & 0.215 & 0.367 & 0.214 & 0.370 & 0.205 & 0.361 & 0.229 & 0.3848 & 0.210 & 0.366 & 0.213 & 0.368 \\
& 10 & \best{0.207} & \best{0.358} & 0.218 & 0.373 & \secondbest{0.212} & \secondbest{0.364} & 0.233 & 0.385 & 0.223 & 0.373 & 0.230 & 0.383 & 0.237 & 0.391 & 0.223 & 0.372 & 0.213 & 0.367 & 0.233 & 0.386 & 0.213 & 0.366 & 0.231 & 0.381 & 0.230 & 0.384 & 0.228 & 0.382 & 0.227 & 0.3830 & 0.233 & 0.385 & 0.227 & 0.379 \\
& 12 & \best{0.211} & \best{0.363} & \secondbest{0.220} & 0.374 & 0.225 & 0.374 & 0.244 & 0.396 & \secondbest{0.220} & \secondbest{0.370} & 0.227 & 0.382 & 0.246 & 0.400 & 0.227 & 0.380 & 0.224 & 0.378 & 0.234 & 0.389 & 0.229 & 0.381 & 0.275 & 0.427 & 0.229 & 0.383 & 0.233 & 0.387 & 0.244 & 0.3984 & 0.234 & 0.386 & 0.229 & 0.382 \\
\cmidrule(lr){2-36}
& Average & \best{0.204} & \best{0.357} & 0.215 & 0.370 & \secondbest{0.212} & \secondbest{0.363} & 0.229 & 0.382 & 0.216 & 0.367 & 0.223 & 0.377 & 0.237 & 0.391 & 0.220 & 0.372 & 0.214 & 0.368 & 0.230 & 0.384 & 0.216 & 0.369 & 0.240 & 0.392 & 0.224 & 0.379 & 0.222 & 0.377 & 0.233 & 0.389 & 0.226 & 0.379 & 0.223 & 0.376 \\
\midrule

% --- Energy ---
\multirow{4}{*}{Energy}
& 8 & \best{0.159} & \best{0.309} & \secondbest{0.161} & \secondbest{0.31} & 0.167 & 0.316 & 0.167 & 0.315 & 0.164 & 0.312 & 0.164 & \best{0.309} & 0.171 & 0.315 & 0.164 & \secondbest{0.310} & 0.170 & 0.319 & 0.165 & 0.314 & 0.165 & 0.311 & 0.172 & 0.316 & 0.165 & 0.315 & 0.166 & 0.313 & 0.170 & 0.316 & 0.167 & 0.316 & 0.168 & 0.314 \\
& 10 & \secondbest{0.175} & \secondbest{0.326} & 0.177 & 0.329 & 0.180 & 0.332 & 0.186 & 0.336 & 0.183 & 0.331 & \best{0.174} & \best{0.325} & 0.186 & 0.333 & 0.181 & 0.327 & 0.185 & 0.336 & 0.182 & 0.332 & 0.177 & \secondbest{0.326} & 0.183 & 0.332 & 0.180 & 0.334 & 0.186 & 0.333 & 0.183 & 0.334 & 0.183 & 0.336 & 0.184 & 0.333 \\
& 12 & \best{0.183} & \best{0.335} & 0.189 & 0.341 & 0.188 & 0.342 & 0.190 & 0.344 & \secondbest{0.185} & \secondbest{0.338} & 0.188 & 0.339 & 0.196 & 0.343 & 0.186 & \best{0.335} & 0.191 & 0.347 & 0.194 & 0.344 & 0.192 & \secondbest{0.338} & 0.194 & 0.343 & 0.189 & 0.342 & 0.197 & 0.344 & 0.196 & 0.344 & 0.195 & 0.349 & 0.192 & 0.345 \\
\cmidrule(lr){2-36}
& Average & \best{0.172} & \best{0.323} & 0.176 & 0.327 & 0.178 & 0.330 & 0.181 & 0.332 & 0.177 & 0.327 & \secondbest{0.175} & \secondbest{0.324} & 0.184 & 0.330 & 0.177 & \secondbest{0.324} & 0.182 & 0.334 & 0.180 & 0.330 & 0.178 & 0.325 & 0.183 & 0.330 & 0.178 & 0.330 & 0.183 & 0.330 & 0.183 & 0.331 & 0.182 & 0.334 & 0.181 & 0.331 \\
\midrule

% --- Environment ---
\multirow{4}{*}{Environment}
& 8 & \secondbest{0.267} & \best{0.363} & \best{0.266} & \secondbest{0.364} & \best{0.266} & 0.369 & 0.268 & 0.367 & \best{0.266} & 0.367 & \secondbest{0.267} & 0.371 & \best{0.266} & 0.369 & \secondbest{0.267} & 0.370 & 0.269 & 0.365 & \best{0.266} & 0.37 & \best{0.266} & 0.370 & \best{0.266} & 0.369 & 0.269 & 0.370 & \secondbest{0.267} & 0.370 & \secondbest{0.267} & 0.368 & \secondbest{0.267} & 0.369 & 0.269 & 0.365 \\
& 10 & \best{0.267} & \secondbest{0.368} & \secondbest{0.268} & 0.373 & \secondbest{0.268} & 0.371 & \secondbest{0.268} & \best{0.367} & \best{0.267} & 0.369 & \secondbest{0.268} & 0.374 & 0.269 & \best{0.367} & \secondbest{0.268} & 0.373 & \secondbest{0.268} & \secondbest{0.368} & 0.269 & 0.373 & \best{0.267} & 0.371 & 0.269 & 0.372 & 0.27 & 0.373 & 0.269 & 0.373 & \secondbest{0.268} & 0.370 & \secondbest{0.268} & \secondbest{0.368} & 0.269 & \best{0.367} \\
& 12 & \best{0.269} & \secondbest{0.368} & \best{0.269} & \secondbest{0.368} & \secondbest{0.270} & 0.371 & \best{0.269} & 0.370 & \best{0.269} & 0.370 & 0.271 & 0.37 & \best{0.269} & 0.371 & \best{0.269} & 0.371 & \best{0.269} & 0.369 & \best{0.269} & 0.369 & \best{0.269} & 0.371 & \best{0.269} & 0.371 & 0.272 & \secondbest{0.368} & 0.271 & 0.373 & \secondbest{0.270} & 0.371 & 0.272 & \secondbest{0.368} & \best{0.269} & \best{0.367} \\
\cmidrule(lr){2-36}
& Average & \best{0.267} & \best{0.366} & \secondbest{0.268} & 0.368 & \secondbest{0.268} & 0.370 & \secondbest{0.268} & 0.368 & \best{0.267} & 0.369 & 0.269 & 0.372 & \secondbest{0.268} & 0.369 & \secondbest{0.268} & 0.371 & 0.269 & \secondbest{0.367} & \secondbest{0.268} & 0.371 & \best{0.267} & 0.371 & \secondbest{0.268} & 0.371 & 0.270 & 0.370 & 0.269 & 0.372 & \secondbest{0.268} & 0.370 & 0.269 & 0.368 & 0.269 & \best{0.366} \\
\midrule

% --- Health ---
\multirow{4}{*}{Health}
& 8 & 0.624 & \best{0.518} & 0.651 & 0.534 & 0.649 & 0.529 & 0.626 & 0.529 & 0.644 & 0.532 & 0.642 & 0.525 & 0.664 & 0.542 & 0.680 & 0.550 & 0.659 & 0.526 & 0.640 & \secondbest{0.524} & 0.667 & 0.539 & \secondbest{0.623} & \secondbest{0.524} & 0.666 & 0.532 & \best{0.606} & 0.525 & 0.645 & 0.528 & 0.653 & 0.535 & 0.697 & 0.543 \\
& 10 & 0.741 & 0.574 & 0.789 & 0.586 & \best{0.736} & 0.581 & 0.783 & 0.587 & 0.748 & 0.590 & 0.808 & 0.604 & 0.758 & 0.586 & 0.824 & 0.611 & 0.761 & 0.581 & 0.752 & 0.580 & 0.782 & 0.585 & 0.786 & 0.591 & 0.756 & \best{0.567} & 0.811 & 0.587 & \secondbest{0.740} & \secondbest{0.570} & 0.782 & 0.590 & 0.836 & 0.616 \\
& 12 & \best{0.832} & \secondbest{0.625} & 0.861 & 0.628 & \secondbest{0.842} & 0.629 & 0.852 & \secondbest{0.625} & 0.875 & 0.636 & 0.882 & 0.632 & 0.854 & 0.633 & 0.921 & 0.663 & 0.862 & 0.631 & 0.862 & 0.631 & 0.849 & 0.632 & 0.894 & 0.636 & 0.853 & \best{0.623} & 0.877 & 0.642 & 0.864 & 0.634 & 0.845 & 0.626 & 0.861 & 0.628 \\
\cmidrule(lr){2-36}
& Average & \best{0.732} & \best{0.572} & 0.767 & 0.583 & \secondbest{0.742} & 0.580 & 0.754 & 0.580 & 0.756 & 0.586 & 0.777 & 0.587 & 0.759 & 0.587 & 0.808 & 0.608 & 0.761 & 0.579 & 0.751 & 0.578 & 0.766 & 0.585 & 0.768 & 0.584 & 0.758 & \secondbest{0.574} & 0.765 & 0.585 & 0.750 & 0.577 & 0.760 & 0.584 & 0.798 & 0.596 \\
\midrule

% --- Security ---
\multirow{4}{*}{Security}
& 8 & \best{72.907} & \best{3.761} & 74.604 & 3.896 & \secondbest{73.436} & 3.794 & 74.098 & 3.943 & 76.199 & 3.998 & 73.443 & 3.807 & 73.741 & \secondbest{3.774} & 75.613 & 3.968 & 77.155 & 4.155 & 76.063 & 4.027 & 75.273 & 3.907 & 73.874 & 3.790 & 75.266 & 3.947 & 77.614 & 4.172 & 74.523 & 3.893 & 75.016 & 3.874 & 75.108 & 3.946 \\
& 10 & \best{74.867} & 3.862 & 75.372 & 3.944 & \secondbest{74.910} & \secondbest{3.854} & 75.773 & 3.963 & 75.957 & 3.950 & 76.350 & 3.963 & 74.979 & \best{3.851} & 76.590 & 4.079 & 76.737 & 4.048 & 76.137 & 3.963 & 75.394 & 3.907 & 75.431 & 4.000 & 76.958 & 4.068 & 76.768 & 4.118 & 76.175 & 3.969 & 75.671 & 3.936 & 75.203 & 3.930 \\
& 12 & 75.674 & \best{3.890} & 75.952 & 3.962 & \secondbest{75.662} & \secondbest{3.893} & 75.695 & 3.955 & 76.425 & 3.972 & 76.713 & 3.968 & 76.535 & 3.948 & 77.603 & 4.147 & 79.106 & 4.246 & 76.510 & 3.993 & 76.387 & 3.960 & 76.446 & 3.940 & 76.979 & 4.045 & 76.266 & 3.997 & 76.109 & 3.923 & 76.715 & 3.994 & \best{75.498} & 3.942 \\
\cmidrule(lr){2-36}
& Average & \best{74.483} & \best{3.838} & 75.309 & 3.934 & \secondbest{74.669} & \secondbest{3.847} & 75.189 & 3.954 & 76.194 & 3.973 & 75.502 & 3.913 & 75.085 & 3.858 & 76.602 & 4.065 & 77.666 & 4.150 & 76.237 & 3.994 & 75.685 & 3.925 & 75.250 & 3.910 & 76.401 & 4.020 & 76.883 & 4.096 & 75.602 & 3.928 & 75.801 & 3.935 & 75.270 & 3.939 \\
\midrule

% --- SocialGood ---
\multirow{4}{*}{SocialGood}
& 8 & \best{0.859} & 0.478 & 0.890 & 0.515 & \secondbest{0.866} & 0.482 & 0.874 & \best{0.455} & 0.946 & 0.485 & 0.880 & 0.476 & 0.891 & 0.496 & 0.873 & 0.487 & 0.880 & 0.492 & 0.887 & 0.497 & 0.983 & 0.536 & 1.021 & 0.566 & 0.892 & \secondbest{0.467} & 0.885 & 0.497 & 0.890 & 0.476 & 0.911 & 0.552 & 0.902 & 0.486 \\
& 10 & \best{0.890} & \best{0.459} & 0.947 & 0.511 & 0.949 & 0.555 & 0.933 & 0.519 & 0.976 & 0.544 & 0.928 & 0.526 & 0.941 & 0.517 & 0.928 & 0.515 & 0.938 & 0.538 & 0.935 & 0.528 & 1.035 & 0.554 & 0.981 & 0.547 & \secondbest{0.913} & 0.497 & 0.929 & \secondbest{0.480} & 0.957 & 0.515 & 0.961 & 0.572 & 0.969 & 0.554 \\
& 12 & \best{0.918} & \best{0.490} & 1.034 & 0.554 & 0.974 & 0.552 & 0.97 & 0.503 & 1.025 & 0.567 & \secondbest{0.957} & \secondbest{0.499} & 1.006 & 0.541 & 0.975 & 0.530 & 0.969 & 0.517 & 0.982 & 0.542 & 1.070 & 0.581 & 1.025 & 0.571 & 0.995 & 0.556 & 1.006 & 0.554 & 0.994 & 0.534 & 0.997 & 0.566 & 1.001 & 0.546 \\
\cmidrule(lr){2-36}
& Average & \best{0.889} & \best{0.476} & 0.957 & 0.527 & 0.930 & 0.530 & 0.926 & \secondbest{0.492} & 0.982 & 0.532 & \secondbest{0.922} & 0.500 & 0.946 & 0.518 & 0.925 & 0.511 & 0.929 & 0.516 & 0.935 & 0.522 & 1.029 & 0.557 & 1.009 & 0.561 & 0.933 & 0.507 & 0.940 & 0.510 & 0.947 & 0.508 & 0.956 & 0.563 & 0.957 & 0.529 \\
\midrule

% --- Traffic ---
\multirow{4}{*}{Traffic}
& 8 & \best{0.157} & 0.224 & 0.173 & 0.224 & 0.178 & 0.228 & 0.172 & 0.225 & 0.179 & 0.234 & 0.163 & 0.234 & 0.183 & 0.249 & 0.170 & 0.240 & 0.167 & \secondbest{0.216} & 0.172 & 0.221 & 0.17 & 0.225 & 0.179 & 0.241 & 0.170 & 0.219 & 0.171 & 0.226 & 0.178 & 0.238 & \secondbest{0.159} & \best{0.214} & 0.195 & 0.245 \\
& 10 & \best{0.157} & 0.228 & 0.174 & 0.221 & 0.179 & 0.232 & 0.169 & \best{0.216} & 0.179 & 0.231 & \secondbest{0.166} & 0.225 & 0.178 & 0.249 & 0.172 & 0.242 & 0.173 & \secondbest{0.217} & 0.171 & 0.218 & 0.172 & 0.222 & 0.176 & 0.241 & 0.172 & 0.220 & 0.171 & 0.222 & 0.182 & 0.238 & 0.173 & 0.219 & 0.197 & 0.249 \\
& 12 & \best{0.176} & 0.236 & 0.196 & 0.239 & 0.186 & 0.238 & 0.191 & 0.239 & 0.194 & 0.242 & \secondbest{0.182} & 0.242 & 0.198 & 0.261 & 0.188 & 0.255 & 0.190 & \best{0.233} & 0.191 & \secondbest{0.234} & 0.190 & 0.238 & 0.195 & 0.254 & 0.193 & 0.237 & 0.198 & 0.242 & 0.195 & 0.251 & 0.19 & \best{0.233} & 0.202 & 0.249 \\
\cmidrule(lr){2-36}
& Average & \best{0.163} & 0.229 & 0.181 & 0.228 & 0.181 & 0.233 & 0.177 & 0.227 & 0.184 & 0.236 & \secondbest{0.170} & 0.234 & 0.186 & 0.253 & 0.177 & 0.246 & 0.177 & \best{0.222} & 0.178 & \secondbest{0.224} & 0.177 & 0.228 & 0.183 & 0.245 & 0.178 & 0.225 & 0.180 & 0.230 & 0.185 & 0.242 & 0.174 & \best{0.222} & 0.198 & 0.248 \\
\midrule

% --- PTF ---
\multirow{4}{*}{PTF}
& 8 & \best{0.239} & \best{0.328} & 0.357 & 0.417 & 0.356 & 0.413 & \secondbest{0.328} & \secondbest{0.395} & 0.629 & 0.555 & 0.384 & 0.425 & 0.994 & 0.746 & 0.666 & 0.587 & 0.482 & 0.475 & 0.813 & 0.657 & 0.454 & 0.442 & 0.973 & 0.736 & 0.539 & 0.512 & 0.947 & 0.728 & 0.396 & 0.440 & 0.830 & 0.670 & 0.385 & 0.427 \\
& 10 & \best{0.248} & \best{0.333} & 0.372 & 0.439 & 0.391 & 0.439 & \secondbest{0.352} & \secondbest{0.412} & 0.723 & 0.607 & 0.432 & 0.460 & 1.125 & 0.805 & 0.747 & 0.628 & 0.477 & 0.472 & 0.882 & 0.687 & 0.494 & 0.470 & 1.041 & 0.763 & 0.594 & 0.550 & 1.018 & 0.757 & 0.437 & 0.474 & 0.935 & 0.725 & 0.423 & 0.441 \\
& 12 & \best{0.282} & \best{0.358} & \secondbest{0.354} & \secondbest{0.417} & 0.414 & 0.452 & 0.388 & 0.438 & 0.707 & 0.596 & 0.446 & 0.475 & 1.193 & 0.825 & 0.779 & 0.653 & 0.544 & 0.521 & 0.930 & 0.707 & 0.494 & 0.486 & 1.114 & 0.799 & 0.621 & 0.563 & 1.068 & 0.780 & 0.460 & 0.486 & 0.971 & 0.735 & 0.439 & 0.448 \\
\cmidrule(lr){2-36}
& Average & \best{0.256} & \best{0.340} & 0.361 & 0.424 & 0.387 & 0.435 & \secondbest{0.356} & \secondbest{0.415} & 0.686 & 0.586 & 0.421 & 0.453 & 1.104 & 0.792 & 0.731 & 0.623 & 0.501 & 0.489 & 0.875 & 0.684 & 0.481 & 0.466 & 1.043 & 0.766 & 0.585 & 0.542 & 1.011 & 0.755 & 0.431 & 0.467 & 0.912 & 0.710 & 0.416 & 0.439 \\
\midrule

% --- LEU ---
\multirow{4}{*}{LEU}
& 8 & \best{0.649} & \best{0.462} & 0.759 & 0.501 & \secondbest{0.738} & 0.493 & 0.754 & 0.505 & 0.794 & 0.521 & 0.817 & 0.482 & 0.863 & 0.502 & 0.840 & 0.495 & 0.871 & 0.497 & 0.862 & 0.497 & 0.820 & 0.488 & 0.855 & 0.495 & 0.814 & 0.485 & 0.877 & 0.509 & 0.771 & \secondbest{0.480} & 0.868 & 0.498 & 0.777 & 0.483 \\
& 10 & \best{0.660} & \best{0.470} & 0.790 & 0.531 & \secondbest{0.765} & 0.514 & 0.812 & 0.518 & 0.842 & 0.523 & 0.844 & 0.506 & 0.912 & 0.520 & 0.886 & 0.512 & 0.914 & 0.519 & 0.910 & 0.516 & 0.819 & 0.494 & 0.902 & 0.513 & 0.843 & 0.499 & 0.923 & 0.526 & 0.799 & 0.494 & 0.913 & 0.522 & 0.798 & \secondbest{0.492} \\
& 12 & \best{0.679} & \best{0.479} & 0.844 & 0.535 & \secondbest{0.793} & 0.539 & 0.825 & 0.527 & 0.865 & 0.555 & 0.866 & 0.508 & 0.948 & 0.536 & 0.916 & 0.526 & 0.955 & 0.540 & 0.948 & 0.534 & 0.837 & \secondbest{0.505} & 0.939 & 0.528 & 0.877 & 0.509 & 0.958 & 0.540 & 0.827 & 0.510 & 0.950 & 0.535 & 0.825 & 0.507 \\
\cmidrule(lr){2-36}
& Average & \best{0.663} & \best{0.470} & 0.798 & 0.522 & \secondbest{0.765} & 0.515 & 0.797 & 0.517 & 0.834 & 0.533 & 0.842 & 0.499 & 0.908 & 0.519 & 0.881 & 0.511 & 0.913 & 0.519 & 0.907 & 0.516 & 0.825 & 0.496 & 0.899 & 0.512 & 0.845 & 0.498 & 0.919 & 0.525 & 0.799 & 0.495 & 0.910 & 0.518 & 0.800 & \secondbest{0.494} \\
\midrule

% --- MSPG ---
\multirow{4}{*}{MSPG}
& 8 & \best{0.333} & \best{0.299} & 0.372 & 0.319 & \secondbest{0.363} & \secondbest{0.311} & 0.373 & \secondbest{0.311} & 0.392 & 0.325 & 0.421 & 0.333 & 0.533 & 0.388 & 0.463 & 0.353 & 0.489 & 0.369 & 0.448 & 0.343 & 0.420 & 0.331 & 0.485 & 0.369 & 0.442 & 0.336 & 0.504 & 0.382 & 0.459 & 0.358 & 0.467 & 0.349 & 0.463 & 0.354 \\
& 10 & \best{0.364} & \best{0.338} & 0.414 & 0.360 & \secondbest{0.410} & \secondbest{0.346} & 0.418 & \best{0.338} & 0.448 & 0.370 & 0.496 & 0.381 & 0.608 & 0.425 & 0.552 & 0.400 & 0.650 & 0.445 & 0.529 & 0.390 & 0.51 & 0.375 & 0.561 & 0.405 & 0.524 & 0.385 & 0.599 & 0.413 & 0.495 & 0.372 & 0.569 & 0.409 & 0.579 & 0.408 \\
& 12 & \best{0.387} & \best{0.341} & \secondbest{0.475} & \secondbest{0.380} & 0.481 & 0.387 & \secondbest{0.475} & 0.406 & 0.508 & 0.399 & 0.585 & 0.424 & 0.710 & 0.472 & 0.668 & 0.447 & 0.73 & 0.464 & 0.637 & 0.444 & 0.566 & 0.406 & 0.690 & 0.458 & 0.614 & 0.420 & 0.704 & 0.473 & 0.647 & 0.439 & 0.672 & 0.456 & 0.632 & 0.440 \\
\cmidrule(lr){2-36}
& Average & \best{0.361} & \best{0.326} & 0.420 & 0.353 & \secondbest{0.418} & \secondbest{0.348} & 0.422 & 0.352 & 0.449 & 0.365 & 0.501 & 0.379 & 0.617 & 0.428 & 0.561 & 0.400 & 0.623 & 0.426 & 0.538 & 0.392 & 0.499 & 0.371 & 0.579 & 0.411 & 0.527 & 0.380 & 0.602 & 0.423 & 0.534 & 0.390 & 0.569 & 0.405 & 0.558 & 0.401 \\
\midrule
\multicolumn{2}{l}{1\textsuperscript{st} Count}
& \multicolumn{2}{c}{\textbf{77}}
& \multicolumn{2}{c}{2}
& \multicolumn{2}{c}{3}
& \multicolumn{2}{c}{5}
& \multicolumn{2}{c}{4}
& \multicolumn{2}{c}{3}
& \multicolumn{2}{c}{4}
& \multicolumn{2}{c}{2}
& \multicolumn{2}{c}{3}
& \multicolumn{2}{c}{2}
& \multicolumn{2}{c}{6}
& \multicolumn{2}{c}{2}
& \multicolumn{2}{c}{5}
& \multicolumn{2}{c}{1}
& \multicolumn{2}{c}{0}
& \multicolumn{2}{c}{3}
& \multicolumn{2}{c}{5} \\

\bottomrule
\end{tabular}%
}
\end{table*}

\begin{table*}[!htbp]
\centering
\caption{Comprehensive performance comparison on Weather and Electricity datasets for different prediction horizons (96, 192, 336, 720). Results are highlighted strictly following the Dense Ranking principle. \textcolor{red}{\textbf{Red}}/\textcolor{blue}{\textbf{Blue}} indicate \textcolor{red}{\textbf{1st}}/\textcolor{blue}{\textbf{2nd}}}
\label{tab:weather_electricity_comparison}
\resizebox{0.88\linewidth}{!}{%
\begin{tabular}{@{}l l *{22}{c}@{}}
\toprule
\multicolumn{2}{c}{Models} & \multicolumn{2}{c}{MIDAPN} & \multicolumn{2}{c}{GTM} & \multicolumn{2}{c}{SE-LLM} & \multicolumn{2}{c}{CORA} & \multicolumn{2}{c}{Aurora} & \multicolumn{2}{c}{DualSG} & \multicolumn{2}{c}{Time-VLM} & \multicolumn{2}{c}{SEMPO} & \multicolumn{2}{c}{Sundial} & \multicolumn{2}{c}{CALF} & \multicolumn{2}{c}{MOIRAI} \\
\multicolumn{2}{c}{\multirow{1}{*}{(Years)}} & \multicolumn{2}{c}{(Ours)} & \multicolumn{2}{c}{(2026)} & \multicolumn{2}{c}{(2026)} & \multicolumn{2}{c}{(2026)} & \multicolumn{2}{c}{(2026)} & \multicolumn{2}{c}{(2025)} & \multicolumn{2}{c}{(2025)} & \multicolumn{2}{c}{(2025)} & \multicolumn{2}{c}{(2025)} & \multicolumn{2}{c}{(2025)} & \multicolumn{2}{c}{(2024)} \\
\cmidrule(lr){3-4} \cmidrule(lr){5-6} \cmidrule(lr){7-8} \cmidrule(lr){9-10} \cmidrule(lr){11-12} \cmidrule(lr){13-14} \cmidrule(lr){15-16} \cmidrule(lr){17-18} \cmidrule(lr){19-20} \cmidrule(lr){21-22} \cmidrule(lr){23-24}
\multicolumn{2}{c}{Metric} & MSE & MAE & MSE & MAE & MSE & MAE & MSE & MAE & MSE & MAE & MSE & MAE & MSE & MAE & MSE & MAE & MSE & MAE & MSE & MAE & MSE & MAE \\
\midrule

% --- Weather Data ---
\multirow{5}{*}{Weather} 
& 96  & \best{0.140} & \best{0.179} & \secondbest{0.147} & 0.197 & 0.150 & 0.200 & 0.149 & \secondbest{0.194} & 0.160 & 0.207 & 0.156 & 0.201 & 0.148 & 0.200 & 0.154 & 0.205 & 0.157 & 0.205 & 0.160 & 0.208 & 0.199 & 0.211 \\
& 192 & \best{0.190} & \best{0.228} & \secondbest{0.192} & 0.241 & 0.194 & 0.243 & 0.193 & \secondbest{0.237} & 0.202 & 0.247 & 0.196 & 0.239 & 0.193 & 0.240 & 0.198 & 0.246 & 0.205 & 0.251 & 0.205 & 0.252 & 0.246 & 0.251 \\
& 336 & 0.245 & \secondbest{0.275} & 0.250 & 0.291 & 0.247 & 0.285 & \best{0.240} & \best{0.274} & 0.252 & 0.288 & 0.248 & 0.280 & \secondbest{0.243} & 0.281 & 0.249 & 0.285 & 0.253 & 0.289 & 0.253 & 0.288 & 0.274 & 0.291 \\
& 720 & 0.316 & 0.328 & \secondbest{0.310} & 0.334 & 0.323 & 0.339 & 0.313 & \best{0.322} & \best{0.307} & \secondbest{0.327} & 0.323 & 0.334 & 0.312 & 0.332 & 0.321 & 0.337 & 0.320 & 0.336 & 0.329 & 0.341 & 0.337 & 0.340 \\
& Average & \best{0.223} & \best{0.253} & 0.225 & 0.266 & 0.229 & 0.267 & \secondbest{0.224} & \secondbest{0.257} & 0.230 & 0.267 & 0.231 & 0.264 & \secondbest{0.224} & 0.263 & 0.231 & 0.268 & 0.234 & 0.270 & 0.237 & 0.272 & 0.264 & 0.273 \\
\midrule

% --- Electricity Data ---
\multirow{5}{*}{Electricity} 
& 96  & 0.132 & \secondbest{0.224} & \secondbest{0.131} & 0.225 & \best{0.130} & 0.225 & 0.139 & 0.236 & 0.134 & 0.234 & \secondbest{0.131} & \best{0.222} & 0.142 & 0.245 & 0.136 & 0.234 & \best{0.130} & 0.227 & 0.134 & 0.231 & 0.153 & 0.241 \\
& 192 & 0.152 & \secondbest{0.242} & \secondbest{0.149} & 0.243 & \best{0.148} & \secondbest{0.242} & 0.152 & 0.248 & 0.161 & 0.258 & 0.151 & \best{0.240} & 0.157 & 0.260 & 0.153 & 0.249 & 0.150 & 0.247 & 0.153 & 0.250 & 0.169 & 0.255 \\
& 336 & \secondbest{0.164} & \best{0.255} & 0.166 & 0.259 & \secondbest{0.164} & 0.259 & 0.169 & 0.263 & 0.193 & 0.287 & \best{0.162} & \secondbest{0.256} & 0.174 & 0.276 & 0.168 & 0.267 & 0.170 & 0.268 & 0.170 & 0.268 & 0.187 & 0.273 \\
& 720 & \best{0.191} & \best{0.279} & 0.201 & \secondbest{0.292} & 0.203 & 0.293 & 0.214 & 0.299 & 0.224 & 0.320 & 0.208 & 0.296 & 0.214 & 0.308 & 0.205 & 0.303 & 0.214 & 0.307 & \secondbest{0.200} & 0.295 & 0.237 & 0.313 \\
& Average & \best{0.160} & \best{0.250} & \secondbest{0.161} & \secondbest{0.254} & \secondbest{0.161} & 0.255 & 0.169 & 0.262 & 0.178 & 0.275 & 0.163 & \secondbest{0.254} & 0.172 & 0.272 & 0.166 & 0.263 & 0.166 & 0.262 & 0.164 & 0.261 & 0.187 & 0.271 \\
\midrule
\multicolumn{2}{l}{1\textsuperscript{st} Count} & \multicolumn{2}{c}{\textbf{11}} & \multicolumn{2}{c}{0} & \multicolumn{2}{c}{2} & \multicolumn{2}{c}{\textbf{3}} & \multicolumn{2}{c}{1} & \multicolumn{2}{c}{3} & \multicolumn{2}{c}{0} & \multicolumn{2}{c}{0} & \multicolumn{2}{c}{1} & \multicolumn{2}{c}{0} & \multicolumn{2}{c}{0} \\
\bottomrule
\end{tabular}%
}
\end{table*}

\begin{table*}[!htbp]
\centering
\caption{Performance comparison on Traffic, Health, and Environment datasets. Results are highlighted strictly following the Dense Ranking principle. \textcolor{red}{\textbf{Red}}/\textcolor{blue}{\textbf{Blue}} indicate \textcolor{red}{\textbf{1st}}/\textcolor{blue}{\textbf{2nd}}}
\label{tab:multi_domain_comparison}
\resizebox{0.88\linewidth}{!}{%
\begin{tabular}{@{}l l *{18}{c}@{}}
\toprule
\multicolumn{2}{c}{Models} & \multicolumn{2}{c}{MIDAPN} & \multicolumn{2}{c}{Aurora} & \multicolumn{2}{c}{SE-LLM} & \multicolumn{2}{c}{VoT} & \multicolumn{2}{c}{Time-VLM}  & \multicolumn{2}{c}{CALF} & \multicolumn{2}{c}{ChatTime} & \multicolumn{2}{c}{GPT4MTS} & \multicolumn{2}{c}{MM-TSFLib} \\
\multicolumn{2}{c}{\multirow{1}{*}{(Years)}} & \multicolumn{2}{c}{(Ours)} & \multicolumn{2}{c}{(2026)} & \multicolumn{2}{c}{(2026)} &\multicolumn{2}{c}{(2026)} & \multicolumn{2}{c}{(2025)} & \multicolumn{2}{c}{(2025)} & \multicolumn{2}{c}{(2025)} & \multicolumn{2}{c}{(2024)} & \multicolumn{2}{c}{(2024)} \\
\cmidrule(lr){3-4} \cmidrule(lr){5-6} \cmidrule(lr){7-8} \cmidrule(lr){9-10} \cmidrule(lr){11-12} \cmidrule(lr){13-14} \cmidrule(lr){15-16} \cmidrule(lr){17-18} \cmidrule(lr){19-20}
\multicolumn{2}{c}{Metric} & MSE & MAE & MSE & MAE & MSE & MAE & MSE & MAE & MSE & MAE & MSE & MAE & MSE & MAE  & MSE & MAE & MSE & MAE \\
\midrule

% --- Traffic Data ---
\multirow{3}{*}{Traffic}
& 8   & \best{0.157} & \best{0.224} & \secondbest{0.158} & 0.286 &0.172 &\secondbest{0.229} &0.167 &0.231 & 0.212 & 0.313 & 0.176 & 0.232 & 0.257 & 0.347 & 0.195 & 0.256 & 0.195 & 0.250 \\
& 12  & \secondbest{0.176} & \secondbest{0.236} & \best{0.168} & 0.294 &0.191 &\best{0.232} &0.181 &0.239 & 0.222 & 0.322 & 0.193 & 0.243 & 0.261 & 0.349 & 0.218 & 0.268 & 0.261 & 0.272 \\
& Average & \secondbest{0.167} & \best{0.230} & \best{0.163} & 0.290 &0.182 &\secondbest{0.231} &0.174 &0.235 & 0.217 & 0.318 & 0.185 & 0.238 & 0.259 & 0.348 & 0.207 & 0.262 & 0.228 & 0.261 \\
\midrule

% --- Health Data ---
\multirow{3}{*}{Health} 
& 24  & \best{0.958} & \best{0.648} & 1.332 & 0.796 &1.232 &0.774 &\secondbest{1.210} &\secondbest{0.707} & 1.491 & 0.839 & 1.451 & 0.749 & 1.758 & 0.888 & 1.513 & 0.802 & 1.562 & 0.768 \\
& 48  & \best{1.238} & \best{0.752} & 1.579 & 0.847 &1.452 &0.836 &\secondbest{1.429} &\secondbest{0.802} & 1.702 & 0.907 & 1.836 & 0.889 & 1.930 & 0.973 & 1.757 & 0.889 & 1.997 & 0.902 \\
& Average & \best{1.098} & \best{0.700} & 1.456 & 0.822 &1.342 &0.805 &\secondbest{1.320} &\secondbest{0.755} & 1.597 & 0.873 & 1.644 & 0.819 & 1.844 & 0.931 & 1.635 & 0.846 & 1.780 & 0.835 \\
\midrule

% --- Environment Data ---
\multirow{3}{*}{Environment} 
& 96  & 0.278 & \best{0.367} & \secondbest{0.271} & \secondbest{0.373} &0.285 &\secondbest{0.373} &\best{0.270} &0.378 & 0.327 & 0.405 & 0.335 & 0.394 & 0.351 & 0.428 & 0.340 & 0.401 & 0.411 & 0.474 \\
& 192 & \best{0.268} & \best{0.363} & \secondbest{0.269} & \secondbest{0.374} &0.274 &\best{0.363} &0.273 &0.384 & 0.328 & 0.403 & 0.341 & 0.394 & 0.345 & 0.423 & 0.336 & 0.411 & 0.423 & 0.482 \\
& Average & \secondbest{0.273} & \best{0.365} & \best{0.270} & 0.374 &0.280 &\secondbest{0.368} &0.272 &0.381 & 0.328 & 0.404 & 0.338 & 0.394 & 0.348 & 0.426 & 0.338 & 0.406 & 0.417 & 0.478 \\
\midrule
\multicolumn{2}{l}{1\textsuperscript{st} Count} & \multicolumn{2}{c}{\textbf{13}} & \multicolumn{2}{c}{3} & \multicolumn{2}{c}{2} &\multicolumn{2}{c}{1} & \multicolumn{2}{c}{0} & \multicolumn{2}{c}{0} & \multicolumn{2}{c}{0} & \multicolumn{2}{c}{0} & \multicolumn{2}{c}{0} \\
\bottomrule
\end{tabular}%
}
\end{table*}

\section{Clarifications for Media Scope}

We define ``media'' in this context not as unrestricted multimedia inputs, but as heterogeneous information carriers structured as temporally aligned variables. Our framework processes both native numerical variates and text-derived pseudo-variates, which are encoded using a frozen PLM and synchronized with the temporal data. This structured representation allows us to deploy a unified forecasting architecture across a wide range of domains, including energy, transportation, weather, healthcare, finance, social systems, and agriculture. Extending this framework to raw images, audio, and video
as auxiliary information remains future work and requires more
comprehensive, temporally aligned multimodal datasets spanning diverse
application domains, together with suitable modality encoders and
alignment mechanisms.

\section{Ethical Considerations}
All experiments use publicly available and aggregated benchmark
datasets. The forecasts are intended for methodological evaluation and should not be directly used for high-stakes medical, financial, or security
decisions without domain-specific validation.

\section{GenAI Usage Disclosure}
GenAI was used solely for lightly language editing.

\end{document}